\documentclass{article} 
\usepackage{iclr2026_conference,times}

\usepackage{amsmath,amsfonts,bm}

\def\eqref#1{equation~\ref{#1}}

\def\1{\bm{1}}

\DeclareMathAlphabet{\mathsfit}{\encodingdefault}{\sfdefault}{m}{sl}
\SetMathAlphabet{\mathsfit}{bold}{\encodingdefault}{\sfdefault}{bx}{n}

\usepackage{algorithm}
\usepackage{algorithmic}
\usepackage{threeparttable}
\usepackage[utf8]{inputenc}
\usepackage[T1]{fontenc}    
\usepackage{hyperref}       
\usepackage{url}            
\usepackage{booktabs}      
\usepackage{amsfonts}       
\usepackage{nicefrac}       
\usepackage{microtype}      
\usepackage{graphicx}
\usepackage{float} 

\usepackage{amsmath}
\usepackage{amsthm}
\usepackage{amssymb}
\usepackage{array}
\usepackage{color}
\usepackage{caption}
\usepackage{mathtools}
\usepackage{mathrsfs}
\usepackage{wrapfig}
\usepackage{multirow}
\usepackage{diagbox}
\usepackage{bm}
\usepackage{tablefootnote}
\usepackage{fancybox}
\usepackage{colortbl}
\usepackage{subcaption}
\usepackage{comment}
\usepackage[table]{xcolor}
\usepackage{pifont,xcolor}
\newcommand{\bluestar}{\textcolor{blue}{\ding{72}}}
\newcommand{\redstar}{\textcolor{red}{\ding{72}}}

\usepackage{fontawesome5}

\definecolor{SDEblue}{RGB}{28 58 88}
\definecolor{cc1}{rgb}{1.0, 0.44, 0.37}
\definecolor{cc2}{rgb}{0.0, 0.2, 0.6}
\definecolor{cc3}{RGB}{255, 191, 0}
\definecolor{cc4}{RGB}{0, 128, 128}
\definecolor{cc5}{RGB}{156, 39, 176}
\definecolor{myblue}{HTML}{4dabf7}

\definecolor{myred}{HTML}{ff7777}

\hypersetup{
 colorlinks=true,
 urlcolor=magenta,
 citecolor=SDEblue,
}
\usepackage{textcomp}
\definecolor{ss1}{RGB}{165, 216, 255}
\definecolor{ss2}{RGB}{116, 192, 252}
\definecolor{ss3}{RGB}{77, 171, 247}
\definecolor{rr1}{RGB}{255, 227, 227}
\definecolor{rr2}{RGB}{255, 201, 201}
\definecolor{rr3}{RGB}{255, 168, 168}
\usepackage{tcolorbox}
\definecolor{lightroyalblue}{HTML}{F6F8FD}
\definecolor{royalblue}{HTML}{4169E1}
\definecolor{lighterblue}{HTML}{f2fafd}
\newtcolorbox{abox}{colback=lightroyalblue,colframe=black}
\definecolor{LightCyan}{rgb}{.9, .95, 1.}

\definecolor{high}{HTML}{ffe5c6}
\definecolor{lower}{HTML}{FF8787}
\definecolor{blank}{HTML}{F1F3F5}
\definecolor{blank1}{HTML}{e5dbff}

\title{The Few Govern the Many:Unveiling Few-Layer Dominance for Time Series Models}

\author{
    Xin Qiu\textsuperscript{1,2}\thanks{Equal contribution.\quad †Corresponding author.} \quad
    Junlong Tong\textsuperscript{1,3}\protect\footnotemark[\value{footnote}] \quad
    Yirong Sun\textsuperscript{1} \quad
    Yunpu Ma\textsuperscript{4} \quad
    Xiaoyu Shen\textsuperscript{1,†} \quad\\
   \textsuperscript{1}Eastern Institute of Technology, Ningbo, China \quad
   \textsuperscript{2}Zhejiang University\quad\\
   \textsuperscript{3}Shanghai Jiao Tong University\quad
   \textsuperscript{4}Ludwig Maximilian University of Munich\\
    \texttt{qiuxinzju@zju.edu.cn}\quad \texttt{xyshen@eitech.edu.cn}
}

\iclrfinalcopy 
\begin{document}

\maketitle

\begin{abstract}
Large-scale models are at the forefront of time series (TS) forecasting, dominated by two paradigms: fine-tuning text-based Large Language Models (LLM4TS) and training Time Series Foundation Models (TSFMs) from scratch. Both approaches share a foundational assumption that scaling up model capacity and data volume leads to improved performance. However, we observe a \textit{\textbf{scaling paradox}} in TS models, revealing a puzzling phenomenon that larger models do \emph{NOT} achieve better performance. Through extensive experiments on two model families across four scales (100M to 1.7B parameters) and diverse data (up to 6B observations), we rigorously confirm that the scaling paradox is a pervasive issue. We then diagnose its root cause by analyzing internal representations, identifying a phenomenon we call \textit{few-layer dominance}: only a small subset of layers are functionally important, while the majority are redundant, under-utilized, and can even distract training. Based on this discovery, we propose a practical method to automatically identify and retain only these dominant layers. In our models, retaining only 21\% of the parameters achieves up to a 12\% accuracy improvement and a 2.7$\times$ inference speedup. We validate the universality of our method on 8 prominent SOTA models (LLM4TS and TSFMs, 90M to 6B), showing that retaining less than 30\% of layers achieves comparable or superior accuracy in over 95\% of tasks.
 
 \looseness=-1  
\end{abstract}
\section{Introduction}
\begin{wraptable}[10]{r}{0.6\textwidth}
\centering
\vspace{-0.45cm}
\caption{\small Scaling paradox (Scaling model leads to higher or negligible reductions in Mean Absolute Error)}
\vspace{-0.1cm}
\resizebox{0.6\textwidth}{!}{ 
  \renewcommand{\arraystretch}{1.3} 
    \begin{tabular}{c|c|c|c}
      \toprule
      {} & Sundial$_{\text{Small}}$ $\rightarrow$ Sundial$_{\text{Large}}$ & Moirai$_{\text{Small}}$ $\rightarrow$ Moirai$_{\text{Large}}$ & Time-LLM$_{\text{GPT-2}}$ $\rightarrow$ Time-LLM$_{\text{LLaMa}}$ \\
      \midrule
       ETTh1   & \cellcolor{gray!30}+0.002 \includegraphics[width=0.06\linewidth]{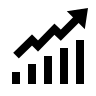} & \cellcolor{gray!30}+0.020 \includegraphics[width=0.06\linewidth]{Figure/increase.png} & \cellcolor{gray!30}+0.016 \includegraphics[width=0.06\linewidth]{Figure/increase.png}  \\
       ETTh2   & $0_{\textcolor{red}{\downarrow 0\%}}$ & \cellcolor{gray!30}+0.006 \includegraphics[width=0.06\linewidth]{Figure/increase.png} & \cellcolor{gray!30}+0.002 \includegraphics[width=0.06\linewidth]{Figure/increase.png}  \\
       ETTm1   & $-0.019_{\textcolor{red}{\downarrow 5\%}}$ & \cellcolor{gray!30}+0.006 \includegraphics[width=0.06\linewidth]{Figure/increase.png} & \cellcolor{gray!30}+0.013 \includegraphics[width=0.06\linewidth]{Figure/increase.png}  \\
       ETTm2   & $-0.009_{\textcolor{red}{\downarrow 3\%}}$ & \cellcolor{gray!30}+0.006 \includegraphics[width=0.06\linewidth]{Figure/increase.png} & $-0.007_{\textcolor{red}{\downarrow 2\%}}$  \\
       ECL     & $-0.009_{\textcolor{red}{\downarrow 1\%}}$ & $-0.004_{\textcolor{red}{\downarrow 1\%}}$ & \cellcolor{gray!30}+0.009 \includegraphics[width=0.06\linewidth]{Figure/increase.png}   \\
       Weather & \cellcolor{gray!30}+0.004 \includegraphics[width=0.06\linewidth]{Figure/increase.png} & $-0.008_{\textcolor{red}{\downarrow 3\%}}$ & $-0.003_{\textcolor{red}{\downarrow 1\%}}$  \\

      \bottomrule
      \end{tabular}
  \label{tab:scaling paradox}
\par\vspace{1em}
}
\end{wraptable}

Time series (TS) forecasting is a critical task in many real-world applications, ranging from financial analysis to climate monitoring \citep{lim2021time,tong2023probabilistic}. However, modeling long-term temporal dependencies and diverse patterns across domains remains a fundamental challenge. Inspired by the success of large-scale models, the TS community has explored two major paradigms. The first, Large Language Models for Time Series (LLM4TS), leverages the existing capabilities of text-based pretrained LLMs by aligning time series embeddings into the text modality, whether at the word \citep{jin2023time,pan2024s}, sentence \citep{liu2024unitime,liu2025timecma}, or latent-space levels \citep{liu2025calf,ICLR2025_e1de63ec}. The second, Time Series Foundation Models (TSFMs), pre-trains models from scratch purely on massive TS data to learn universal representations \citep{woo2024unified,liu2024timer}. Despite their methodological differences, both approaches share the underlying assumption that scaling up model capacity and data volume will improve performance \citep{zhang2024large, liu2025sundialfamilyhighlycapable}.

However, this foundational assumption is challenged by a series of scattered observations. For instance, in LLM4TS, employing full-scale LLMs as backbones does not necessarily yield better performance than truncating lower layers \citep{zhou2023one,liu2025timecma,ICLR2025_e1de63ec}, and advanced LLMs often fail to show expected advantages over weaker ones \citep{jin2023time}. Likewise, in TSFMs, larger models such as Moirai${_\text{Large}}$ often lag behind smaller ones like Moirai${_\text{Small}}$ \citep{goswami2024moment,das2024decoder,liu2025sundialfamilyhighlycapable}, with performance gains being marginal at best.

These findings, however, have so far been treated as isolated or scattered observations. They have lacked a comprehensive and in-depth investigation to determine if they are a systemic issue. In this paper, we are the first to formalize this problem and conduct a comprehensive study to confirm its wide existence. We term these collective observations the \textit{scaling paradox}, a phenomenon that stands in sharp contrast to the scaling laws widely observed in other domains, as evidenced by the empirical results summarized in Tab.~\ref{tab:scaling paradox}.\color{black}{\footnote{The results are reproduced from \citep{pan2024s,liu2025sundialfamilyhighlycapable}.}}

To explore this paradox, we compare {\color{cc1}{two architectural}} model families, each covering {\color{cc2}{four scales}} ((i)Tiny, (ii)Small, (iii)Base, and (iv)Large) with parameter sizes ranging {\color{cc2}{from under 100M to over 1.7B}}. These models are trained on datasets spanning data volumes {\color{cc3}{from 60K to 6B}} observations and up to {\color{cc4}{41 data distributions}}, following both {\color{cc5}{single-dataset}} and {\color{cc5}{cross-dataset learning}} strategy \citep{chang2025llm4ts}. We evaluate the in-domain and out-of-domain forecasting abilities of the 8 trained models across {\color{cc4}{16 data distributions}}.
The results confirm that the \textit{\textbf{scaling paradox}} is a pervasive phenomenon across different models and data distributions, and it is particularly pronounced in LLM4TS. 
After confirming its existence, we conducted a careful analysis of the models' internal representations to diagnose the root cause. Further analysis reveals the paradox shows weak correlations with data volume, diversity, and learning strategy. Instead, we find the cause is \textit{``\textit{\textbf{few-layer dominance}}''}: (1)\emph{ not all layers contribute effectively to the prediction process}, and (2)\emph{ retaining only a few true executors can achieve performance comparable to the full model}. This suggests that most layers in large TS models are redundant and the larger parameters are largely under-utilized. In contrast, we find they can even distract the training process, leading to performance degradation.

Based on this discovery, we propose a practical method to identify and retain only critical layers. Specifically, utilizing only \textbf{21\%} of the parameters achieves superior performance, leading to a \textbf{2.7×} faster inference speed and up to a \textbf{12\%} improvement in accuracy, while the performance degradation in the worst case is as small as \textbf{0.6\%} in our model families. Moreover, we validate the universality of our method by applying it to \textbf{8 famous TS models} (4 LLM4TS \& 4 TSFMs). The model architectures cover three paradigms: {\color{cc1}{encoder-only}}, {\color{cc1}{decoder-only}}, and {\color{cc1}{encoder–decoder}}, while the TS embeddings operate at three levels: {\color{cc1}{point-wise}}, {\color{cc1}{patch-wise}}, and {\color{cc1}{variable-wise}}. Their model sizes range {\color{cc2}{from 90M to 6B}}. The TSFMs vary in their pre-training scales {\color{cc3}{from 100B to 1,000B}} observations, and LLM4TS differ in prompt designs and alignment strategies (word level, sentence level, and latent-space level). Experiments are conducted on {\color{cc4}{13 data distributions}} with three input lengths and four forecasting horizons. Remarkably, by retaining less than \textbf{30\%} of the critical layers on average, our method achieves comparable or even superior forecasting accuracy in over \textbf{95\%} of tasks, with the average accuracy loss in the remaining cases kept below \textbf{0.5\%}. Our main contributions are threefold:
\begin{enumerate}
    \item We provide the first analysis of a \emph{scaling paradox} in large-scale TS models, rigorously demonstrating that \emph{"larger is not always better"} across both TSFM and LLM4TS paradigms.

    \item We diagnose the root cause of this paradox, identifying \emph{few-layer dominance}, where only a small subset of layers are functionally important.

    \item We propose a practical \emph{pruning method} that identifies and retains the critical layers, producing smaller and faster models that maintain or even improve forecasting accuracy.
\end{enumerate}
\vspace{-0.1cm}
\section{Related Work}
\vspace{-0.1cm}
\paragraph{Large-scale Time Series Models.}
Large-scale TS models consist of two categories: LLM4TS and TSFMs. Applying pre-trained LLMs to TS poses a fundamental challenge, achieving effective modality alignment between textual and temporal representations. \citep{zhou2023one,jin2023time,ICLR2025_e1de63ec,liu2025calf}. Existing approaches either preserve the original LLM architecture or truncate the shallow layers to improve computational efficiency \citep{zhou2023one,jin2023time}. The former often introduces task-irrelevant computational overhead, whereas the latter compromises the model’s linguistic expressiveness. TSFMs aim to build general-purpose TS models, enabling cross-domain generalization \citep{ansari2024chronos,goswami2024moment,wang2025lightgts}. Although scaling up model size has become a common strategy to enhance performance \citep{wilinski2024exploring,das2024decoder,goswami2024moment,xiaoming2025time}, empirical observations indicate that forecasting gains are often marginal and may even degrade in certain cases.

\vspace{-0.3cm}
\paragraph{Redundancy Analysis in Large-scale Models.}

Redundancy has been extensively studied in large-scale models across language \citep{yao2022zeroquant,ma2023llm, men2024shortgpt,zhao2025skipgpt} and vision domains \citep{sung2023ecoflap,fan2025visipruner,liang2025efficientllava}. These studies demonstrate that a substantial amount of parameter redundancy exists within such models, yet observes that performance still scales positively with model size \citep{kaplan2020scaling, aghajanyan2023scaling, chen2024internvl}. This phenomenon is commonly attributed to the strong representational capacity and optimization stability brought by overparameterization. In contrast, our findings on large-scale TS models reveal a scaling paradox, where enlarging model size does not necessarily yield better forecasting performance, yet the underlying causes remain largely unexplored.
\vspace{-0.1cm}
\section{Background}

\vspace{-0.2cm}
In this section, we provide the necessary background, including task definitions (Sec. \ref{subsec:Time Series Forecasting Task}), general architectures and core components of LLM4TS (Sec. \ref{subsec:Architecture of LLM4TS}) and TSFMs (Sec. \ref{subsec:Architecture of TSFMs}). Based on them, we design the model families employed in the experimental analysis of our work.

\subsection{Time Series Forecasting Task}
\label{subsec:Time Series Forecasting Task}

\noindent Let $\boldsymbol{X} \in \mathbb{R}^{T \times V}$ denote the historical sequence of TS, where $T$ is sequence length and $V$ is variable dimension. The objective of TS forecasting is to learn a mapping function $\boldsymbol{F}_{\boldsymbol{\Theta_F}}: \mathbb{R}^{T \times V} \to \mathbb{R}^{T' \times V}$, parameterized by $\boldsymbol{\Theta_F}$, that predicts the future sequence over next $T'$ time steps. Given a collection of $K$ distinct TS data distributions $\{\mathcal{D}_1, \mathcal{D}_2, \dots, \mathcal{D}_K\}$, forecasting task can be formulated as:
\begin{align}
    \hat{\boldsymbol{X}}_{[T:T+T')} &= \boldsymbol{F}_{\boldsymbol{\Theta_F}}\big( \boldsymbol{X}_{[0:T)} \big), 
    \boldsymbol{\Theta_F}^* = \arg\min_{\boldsymbol{\Theta_F}} \ \mathbb{E}_{k \sim \{1,\dots,K\}} \ \mathbb{E}_{\boldsymbol{X}, \hat{\boldsymbol{X}} \sim \mathcal{D}_k} \Big[ {\mathcal{L}}\big(\hat{\boldsymbol{X}}, \boldsymbol{X}_{[T:T+T')} \big) \Big].
\end{align}
\vspace{-0.3cm}

\subsection{Architecture of LLM4TS}
\label{subsec:Architecture of LLM4TS}
The main challenge of LLM4TS is to represent TS in a way that LLMs can effectively process, while minimizing the modality gap through proper strategies. Its \textbf{input embed} module consists of a TS embed module, which maps TS into embedding, and a tokenizer that injects prompts into the LLM. The \textbf{LLM4TS backbone} then aligns the TS embeddings with the LLM representation space through fine-tuning. The aligned TS embeddings are passed through a \textbf{prediction head} to generate the prediction results, with the model optimized by minimizing the MSE objective.
\vspace{-0.25cm}
\paragraph{Input embed.} Patch-wise \citep{Yuqietal-2023-PatchTST} and variable-wise \citep{ICLR2024_2ea18fdc} TS embed are widely used to transform TS into embeddings. The LLM tokenizer encodes the prompts and injects them into the model, allowing the LLM to exploit its intrinsic understand and reasoning advantages.
\vspace{-0.25cm}
\paragraph{LLM4TS backbone.} The backbone can be instantiated with LLMs such as GPT-2 \citep{radford2018improving}, LLaMA \citep{touvron2023llamaopenefficientfoundation}, or Qwen-3 \citep{yang2025qwen3technicalreport}. By fine-tuning, the model performs alignment at the word, sentence, or latent-space levels, strengthening its understanding of historical patterns and reasoning over future steps \citep{pan2024s,liu2025timecma,ICLR2025_e1de63ec}. 
\vspace{-0.65cm}
\paragraph{Prediction head.} It serves as a decoder that transforms TS embeddings into prediction results. It typically adopts an autoregressive or a parallel decoding strategy. The parallel decoding paradigm mitigates error accumulation and lowers decoding latency \citep{Zhou_Zhang_Peng_Zhang_Li_Xiong_Zhang_2021,zeng2023transformers}.

\subsection{Architecture of TSFMs}
\label{subsec:Architecture of TSFMs}
Compared with LLM4TS, TSFMs feature a more flexible architecture, consisting of a \textbf{TS tokenization} module and a \textbf{TSFM backbone}. Instead of adapting pre-trained LLMs, they design specialized structures and temporal modeling modules to capture dependencies offering greater adaptability to diverse forecasting tasks \citep{chen2023long}. TSFMs can also employ more diverse supervision losses, allowing greater flexibility in optimizing objectives.
\vspace{-0.25cm}
\paragraph{TS Tokenization.} Besides patch-based embedding, TSFMs may encode TS data through scaling and quantization, or alternatively, adopt point-wise tokenization to preserve the completeness of temporal information \citep{ansari2024chronos,liu2025sundialfamilyhighlycapable}. 
\vspace{-0.25cm}
\paragraph{TSFMs backbone.} 
Beyond LLM4TS typically adopts a decoder-only architecture to predict the next token, some TSFMs instead follow an encoder–decoder paradigm to update temporal tokens. Meanwhile, an explicit prediction head is unnecessary, since the backbone itself can map tokens back to numerical values through strategies such as autoregressive sampling or multi-scale forecasting \citep{das2024decoder,xiaoming2025time}.

\begin{figure}[h]
\vspace{-0.5cm}
    \centering
    \includegraphics[width=0.95\linewidth]{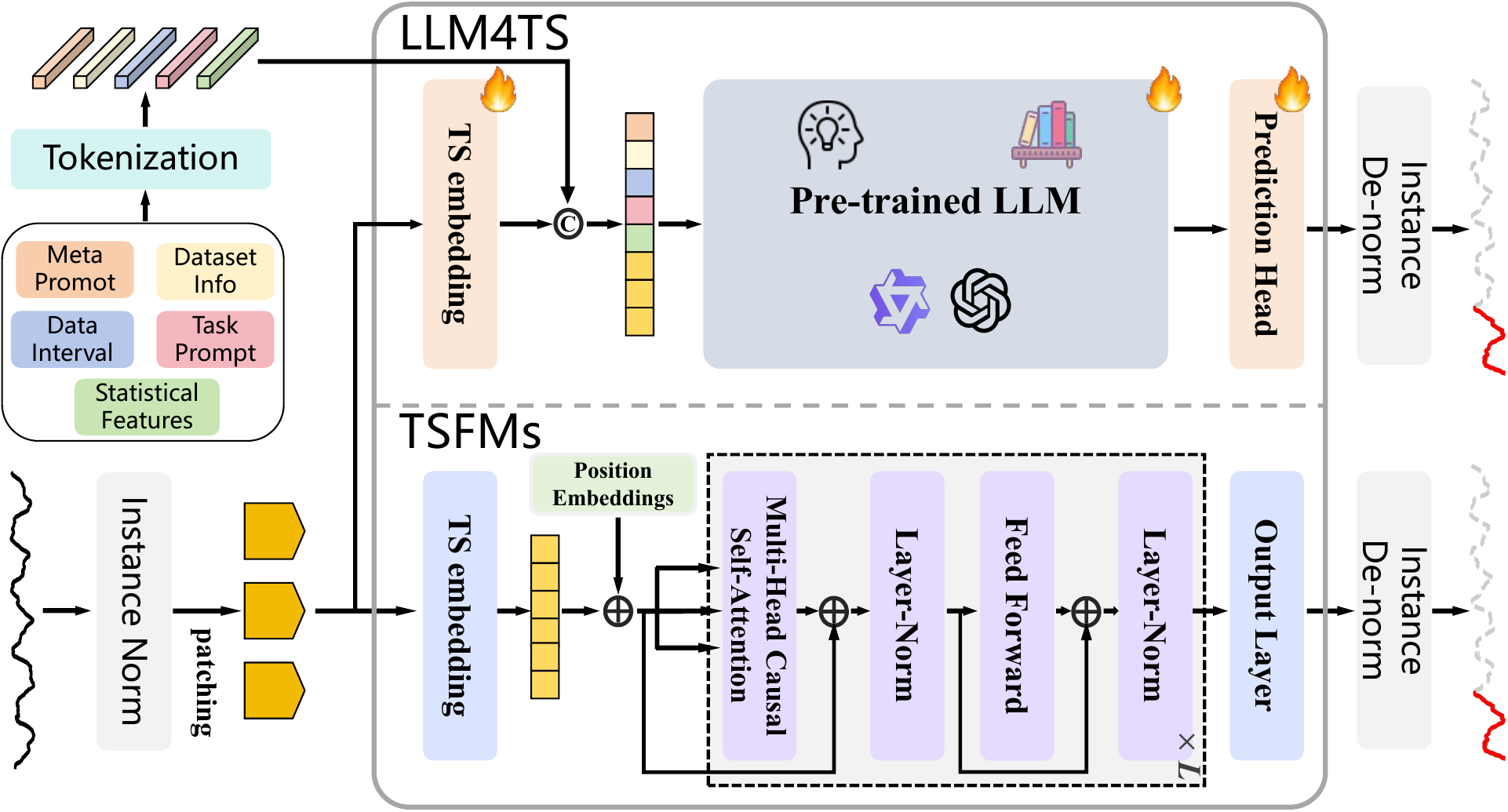}
    \vspace{-0.2cm}
    \caption{Overall structure of our LLM4TS and TSFMs families. Each family consists of four models at different scales. LLM4TS fine-tunes only a small subset of parameters to transfer the extensive knowledge and reasoning capability of LLMs to TS tasks, while incorporating prompts to enhance representation learning. TSFMs adopt a decoder-only architecture that predicts the next token to forecast future.}
    \label{fig:main architexture}
\end{figure}

\begin{figure}[h]
\vspace{-0.1cm}
    \centering
    \includegraphics[width=0.95\linewidth]{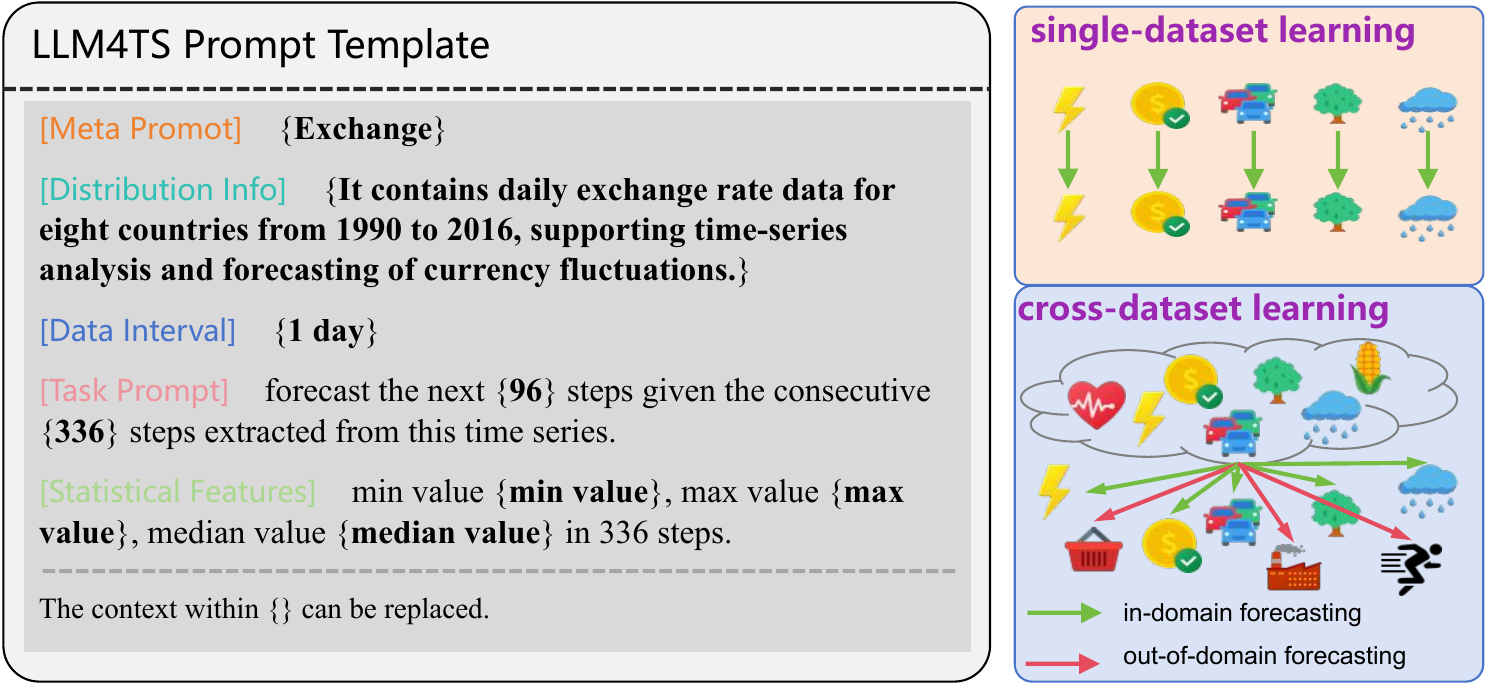}
    \vspace{-0.2cm}
    \caption{\textbf{(\emph{Left})} The prompt consists of five subtypes, illustrated here using Exchange dataset as an example. \textbf{(\emph{Right})} In single-dataset learning, the model is trained in a one-to-one manner on individual datasets, while in cross-dataset learning, a many-to-many scheme is adopted to enhance the model’s generalization across datasets.}
    \label{fig:Prompt Template}
\end{figure}
\vspace{-0.3cm}
\section{Scaling Paradox}
\label{section:Scaling Paradox}
\vspace{-0.3cm}
In this section, we systematically investigate the \textit{scaling paradox} phenomenon observed in large-scale TS models.
 Sec. \ref{subsec:Prelimilary} provides an overview of our prior work, detailing the constructed TS model family, the employed datasets, training strategies, and evaluation schemes. Based on this foundation, the following sections investigate \emph{five potential factors} affecting forecasting performance,  and analyze external factors related to the scaling paradox.

\subsection{Analysis Framework}
\label{subsec:Prelimilary}
We systematically investigate key factors of TS forecasting task, including {\color{cc1}{architecture}}, {\color{cc2}{model size}}, {\color{cc3}{data volume}}, {\color{cc4}{distribution}}, and {\color{cc5}{learning strategies}}, to illustrate how they influence forecasting performance and the emergence of scaling paradox.

\vspace{-0.25cm}

\paragraph{Model Architecture.} We employed {\color{cc1}{two representative architectural designs}} that are prevalent in contemporary TS research~\citep{zhou2023one,das2024decoder}, which also avoids confounding effects from introducing too many variables at once. To mitigate semantic shifts from uneven data distributions, we apply instance normalization and patch-based embedding~\citep{Yuqietal-2023-PatchTST}. Since the variables are mostly independent~\citep{chencloser}, we adopt a channel-independent design to reduce overfitting. For the LLM4TS model family, only positional encoding and layer normalization are fine-tuned to avoid catastrophic forgetting~\citep{zhou2023one}. We selected lightweight GPT-2~\citep{radford2018improving} and more powerful Qwen-3~\citep{yang2025qwen3technicalreport} as the LLM4TS backbones. For the TSFMs model family, they are trained end-to-end with full-parameter. A detailed schematic of the overall architecture is provided in Fig.~\ref{fig:main architexture}.

\paragraph{Model Size.} 
We investigate {\color{cc2}{four distinct model scales}}: (i)Tiny, (ii)Small, (iii)Base, and (iv)Large. As detailed in Tab.~\ref{Main model parameters}, we control the configurations for both the LLM4TS and TSFMs families at each scale, such as the number of layers and backbone choice. This ensures that their learnable parameter counts are approximately comparable within each corresponding size, facilitating a fair comparison.
\begin{table}[h]
\centering
\caption{Main model parameters. For LLM4TS, the learnable parameters exclude the frozen components within the underlying LLMs, while for TSFMs, since no modules are frozen. “–” indicates no pre-training (trained from scratch).}
\vspace{-0.2cm}
\label{Main model parameters}
\resizebox{1\linewidth}{!}{
  \begin{tabular}{lcccccc}
    \toprule
    \textbf{Scale} & \textbf{Model family} & \textbf{Layers} & \textbf{Backbone} & \textbf{Channels} & \textbf{Learnable Parameters} & \textbf{Token Level}\\
    \midrule
    Tiny  & LLM4TS / TSFMs & 6 / 6  & GPT-2 / -& 768 / 768   & 3.92M / 85M & Patch / Patch \\
    Small & LLM4TS / TSFMs & 12 / 12 & GPT-2 / -& 768 / 768   & 3.93M / 128M & Patch / Patch\\
    Base & LLM4TS / TSFMs & 28 / 28 & Qwen-3 / -& 1,024 / 1,024 & 0.16B / 0.60B & Patch / Patch\\
    Large & LLM4TS / TSFMs & 28 / 28 & Qwen-3 / -& 2,048 / 2,048 & 0.32B / 1.73B & Patch / Patch\\
    \bottomrule
  \end{tabular}
}
\end{table}
\vspace{-0.4cm}
\paragraph{Data Volume.} As shown in Tab.~\ref{tab:Statistics of datasets}, the nine selected distributions differ widely in size, containing between {\color{cc3}{60K and 15M}} observations. Collectively, they encompass roughly {\color{cc3}{33M}} records, providing a rich and diverse basis for subsequent analysis.

\paragraph{Data Distribution.} The TS datasets are derived from {\color{cc4}{nine real-world data distributions}}, as detailed in Tab.~\ref{tab:Statistics of datasets}. These datasets were intentionally selected to cover a wide spectrum of temporal characteristics, including different levels of seasonality, trend, and noise. 
\begin{table}[h]
\caption{Statistics of datasets, collected from multiple real-world scenarios \citep{ICLR2024_2ea18fdc}.}
\vspace{-.4cm}
\begin{center}
\resizebox{1\textwidth}{!}{
\begin{tabular}{l|ccccccccc}
\toprule
\textbf{Datasets} & \textbf{ETTh1} & \textbf{ETTh2} & \textbf{ETTm1} & \textbf{ETTm2} & \textbf{ECL} & \textbf{Exchange} & \textbf{Solar} & \textbf{Weather} & \textbf{Traffic}\\
\midrule
Variates  & 7 & 7  & 7 & 7 & 321 & 8  & 137 & 21 & 862\\
Observations & 121,940 & 121,940 & 487,760 & 487,760 & 8,443,584 & 60,704 & 7,200,720 & 1,106,616 & 15,122,928\\
\bottomrule
\end{tabular}
}
\end{center}
\vspace{-.4cm}
\label{tab:Statistics of datasets}
\end{table}
\vspace{-0.3cm}
\paragraph{Learning Strategies.} We employ two different strategies: {\color{cc5}{single-dataset learning}} and {\color{cc5}{cross-dataset learning}} strategy, shown in Fig.~\ref{fig:Prompt Template}. In the former, each model is trained and evaluated on a single dataset (e.g., ETTh1) to measure in-domain performance. In the latter, models are trained on the combined training sets and share weights across the test sets of all datasets~\citep{chang2025llm4ts}. 
We follow \textsc{Fsca} \citep{ICLR2025_e1de63ec} for dataset partitioning, using a input of 336 steps to forecast a horizon of 96. In the single-dataset learning setting, models are evaluated on the same distribution to measure in-domain performance. cross-dataset learning additionally tests on on {\color{cc4}{eight new distributions}} (NN5, PDB, Sceaux, Smart, Spanish, Sunspot Rain, US Births, and Wind Power) to assess out-of-domain generalization \citep{goswami2024moment}.

To structure our investigation, we organize the analysis of these five factors into three key Research Questions (RQs). RQ1 (Sec. \ref{subsection:Does Backbone Scaling Improve Forecasting Performance? (RQ1)}) addresses the most fundamental factors, model architecture and model size, to determine if the scaling paradox is universal. RQ2 (Sec.~\ref{subsection:Does Limited Data Volumn Hinder the Advantages of Backbone? (RQ2)}) focuses on examining how variations in data volume within the same distribution influence the observed paradox. Finally, RQ3 (Sec.~\ref{Does Performance Degradation Stem From Dataset Homogeneity? (RQ3)}) investigates how the interplay between data distribution and learning strategies (particularly cross-dataset learning) influences the emergence and severity of the paradox.

\subsection{Examining the Universality of the Scaling Paradox (RQ1)}
\label{subsection:Does Backbone Scaling Improve Forecasting Performance? (RQ1)}
\begin{figure}[h]
    \centering
    \vspace{-0.1cm}
    \begin{subfigure}{0.48\linewidth}
        \centering
        \includegraphics[width=\linewidth]{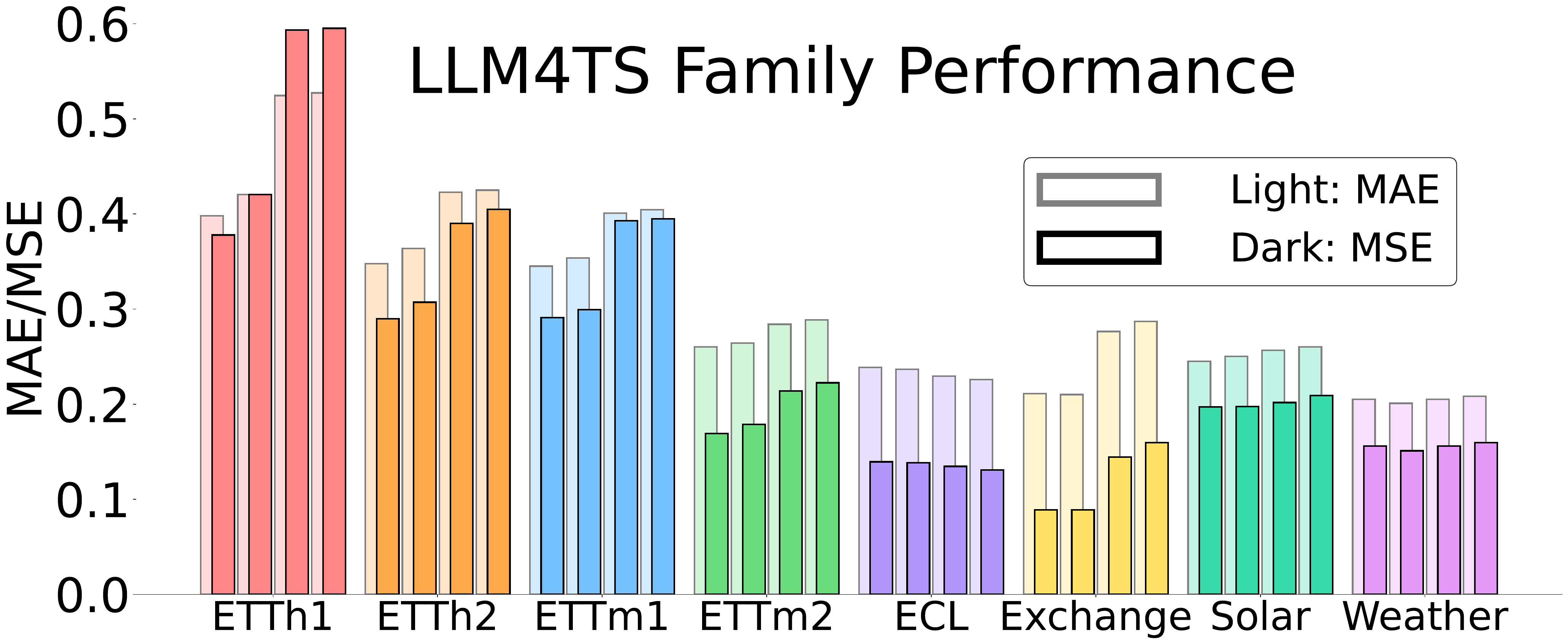}
    \end{subfigure}
    \begin{subfigure}{0.48\linewidth}
        \centering
        \includegraphics[width=\linewidth]{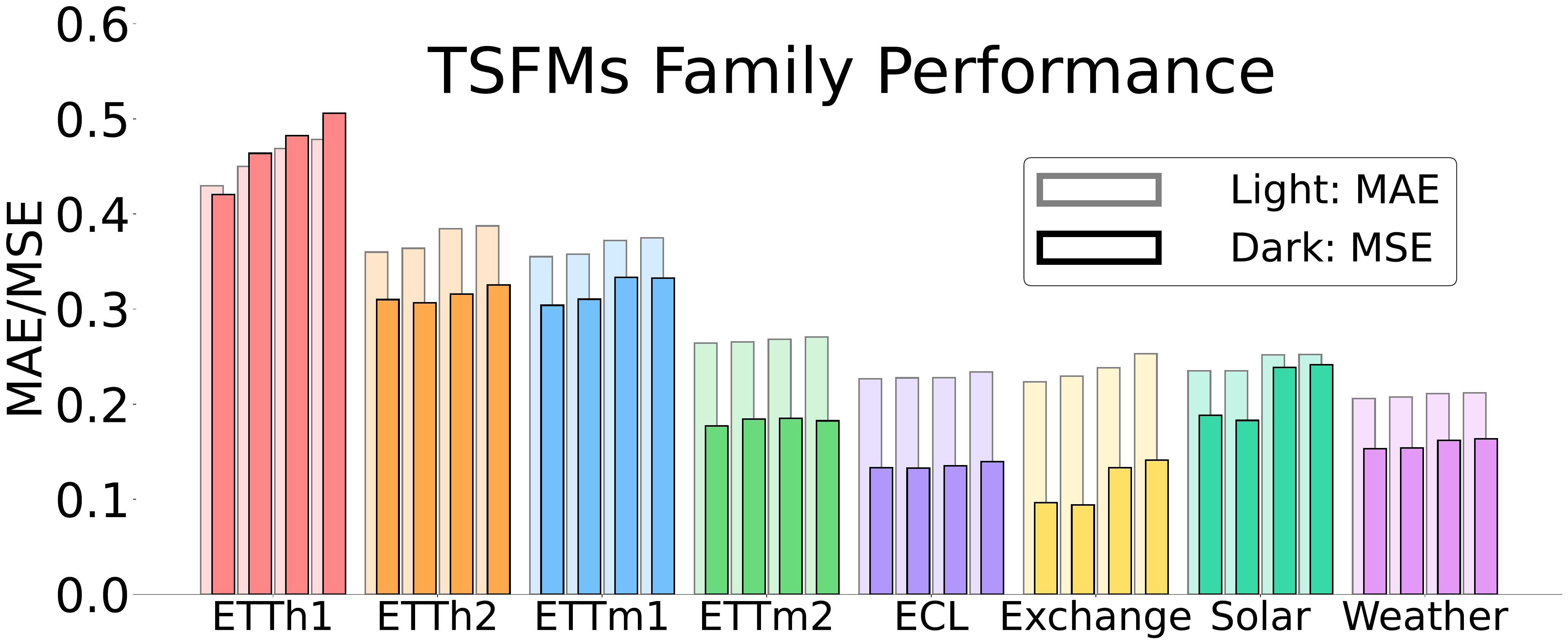}
    \end{subfigure}
    \captionsetup{justification=justified, singlelinecheck=false}
    \caption{Average performance across different model scales and data distribution. \textbf{(\emph{Left})} Metrics of the LLM4TS family. \textbf{(\emph{Right})} Metrics of the TSFMs family. Lighter bars is
    MAE, and darker bars is MSE. For each dataset, the 4 bars from {\color{cc2}{\emph{“left to right”} represent \emph{“Tiny to Large”}}}. The scaling paradox appears across most data distributions and is more pronounced in LLM4TS.
} 
    \label{fig:backbone-scaling}
    \vspace{-.5cm}
\end{figure}

\paragraph{Rationale.} Empirically, regardless of the large-scale TS models paradigm, a larger model enhances performance, whether strength comes from powerful prior knowledge endowed by LLMs in LLM4TS, or from the fully learnable capacity of TSFMs \citep{Chowdhery2022PaLMSL,zhuocheng-etal-2023-scaling,isik2024scaling}. However, in the field of TS, this theory does not seem to hold. 
\vspace{-0.2cm}
\paragraph{Setup.} We adopt a {\color{cc5}{single-dataset learning}} strategy to evaluate model performance across {\color{cc4}{eight datasets}} with {\color{cc3}{different data volumes}}, with a particular focus on investigating whether the scaling paradox is consistently observed across the {\color{cc1}{two architectures}} within model family.
\vspace{-0.2cm}
\paragraph{Results.} Fig. \ref{fig:backbone-scaling} shows that MAE \& MSE do not significant decrease as the model scales up. Interestingly, across most datasets, larger models tend to underperform, irrespective of the dataset size, be it the large-scale ECL or the relatively small ETTh. It means that neither the prior knowledge in bigger LLM4TS nor the stronger parameterization in larger TSFMs leads to better performance. 

\begin{abox} 
    \looseness -1 \textbf{Takeaways 1:} Scaling paradox is a pervasive phenomenon in TS modeling, consistently observed across varying data distributions and volumes, and appears to be more pronounced in LLM4TS family.
\end{abox}
\vspace{-0.2cm}
\subsection{Impact of Intra-domain Data Scale on the Scaling Paradox (RQ2)}
\label{subsection:Does Limited Data Volumn Hinder the Advantages of Backbone? (RQ2)}

\begin{figure}[h]
\vspace{-0.0cm}
    \centering
    \includegraphics[width=0.9\linewidth]{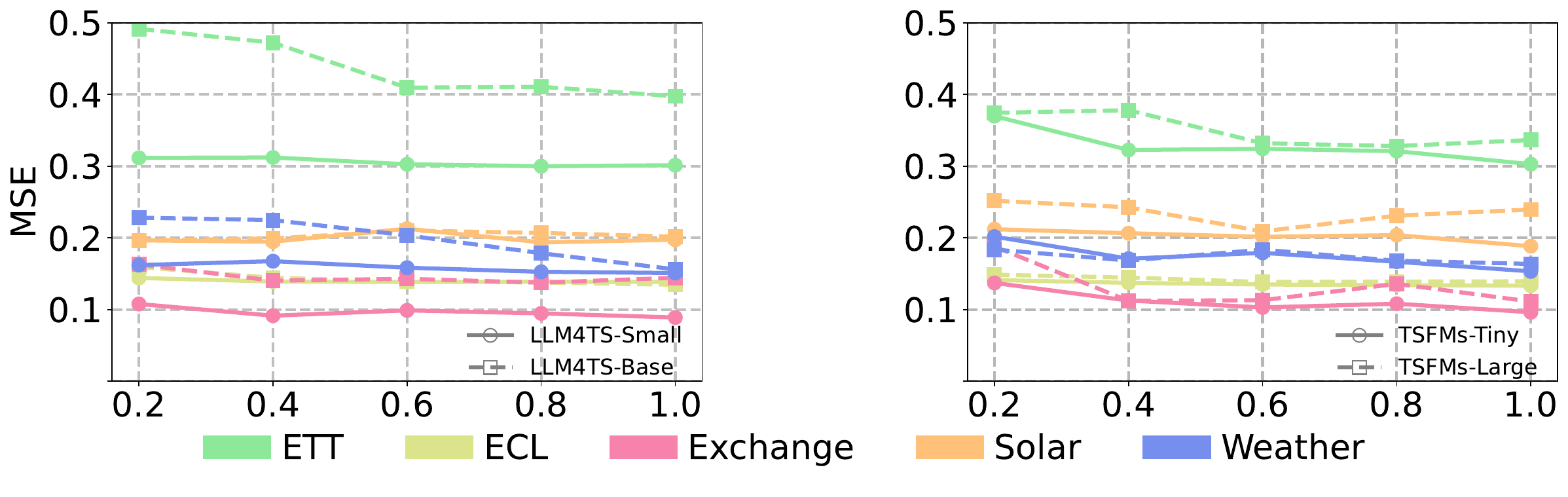}
    \vspace{-0.2cm}
    \caption{MSE under data ratios from 20\% to 100\%. \textbf{(\emph{Left})} MSE of LLM4TS at \emph{Small} and \emph{Base} scales. \textbf{(\emph{Right})} MSE of TSFMs at the \emph{Tiny} and \emph{Large} scales. In the two subfigures, the MSE of large and small models decreases as ratio increases, but the MSE gap does not consistently narrow.}
    \label{fig:Performance under training data ratios}
\end{figure}

\vspace{-0.2cm}
\paragraph{Rationale.} Across almost all datasets, despite large variations in distribution and data volume (from 60K to 15M), model scaling yields no significant performance gains (Fig. \ref{fig:backbone-scaling}). Therefore, in this section, we investigate how data volume relates to the scaling paradox under a consistent distribution.
\vspace{-0.3cm}
\paragraph{Setup.} We conduct experiments using two scales from each family: {\color{cc2}{Small and Large}} for {\color{cc1}{LLM4TS}}, and {\color{cc2}{Tiny and Large}} for {\color{cc1}{TSFMs}}. We use {\color{cc4}{eight datasets}} and {\color{cc5}{single-dataset learning}} as in Sec. \ref{subsection:Does Backbone Scaling Improve Forecasting Performance? (RQ1)}, while varying the training data ratio {\color{cc3}{from 20\% to 100\%}} with fixed validation and test sets. 
\vspace{-0.2cm}
\paragraph{Results.}As the data volume gradually increases from 20\%, MAE and MSE exhibit a downward trend. However, larger models still underperform or perform comparably to their smaller counterparts, and the MSE gap between them does not significantly narrow, as shown in Fig. \ref{fig:Performance under training data ratios}. It suggests that increasing data volumes helps to improve the performance of both small and large models, but it does not alleviate the scaling paradox — larger models offer no consistent gains.

\begin{abox} 
    \looseness -1 \textbf{Takeaways 2:} Under the same data distribution, increasing the training data improves overall performance but brings no significant relief to the scaling paradox.
\end{abox}

\subsection{Impact of Data Distribution Diversity and Learning Strategies on the Scaling Paradox (RQ3)}
\label{Does Performance Degradation Stem From Dataset Homogeneity? (RQ3)}

\begin{figure}[h]
    \centering
    \vspace{-0.8cm}
    \begin{minipage}[h]{0.36\linewidth}
        \centering
        \vspace{-0.5cm}
        \includegraphics[width=\linewidth]{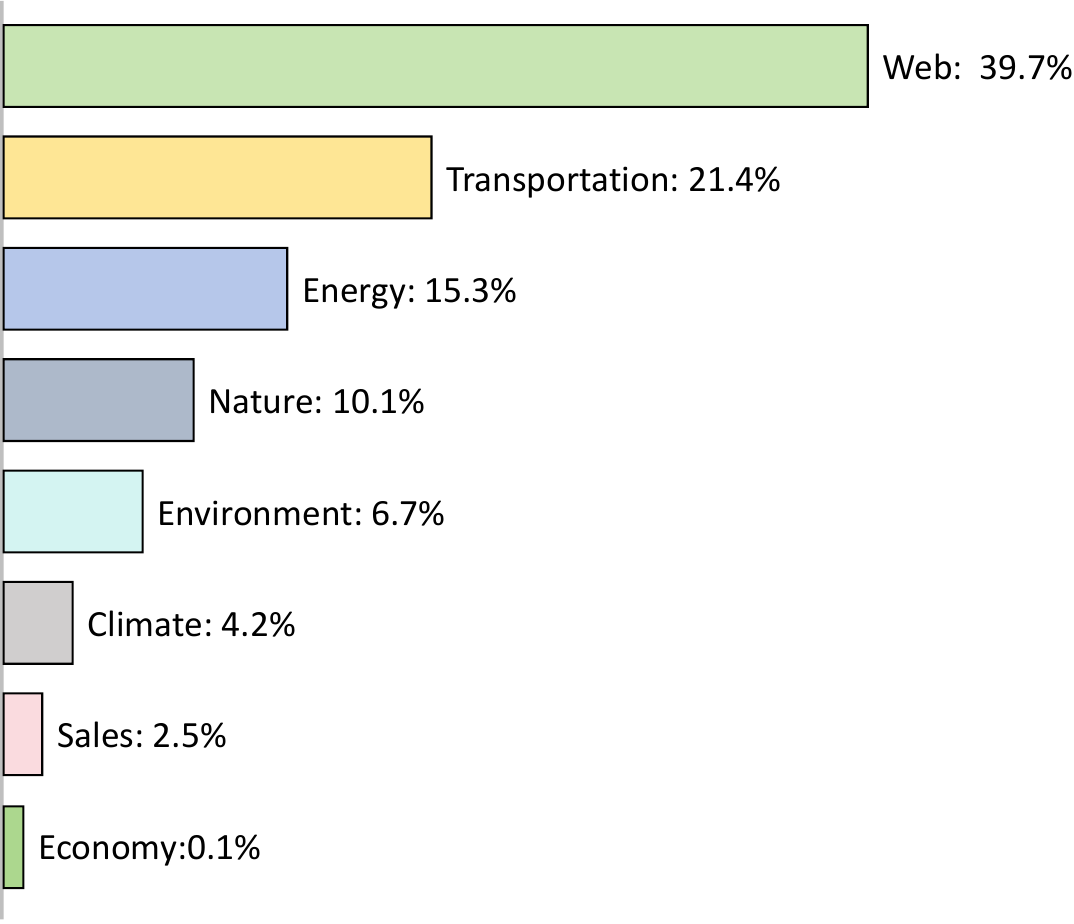}
        \vspace{-0.4cm}
        \caption{Dataset sources.}
        \label{fig:Dataset sources}
    \end{minipage}
    \hspace{0.1\linewidth}
    \begin{minipage}[h]{0.5\linewidth}
        \centering
        \includegraphics[width=\linewidth]{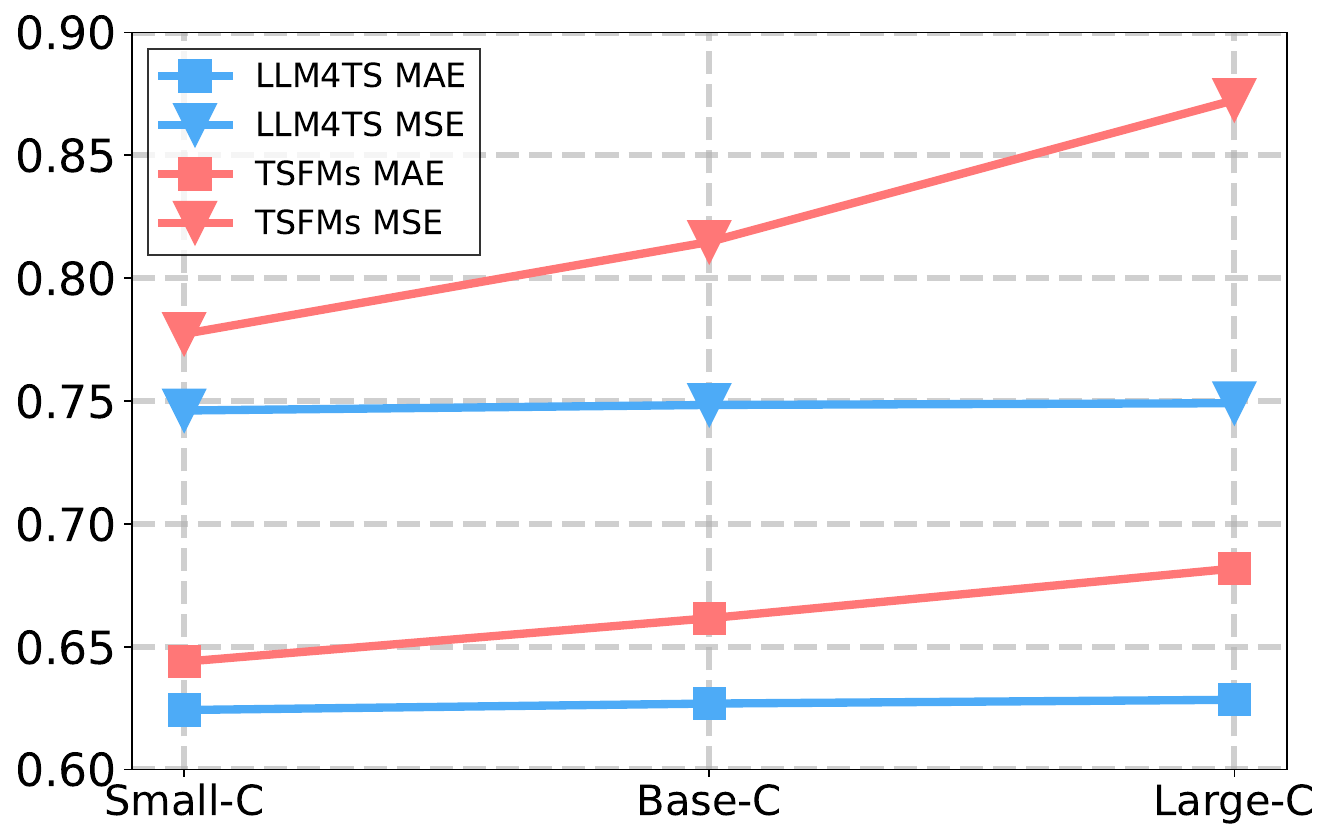}
        \caption{Average errors of three scales of \colorbox[HTML]{4dabf7}{LLM4TS} and \colorbox[HTML]{ff7777}{TSFMs} models on eight out-of-domain datasets.}
        \label{fig:ood_performance}
    \end{minipage}
    \vspace{-0.2cm}
\end{figure}

\vspace{-0.2cm}
\paragraph{Rationale.} Training on a narrow data source, such as a single energy dataset, can cause the model to overfit to specific temporal resolutions and limited semantic patterns \citep{woo2024unified,huang2025exploiting}. Such constraints may hinder models from fully realizing their representational and generalization potential. We adopt mixed-distribution learning strategy to increase data diversity and scale, and analyze their relationship with model scaling.
\vspace{-0.2cm}
\paragraph{Setup.} 
We aggregated the training sets listed in Tab. \ref{tab:Statistics of datasets} and additionally collected 32 datasets, resulting in a total of {\color{cc4}{41 datasets}}, and then randomly shuffled them for {\color{cc5}{cross-dataset learning}}, resulting in six models: {\color{cc1}{LLM4TS}}-({\color{cc2}{Small}}, {\color{cc2}{Base}}, {\color{cc2}{Large}})-C and {\color{cc1}{TSFMs}}-({\color{cc2}{Small}}, {\color{cc2}{Base}}, {\color{cc2}{Large}})-C. Unlike the training phase, during evaluation each model is assessed sequentially on each individual test set. The aggregated dataset encompasses multiple domains with diverse and markedly different distributions, as illustrated in Fig. \ref{fig:Dataset sources}. In this setting, the data are not only more diverse in distribution, sourced from {\color{cc4}{eight major domains}} across {\color{cc4}{41 datasets}}, but also comprise up to {\color{cc3}{6B}} observations. After model evaluation, we obtained in-domain (as shown in Tab. \ref{Performance comparison of LLM4TS and TSFMs under single-dataset and cross-dataset learning}) and out-of-domain (as shown in Fig. \ref{fig:ood_performance}) results, and compared the in-domain results with those presented in Sec. \ref{subsection:Does Backbone Scaling Improve Forecasting Performance? (RQ1)}.

\vspace{-0.2cm}
\paragraph{Results.} 
In Tab. \ref{Performance comparison of LLM4TS and TSFMs under single-dataset and cross-dataset learning}, cross-dataset learning improves the in-domain performance of larger models; however, somewhat surprisingly, the best performance still falls slightly below that of \bluestar. In contrast, scaling paradox in TSFMs remains pronounced, and the smallest model achieves the best results {\redstar}. In addiction, out-of-domain evaluations fail to gain improvements from larger scales, shown in Fig. \ref{fig:ood_performance}.

\begin{table}[h]
\centering
\caption{Performance under single-dataset and cross-dataset learning. The prediction errors are averaged across all datasets. Darker cell colors indicate larger prediction errors. 
\textcolor{green!60}{$\blacktriangle$} denotes increasing errors with larger model, while \textcolor{red!90}{$\blacktriangledown$} indicates decreasing errors.
\bluestar\ marks the best performance within the LLM4TS family, and \redstar\ represents the best performance within the TSFMs family.}
\label{Performance comparison of LLM4TS and TSFMs under single-dataset and cross-dataset learning}
\vspace{-0.2cm}
\resizebox{1\linewidth}{!}{%
\begin{tabular}{c|ccc|ccc|ccc|ccc}
    \toprule
    \textbf{strategy}&\multicolumn{6}{c|}{\cellcolor{gray!20}\textbf{{\color{cc5}{Single-dataset learning }}}} & \multicolumn{6}{c}{\cellcolor{gray!20}\textbf{{\color{cc5}{Cross-dataset learning }}}}\\
    \midrule
     \multirow{2}{*}{\textbf{Models}} & \multicolumn{3}{c|}{\textbf{LLM4TS-}} & \multicolumn{3}{c|}{\textbf{TSFMs-}} & \multicolumn{3}{c|}{\textbf{LLM4TS-}} & \multicolumn{3}{c}{\textbf{TSFMs-}}\\
     & Small & Base & Large & Small & Base & Large & Small-C & Base-C & Large-C & Small-C & Base-C & Large-C\\
    \midrule
    \textbf{MAE} & {\cellcolor{ss1}{0.2875\bluestar}} & {\cellcolor{ss2}{0.3250}} & {\cellcolor{ss3}{0.3283}} & {\cellcolor{rr1}{0.2921}} & {\cellcolor{rr2}{0.3028}} & {\cellcolor{rr3}{0.3078}} & {\cellcolor{ss3}{0.2986}} & {\cellcolor{ss2}{0.2894}} & {\cellcolor{ss1}{0.2875}} & {\cellcolor{rr1}{0.2872\redstar}} & {\cellcolor{rr2}{0.2986}} & {\cellcolor{rr3}{0.3006}} \\
    \textbf{MSE} & {\cellcolor{ss1}{0.2226\bluestar}} & {\cellcolor{ss2}{0.2784}} & {\cellcolor{ss3}{0.2846}} & {\cellcolor{rr1}{0.2286}} & {\cellcolor{rr2}{0.2482}} & {\cellcolor{rr3}{0.2540}} & {\cellcolor{ss3}{0.2367}} & {\cellcolor{ss2}{0.2263}} & {\cellcolor{ss1}{0.2233}} & {\cellcolor{rr1}{0.2257\redstar}} & {\cellcolor{rr2}{0.2496}} & {\cellcolor{rr3}{0.2543}} \\
    \midrule
\textbf{Trend} & \multicolumn{3}{c|}{\textcolor{green!60}{\large$\blacktriangle$}} 
      & \multicolumn{3}{c|}{\textcolor{green!60}{\large$\blacktriangle$}}  
      & \multicolumn{3}{c|}{\textcolor{red!90}{\large$\blacktriangledown$}}  
      & \multicolumn{3}{c|}{\textcolor{green!60}{\large$\blacktriangle$}}  \\

    \bottomrule
\end{tabular}%
}
\end{table}
\vspace{+0.4cm}

\begin{abox} 
    \looseness -1 \textbf{Takeaways 3:} In in-domain tests, larger LLM4TS models leverage LLMs’ multi-domain knowledge under mixed-distribution learning but still underperform smaller models trained on single datasets. For TSFMs, mixed learning improves performance at the same scale, but the scaling paradox persists as the model size increases. In out-of-domain testing, larger-scale models show no advantage.
\end{abox}

\section{A Few Layers Dominate the Majority of Performance}
Although larger models contain more layers, they do not yield better performance. As the previous analyses in Sec.~\ref{section:Scaling Paradox} have shown this phenomenon is not caused by other factors such as data or training strategies. Therefore, we propose possible hypotheses: \emph{"Not all layers contribute effectively to the prediction process (hypothesis 1)"} and \emph{"Retaining a few true executors can rival the full (hypothesis 2)"}. Experimental results confirm the proposed hypothesis and motivate a new approach for identifying and pruning redundant layers.

\subsection{Not All Layers Contribute to the Final Predictions (Hypothesis 1)}
\label{Do All layers of Backbone Contribute to Final Predictions? (RQ4)}

\vspace{-0.3cm}
\paragraph{Inter-layer Representations.} We use Euclidean distance to measure absolute vector differences and reflects overall scaling variations in a geometrical way. Let $\boldsymbol{H}^{l-1}$ and $\boldsymbol{H}^{l}$ denote the input and output of  $l$-th layer. The directional shifts are measured using cosine similarity due to its scale invariance \citep{pmlr-v97-kornblith19a}. The directional shifts are presented below:
\begingroup
\setlength{\abovedisplayskip}{-0.2em}
\setlength{\belowdisplayskip}{-0.6em}
\begin{equation}
Dist^{l} = \|\boldsymbol{H}^{l} - \boldsymbol{H}^{l-1}\|_2, \;\;
Sim(\boldsymbol{H}^{l}, \boldsymbol{H}^{l-1}) = 
\frac{\langle \boldsymbol{H}^{l}, \boldsymbol{H}^{l-1} \rangle}
{\|\boldsymbol{H}^{l}\|_2 \, \|\boldsymbol{H}^{l-1}\|_2}.
\end{equation}
\endgroup

\begin{figure}[h]
\vspace{-0.2cm}
    \centering
    \begin{subfigure}{0.49\linewidth}
        \includegraphics[width=\linewidth]{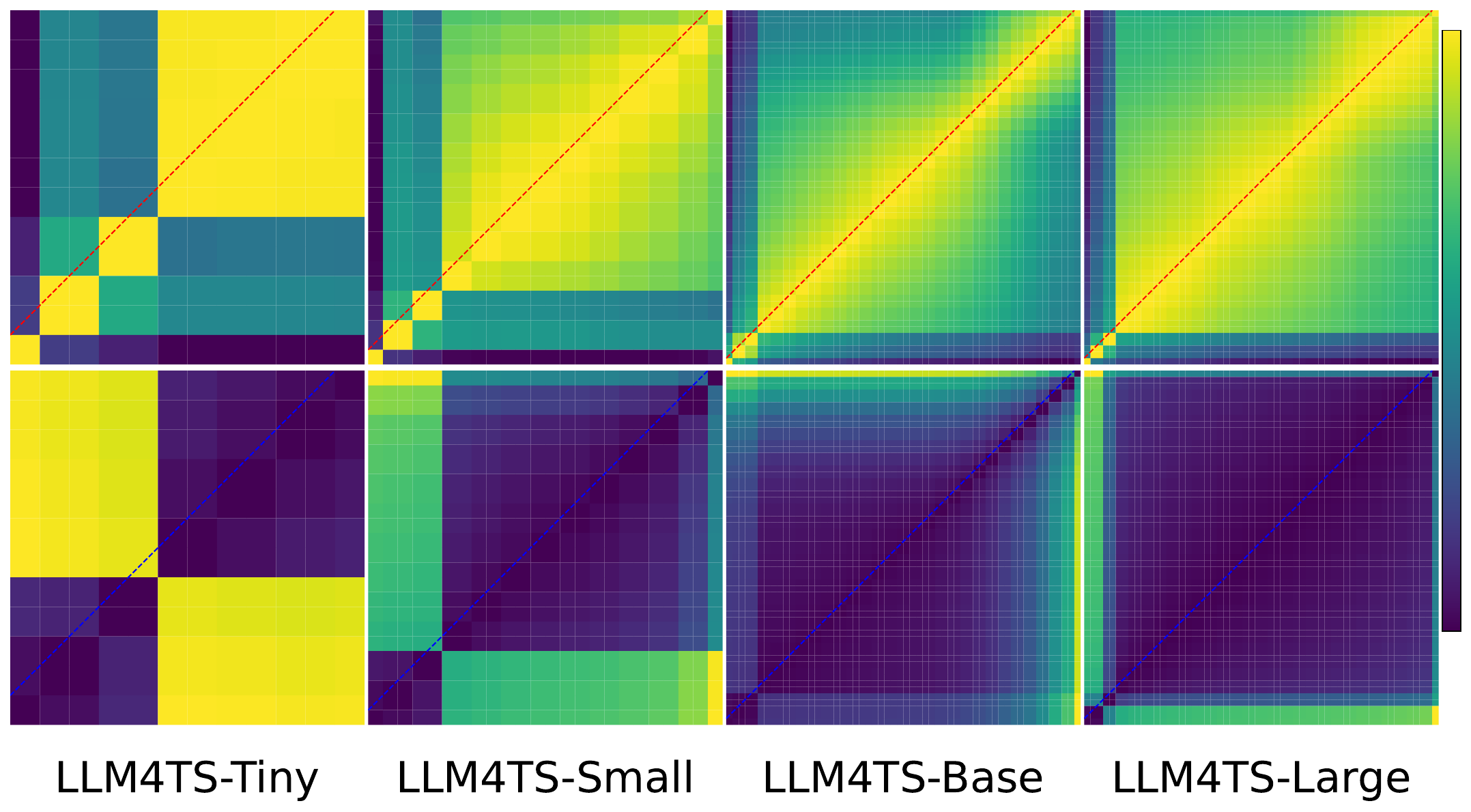}
    \end{subfigure}
    \hspace{0.001\linewidth}
    \begin{subfigure}{0.49\linewidth}
        \includegraphics[width=\linewidth]{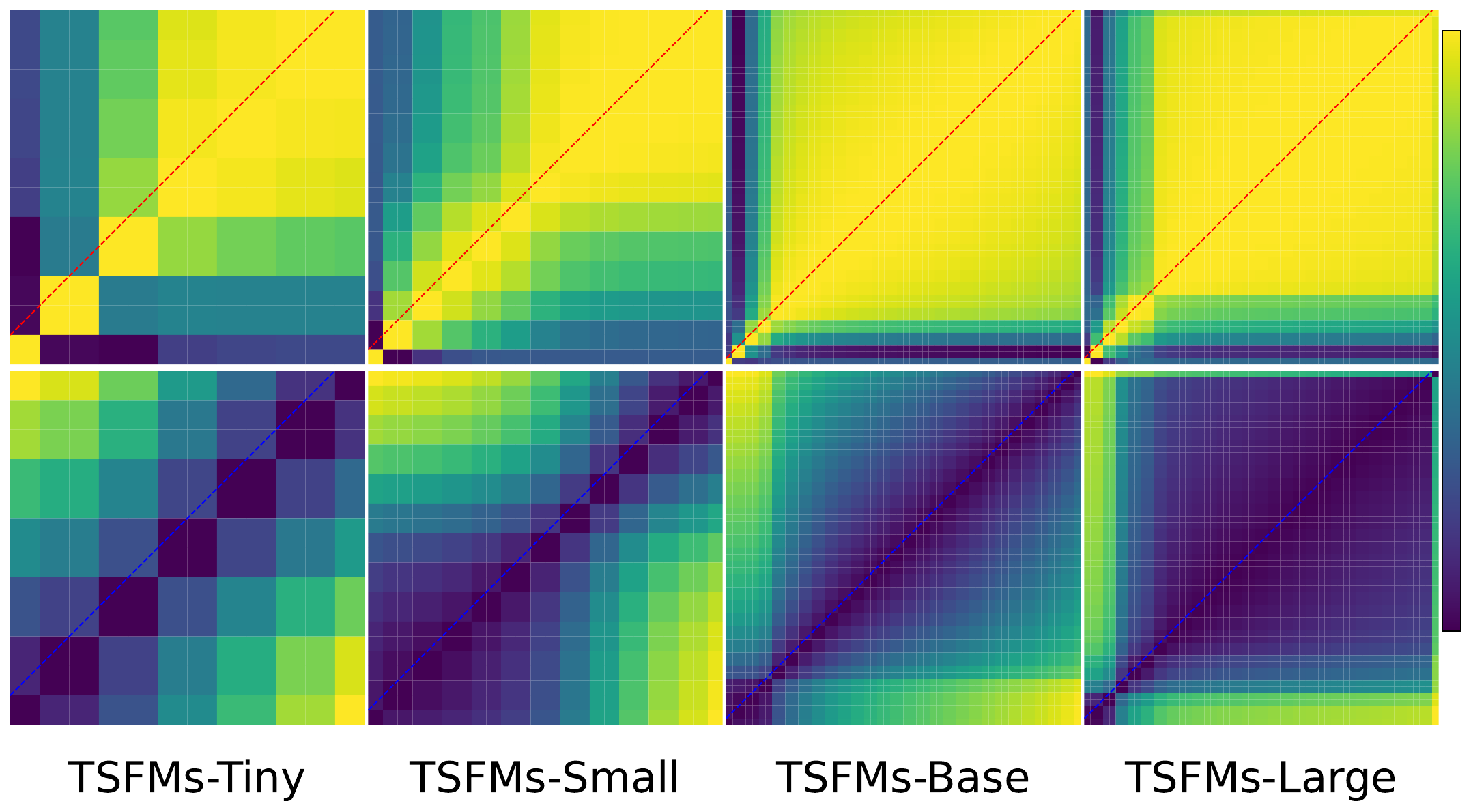}
    \end{subfigure}
    \caption{Inter-layer cosine similarity (\textcolor{red}{row 1}) and Euclidean distance (\textcolor{blue}{row 2}). Brighter areas indicate higher values, darker areas lower values. It can be observed that across eight models of different scales and architectures, inter-layer representations exhibit high similarity and low Euclidean distance.}
    \label{fig:Inter-layer cosine similarity and Euclidean distance}
\end{figure}

\vspace{-0.3cm}
\paragraph{Intra-layer Representations.}  
Each attention head can be regarded as an independent relational learner \citep{liu2021attention}, parameterized by distinct projection matrices. Given the embeddings $\boldsymbol{X}^l \in \mathbb{R}^{N \times d_{\text{model}}}$ at layer $l$, the $i$-th head generates its attention weights:
\begingroup
\begin{equation}
\boldsymbol{A}_i^l = 
\operatorname{softmax}\!\left(\boldsymbol{Q}_i^l {\boldsymbol{K}_i^l}^\top / \sqrt{d}\right), 
\quad 
\boldsymbol{Q}_i^l = \boldsymbol{X}^l \boldsymbol{W}_i^Q,\;
\boldsymbol{K}_i^l = \boldsymbol{X}^l \boldsymbol{W}_i^K .
\end{equation}
\endgroup
$\boldsymbol{W}_i^Q, \boldsymbol{W}_i^K \in \mathbb{R}^{d_{\text{model}} \times d}$ are head-specific projections and $d = d_{\text{model}}/H$ denotes the per-head dimensionality. We measure the average pairwise similarity across all head attentions in layer $l$. A lower $\bar{s}^{(l)}$ indicates higher functional diversity among heads, reflecting that each head captures distinct aspects of the TS representations. $\bar{s}^{l}$ be defined as:
\begingroup
\setlength{\belowdisplayskip}{-0.5em}
\begin{equation}
\bar{s}^{l} \;=\; \frac{2}{H(H-1)} \sum_{1 \leq i < j \leq H} 
      \mathrm{sim}\!\left(\boldsymbol{A}_i^l, \boldsymbol{A}_j^l\right).
\label{equation:average pairwise similarity across all head attentions}
\end{equation}
\endgroup
\vspace{-0.3cm}
\paragraph{Evidence 1.} By analyzing the representations of the eight trained models during the inference process, we find that only a few layers contribute substantial representation shifts, while many layers function redundantly as shown in Fig. \ref{fig:Inter-layer cosine similarity and Euclidean distance}. This layer-wise redundancy phenomenon is also observed in cross-dataset learning. Our experiments further demonstrate that even with increased training data and greater distribution diversity, larger models still exhibit substantial redundancy across most layers, which helps explain why scaling up model size does not always lead to improved performance.

\begin{figure}[H]
\vspace{-0.3cm}
    \centering
    \includegraphics[width=\linewidth]{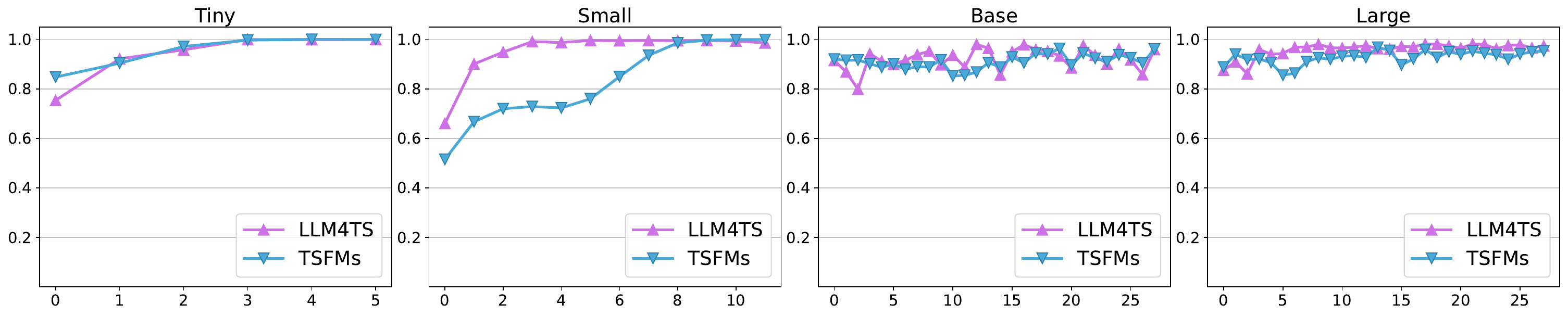}
        \vspace{-.5cm}
    \caption{Average inter-layer pairwise similarity across all head attentions.}
    \vspace{-.5cm}
    \label{fig:Inter-layer average pairwise similarity across all head attentions}
\end{figure}

\vspace{-.5cm}
\paragraph{Evidence 2.}
Eight model from two families exhibit similar trends in intra-layer representations, as shown in Fig. \ref{fig:Inter-layer average pairwise similarity across all head attentions}. For models with smaller backbones (6 or 12 layers), heads in the shallow layers primarily allocate their attention to distinct semantic patterns. However, in the middle and deeper layers, $\bar{s}^{l}$ approaches 1, with most heads learning within the similar subspace. For models with larger backbones (28 layers), heads across all layers almost exhibit high similarity.
\begin{abox} 
    \looseness -1 \textbf{Conclusion:} For LLM4TS and TSFMs, only a small fraction of layers actively contribute to representation learning, and the vast majority remain largely passive spectators.
\end{abox}

\subsection{Retaining a Few True Executors Can Rival the Full (Hypothesis 2)}
\label{subsection:Can a Few Layers Match Full-Model Forecasting Performance? (RQ5)}
\begin{figure}[t]
    \centering
    \vspace{+1.0cm}
    \begin{minipage}[t]{0.47\linewidth}
    \vspace{-1.2cm}
        \centering
        \includegraphics[width=\linewidth]{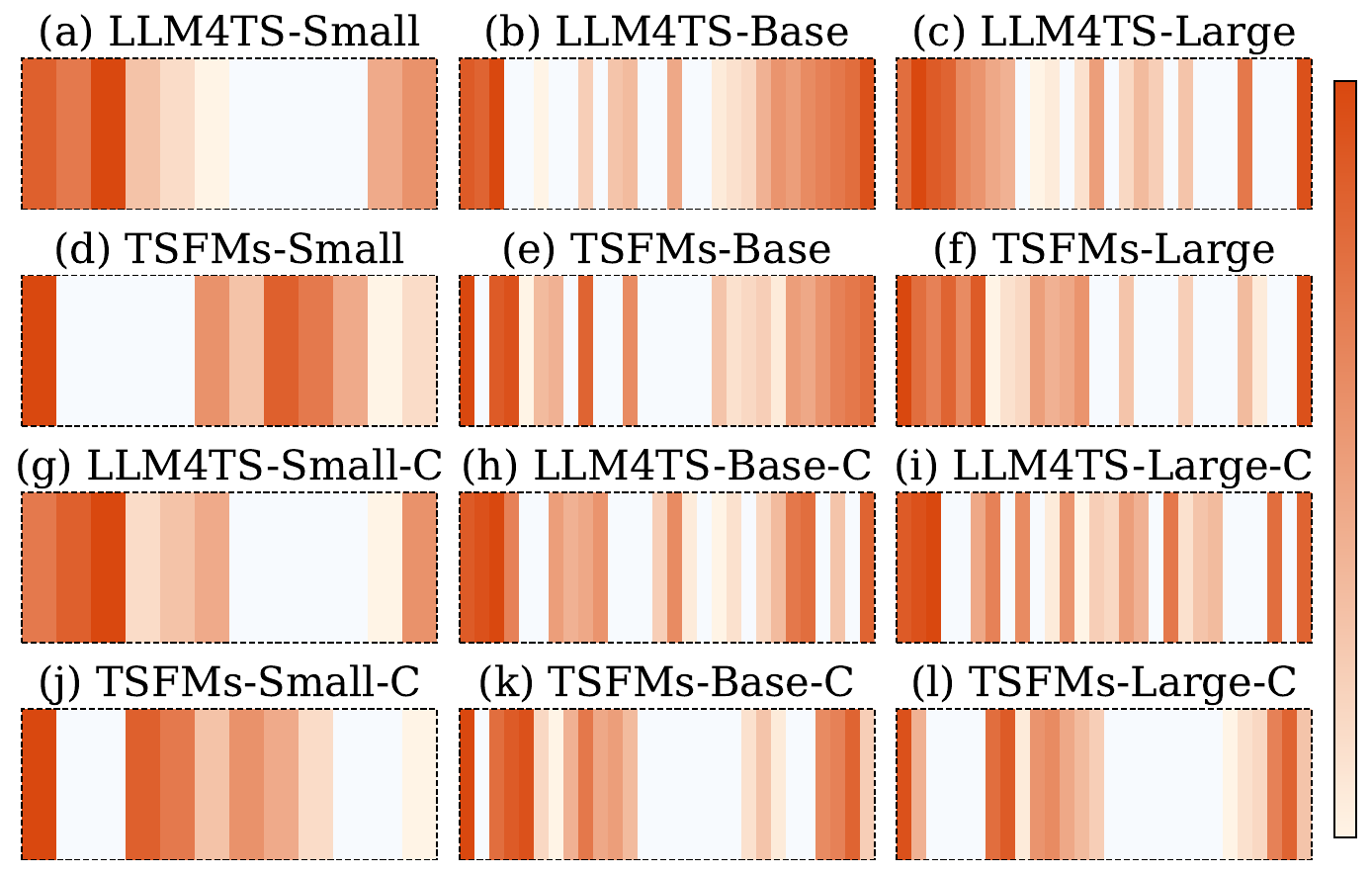}
        \vspace{-0.5cm}
        \captionsetup{singlelinecheck=true}
        \caption{Average layer-wise importance, and darker colors indicate greater importance. The key layers are located in the initial and final portions.}
        \label{fig:Average layer-wise importance}
    \end{minipage}
    \hspace{0.005\linewidth}
    \begin{minipage}[t]{0.44\linewidth}
    \vspace{-1.6cm}
        \centering
        \renewcommand{\arraystretch}{1.2} 
        \vspace{-0.2cm}
        \begin{table}[H]
        \caption{Critical layer identification}
        \label{tab:Critical layer identification}
\resizebox{\linewidth}{!}{

\begin{tabular}{l}
\hline
\textbf{Algorithm: Critical Layer Identification} \\
\hline
\textbf{Input:} Representations $\boldsymbol{H}^{0}, \dots, \boldsymbol{H}^{L-1}$, attention weights $\boldsymbol{A}_i^l$, \\
\quad decay factor $\alpha$, top percentile $\tau$, number of preceding layers $K$ \\
\textbf{Output:} Ordered list of critical layers \\[1mm]

\textbf{for} $l = 0$ \textbf{to} $L-1$ \textbf{do} \\
\quad $Dist^{l} \gets \| \boldsymbol{H}^{l} - \boldsymbol{H}^{l-1} \|_2$, $ \gets Sim(\boldsymbol{H}^{l}, \boldsymbol{H}^{l-1})$ \\[1mm]
\quad $\bar{s}^{l} \gets \tfrac{1}{\binom{H}{2}} \sum_{i<j} Sim(\boldsymbol{A}_i^l, \boldsymbol{A}_j^l)$ \\[1mm]
\quad $w_k \gets \alpha^k / \sum_{i=1}^{K} \alpha^i$, \quad 
      $R^{l} \gets \sum_{k=1}^{K} w_k \cdot Sim(\boldsymbol{H}^{l}, \boldsymbol{H}^{l-k})$ \\[1mm]
\quad $I^{l} \gets (1 - R^{l}) \cdot (1-\bar{s}^{l})$ \\
\textbf{end for} \\[1mm]
\textbf{Let} $\mathcal{S}$ be the indices of Top-$\tau\%$ layers ranked by $Dist^{l}$. \\
\textbf{for} $l = 1$ \textbf{to} $L-2$ \textbf{do} \\
\quad \textbf{if} $l \notin \mathcal{S}$ \textbf{then} $I^{l} \gets 0$ \\
\textbf{end for} \\[1mm]
\textbf{Rank} all layers by $I^{l}$ in descending order: \\
\hline
\end{tabular}
}
\end{table}
    \end{minipage}
    \vspace{-0.2cm}
\end{figure}

\paragraph{Rationale.} Building on the above, we fine only a small subset making the primary contribution. Then, if only a few seemingly more influential layers are retained while pruning the remaining “less important” ones, would the model still exhibit comparable performance? This motivates the need for a method to accurately assess each layer’s contribution and identify critical layers. 
\vspace{-0.2cm}
\paragraph{Critical Layer Identification.} In progressive networks, each layer builds upon its predecessors, refining them toward higher-level, task-relevant features. Consequently, the contribution of early layers gradually decreases, highlighting a natural hierarchy in feature abstraction. Therefore, static properties of an individual layer may not fully reflect its contribution to last predictions. Considering the decaying influence of preceding layers on deeper layers, importance score of the $l$-th layer is: 
\begin{equation}
I^{l} = \mathbf{1}\{l \in \text{Top-}\tau\%\} \cdot \left( 1 - R^{l} \right) \cdot \left( 1-\bar{s}^{l} \right),
R^{l} = \sum_{k=1}^{K} w_k \cdot Sim\!\big(\boldsymbol{H}^{l}, \boldsymbol{H}^{l-k}\big).
\end{equation}
$\alpha \in (0,1)$ is a decay factor, $w_k = \alpha^k/{\sum_{i=1}^{K} \alpha^i}$ the normalized decaying weight, $\bar{s}^{l}$ is defined in formula \ref{equation:average pairwise similarity across all head attentions} and $K$ is the number of preceding layers. We use the multiplicative form $(1 - R^{l})\cdot(1 - \bar{s}^{l})$ instead of a weighted sum. 
This ensures that high redundancy or instability suppresses a layer's importance and introduces a nonlinear gating effect, so only layers that are both distinctive and stable are highly weighted. 
A weighted sum could let one factor offset the other, hiding their joint effect in hierarchical feature abstraction.
$\tau\%$ specifies the top percentile of layers ranked by $Dist^{l}$. By ranking all layers according to $I^{l}$ in descending order, those contributing most can be identified. The first and last layers, connecting the time series and embeddings, are exempt from selection. 
The remaining layers are retained if they satisfy either of the two criteria: being within the top $\alpha\%$ or having a cumulative score proportion exceeding $\beta\%$, with the latter given higher priority. Details are summarized in Tab. \ref{tab:Critical layer identification}.

\vspace{-0.2cm}
\paragraph{Method Evaluation.} We conduct extensive experiments across 12 models and 9 data distributions, and visualize the average distribution of layer importance over the validation sets of different datasets, as shown in Fig. \ref{fig:Average layer-wise importance}. The three hyperparameters $\tau\%$, $\alpha\%$, and $\beta\%$ are set to 80\%, 50\%, and 85\%, respectively. Based on the importance ranking, we prune redundant layers and fine-tune the pruned model on the training sets to realign it with the data distribution. We comprehensively evaluate the pruned and original models in both in-domain and out-of-domain forecasting scenarios, considering prediction errors and inference efficiency.
\vspace{-0.2cm}
\paragraph{Results.}  MAE and MSE are reported as the averages over all datasets. Efficiency is quantified as the ratio between the inference time of the original model and that of the pruned model (defined as $ {T_{\text{original model}}}/{T_{\text{pruned model}}}$, and $T$ is inference time). All experiments are conducted under identical settings to ensure fairness. Main results are reported in Tab. \ref{Prediction performance and inference efficiency under pruning}. Results show that the pruned models (minimum retained proportion $\approx \textbf{20\%}$.) achieve an average of 2× (up to \textbf{2.7×}) inference speedup while almost universally outperforming the original models. The maximum performance gain reaches \textbf{12\%}, whereas the largest loss is only \textbf{0.6\%}, which is negligible.

\begin{table}[htbp]
\centering
\vspace{+0.2cm}
\captionsetup{justification=justified,singlelinecheck=false}
\caption{Comparison of performance and inference efficiency between the original and pruned models. “ID” and “OOD” refer to in-domain and out-of-domain, respectively. OOD evaluation is not applicable for single-dataset learning, denoted as “–”. \colorbox[HTML]{FFE5C6}{\strut Performance gain} and \colorbox[HTML]{FF8787}{\strut Performance loss} denote improved and degraded metrics.
}
\vspace{-0.25cm}
\label{Prediction performance and inference efficiency under pruning}
\renewcommand{\arraystretch}{1.2}
\resizebox{1\linewidth}{!}{
\begin{tabular}{c|cc|cc|cc|cc|c|c}
\toprule
\multirow{2}{*}{\textbf{Model}} & \multicolumn{2}{c|}{\textbf{MAE-ID}} & \multicolumn{2}{c|}{\textbf{MSE-ID}} & \multicolumn{2}{c|}{\textbf{MAE-OOD}}& \multicolumn{2}{c|}{\textbf{MSE-OOD}} & \multirow{2}{*}{\textbf{Efficiency$\uparrow$}} & \multirow{1}{*}{\textbf{Critical layer ratio}}\\
& Pruned & Original & Pruned & Original & Pruned & Original & Pruned & Original &  &${L_{\text{pruned model}}}/{L_{\text{original model}}}$\\
\midrule
\multicolumn{11}{c}{
  \rule{0pt}{2.2ex}
  \cellcolor{gray!20}\raisebox{0.3ex}{\textbf{{\color{cc5}{Single-dataset learning}}}}
  \rule[-0.8ex]{0pt}{2.2ex}
} \\
LLM4TS-Small  & {\cellcolor{high}{0.283}} & {\cellcolor{high}{0.288}} & {\cellcolor{high}{0.218}} & {\cellcolor{high}{0.223}} & {\cellcolor{blank}{--}}  & {\cellcolor{blank}{--}}  & {\cellcolor{blank}{--}}  & {\cellcolor{blank}{--}}  & \textbf{2.7} & 28\%\\
LLM4TS-Base   & {\cellcolor{high}{0.293}} & {\cellcolor{high}{0.325}} & {\cellcolor{high}{0.226}} & {\cellcolor{high}{0.278}} & {\cellcolor{blank}{--}}  & {\cellcolor{blank}{--}}  & {\cellcolor{blank}{--}}  & {\cellcolor{blank}{--}}  & 2.2 & 23\%\\
LLM4TS-Large  & {\cellcolor{high}{0.283}} & {\cellcolor{high}{0.328}} & {\cellcolor{high}{0.219}} & {\cellcolor{high}{0.285}} & {\cellcolor{blank}{--}}  & {\cellcolor{blank}{--}}  & {\cellcolor{blank}{--}}  & {\cellcolor{blank}{--}}  & 2.4 & \textbf{21\%} \\
\midrule
TSFMs-Small   & {\cellcolor{high}{0.287}} & {\cellcolor{high}{0.292}} & {\cellcolor{high}{0.221}} & {\cellcolor{high}{0.227}} & {\cellcolor{blank}{--}}  & {\cellcolor{blank}{--}}  & {\cellcolor{blank}{--}} & {\cellcolor{blank}{--}}  & 2.5 &24\%\\
TSFMs-Base    & {\cellcolor{high}{0.284}} & {\cellcolor{high}{0.303}} & {\cellcolor{high}{0.220}} & {\cellcolor{high}{0.248}} & {\cellcolor{blank}{--}}  & {\cellcolor{blank}{--}}  & {\cellcolor{blank}{--}}  & {\cellcolor{blank}{--}}  & \textbf{2.7} &\textbf{21\%} \\
TSFMs-Large   & {\cellcolor{high}{0.290}} & {\cellcolor{high}{0.308}} & {\cellcolor{high}{0.231}} & {\cellcolor{high}{0.254}} & {\cellcolor{blank}{--}}  & {\cellcolor{blank}{--}}  & {\cellcolor{blank}{--}}  & {\cellcolor{blank}{--}}  & \textbf{2.7} &23\%\\
\midrule
\multicolumn{11}{c}{
  \rule{0pt}{2.2ex}
  \cellcolor{gray!20}\raisebox{0.3ex}{\textbf{{\color{cc5}{Cross-dataset learning}}}}
  \rule[-0.8ex]{0pt}{2.2ex}
} \\
LLM4TS-Small-C  & {\cellcolor{high}{0.298}} & {\cellcolor{high}{0.299}} & {\cellcolor{high}{0.233}} & {\cellcolor{high}{0.237}} & {\cellcolor{lower}{0.625}}  & {\cellcolor{lower}{0.624}}  & {\cellcolor{high}{0.740}}  & {\cellcolor{high}{0.746}}  & 2.2 & 33\% \\
LLM4TS-Base-C   & {\cellcolor{high}{0.288}} & {\cellcolor{high}{0.289}} & {\cellcolor{high}{0.225}} & {\cellcolor{high}{0.226}} & {\cellcolor{high}{0.618}}  & {\cellcolor{high}{0.627}}  & {\cellcolor{high}{0.738}}  & {\cellcolor{high}{0.748}}  & 2.1 & 36\%\\
LLM4TS-Large-C  & {\cellcolor{high}{0.286}} & {\cellcolor{high}{0.288}} & {\cellcolor{lower}{0.226}} & {\cellcolor{lower}{0.223}} & {\cellcolor{high}{0.620}}  & {\cellcolor{high}{0.628}}  & {\cellcolor{high}{0.727}}  & {\cellcolor{high}{0.749}}  & 2.2 & 25\%\\
\midrule
TSFMs-Small-C   & {\cellcolor{high}{0.286}} & {\cellcolor{high}{0.287}} & {\cellcolor{lower}{0.227}} & {\cellcolor{lower}{0.226}} & {\cellcolor{high}{0.635}}  & {\cellcolor{high}{0.644}}  & {\cellcolor{lower}{0.782}}  & {\cellcolor{lower}{0.777}}  & 1.9 & 33\%\\
TSFMs-Base-C    & {\cellcolor{high}{0.293}} & {\cellcolor{high}{0.299}} & {\cellcolor{high}{0.247}} & {\cellcolor{high}{0.250}} & {\cellcolor{high}{0.654}}  & {\cellcolor{high}{0.662}}  & {\cellcolor{high}{0.773}}  & {\cellcolor{high}{0.815}}  & 2.5 & 29\%\\
TSFMs-Large-C   & {\cellcolor{high}{0.282}} & {\cellcolor{high}{0.301}} & {\cellcolor{high}{0.224}} & {\cellcolor{high}{0.254}} & {\cellcolor{high}{0.659}}  & {\cellcolor{high}{0.682}}  & {\cellcolor{high}{0.821}}  & {\cellcolor{high}{0.872}}  & \textbf{2.7} & 25\%\\
\bottomrule
\end{tabular}
}

\end{table}
\begin{abox} 
    \looseness -1 \textbf{Conclusion:} Retaining only the critical layers, followed by fine-tuning, can preserve or even improve forecasting accuracy, while substantially inference latency.
\end{abox}

\section{Further Investigation of Our Method}
\label{section:Experiments and Analysis}
In the previous section, our method demonstrated strong effectiveness when applied to model families. In Sec. \ref{subsection:Method Adaptation for Baselines}, we extend this method to other model architectures to validate its transferability. Sec. \ref{Full vs. Pruned} investigates 3 configurations, keeping all layers, keeping only a few important layers, and randomly keep same number of layers, to assess their impact on performance. Finally, Sec. \ref{subsec:Inference Overhead and Performance} conducts an ablation study.

\subsection{Method Scalability}
\label{subsection:Method Adaptation for Baselines}
\paragraph{Models and Datasets.} Besides the datasets listed in Tab. \ref{tab:Statistics of datasets}, four PEMS subsets are included for broader coverage, with the splits following \cite{ICLR2024_2ea18fdc}. We extend the method to four LLM4TS models: \textsc{Fsca}\citep{ICLR2025_e1de63ec}, \textsc{Calf}\citep{liu2025calf}, \textsc{Time-LLM(G)} (GPT-2 as backbone)\citep{jin2023time}, and \textsc{Ofa}\citep{zhou2023one}, and four TSFMs models: \textsc{Sundial}$_{\text{Large}}$\citep{liu2025sundialfamilyhighlycapable}, \textsc{Chronos}$_{\text{Base}}$\citep{ansari2024chronos}, \textsc{Moirai}$_{\text{Large}}$\citep{moirai} and \textsc{TimesFM}\citep{das2024decoder}. With considerable heterogeneity in architecture, scale, temporal encoding method and paradigm, these models provide ideal conditions for applying and assessing the transferability of our method, as shown in Tab. \ref{tab:model info}. To preserve the full capabilities of LLM in LLM4TS, we start from the complete GPT-2, rather than coarsely truncating the first few layers as done in \textsc{Fsca}, \textsc{Calf}, and \textsc{Ofa}.

\begin{table}[h]
  \centering
  \vspace{-0.2cm}
  \caption{Based on these models, we perform layer identification and pruning across a total of 13 data distributions.}
  \label{tab:model info}
  \renewcommand{\arraystretch}{1.2} 
  \vspace{-0.2cm}
  \makebox[\textwidth][c]{%
    \resizebox{1\linewidth}{!}{
      \begin{tabular}{c|ccccccccc}
        \toprule
        {} & \textbf{Model} & \textbf{Backbone} & \textbf{Paradigm} & \textbf{Token Level} & \textbf{Channels} & \textbf{Prompt} & \textbf{Alignment} & \textbf{Input Length} & \textbf{Horizons}\\
        \midrule
        \multirow{4}{*}{\rotatebox{90}{\textbf{LLM4TS}}}
          & FSCA & GPT-2 & decoder-only & Patch & 768 & {\color{green}{\ding{52}}} & sentence & 512 &$\{96,192,336,720\}$\\
          & CALF & GPT-2 & decoder-only & Variable & 768 & {\color{red}{\ding{56}}} & word& 96 &$\{96,192,336,720\}$ \\
          & Time-LLM(G) & GPT-2 & decoder-only & Patch & 768 & {\color{green}{\ding{52}}} & word  & 512 &$\{96,192,336,720\}$\\
          & OFA & GPT-2 & decoder-only & Patch & 768 & {\color{red}{\ding{56}}} & latent space & 336 &$\{96,192,336,720\}$\\
        \toprule
        {} & \textbf{Model} & \textbf{Backbone} & \textbf{Paradigm} & \textbf{Token Level} & \textbf{Channels} & \textbf{Pre-training Scale} & \textbf{Model Size} & \textbf{Input Length} & \textbf{Horizons}\\
        \midrule
        \multirow{4}{*}{\rotatebox{90}{\textbf{TSFMs}}}
          & {Sundial}$_{\text{Large}}$ & $Layers:24$ & decoder-only & Patch & 1,024 & 1,000B & 450M & 336 &$\{96,192,336,720\}$\\
          & {Chronos}$_{\text{Base}}$ & $Layers:24$ & encoder-decoder & Point & 1,024 & 200B & 200M & 96 &$\{96,192,336,720\}$\\
          &\textsc{Moirai}$_{\text{Large}}$ & $Layers:24$ & encoder-only & Patch& 1,024 & 230B & 300M & 96 &$\{96,192,336,720\}$\\
          & TimesFM & $Layers:50$ & decoder-only & Patch & 1,280 & 100B & 500M & 96 &$\{96,192,336,720\}$\\
        \bottomrule
      \end{tabular}
    }
  }
\end{table}
\vspace{-0.2cm}
\paragraph{Implementation Details.} We begin by examining the models’ inter-layer and intra-layer representations on the validation sets. Following the procedure introduced earlier, we compute cosine similarity across layers, and pairwise similarity of attention weights among heads, to derive a score for each layer. Additionally, excluding the first and last layers, we calculate the Euclidean distance to measure transformation magnitude. Layers whose distance differences do not rank within the top 80\% are assigned zero scores and excluded from the importance ranking. Based on this, we obtain a hierarchical ranking and retain layers meeting either of the two criteria: top 50\% or cumulative score proportion exceeding 90\%, with the latter given higher priority. Finally, the pruned models are fine-tuned using the original training strategy to realign the models with the data distribution. Note: The four TSFMs are fine-tuned on downstream datasets to strengthen generalization and ensure comparable parameter distributions before and after pruning \citep{zhao2025less}. 

\begin{table}[h]
  \centering
\captionsetup{justification=justified,singlelinecheck=false}
\caption{Comparison of pre- and post-pruning metrics in four LLM4TS. Results are averaged across horizons $\in \{96,192,336,720\}$. Original: model from original paper. \colorbox[HTML]{FFE5C6}{\strut Performance gain} and \colorbox[HTML]{FF8787}{\strut Performance loss} denote improved and degraded metrics.} 
\label{Method adaptation for baselines}
\begin{threeparttable}
\vspace{-0.2cm}
\renewcommand{\arraystretch}{1.4}
  \resizebox{\columnwidth}{!}{\small 
\begin{tabular}{c|cccc|cccc|cccc|cccc}
\toprule
\multirow{3}{*}{\textbf{Models}} & \multicolumn{16}{c}{\cellcolor{gray!40}\textbf{LLM4TS}} \\
 &\multicolumn{4}{c|}{\textbf{FSCA \citeyearpar{ICLR2025_e1de63ec}}} & \multicolumn{4}{c|}{\textbf{CALF \citeyearpar{liu2025calf}}}& \multicolumn{4}{c|}{\textbf{Time-LLM(G) \citeyearpar{jin2023time}}}& \multicolumn{4}{c}{\textbf{OFA \citeyearpar{zhou2023one}}} \\
& \multicolumn{2}{c}{\textbf{Pruned}} & \multicolumn{2}{c|}{\textbf{Original}}& \multicolumn{2}{c}{\textbf{Pruned}} & \multicolumn{2}{c|}{\textbf{Original}}& \multicolumn{2}{c}{\textbf{Pruned}} & \multicolumn{2}{c|}{\textbf{Original}}& \multicolumn{2}{c}{\textbf{Pruned}} & \multicolumn{2}{c}{\textbf{Original}}\\
\midrule
{\textbf{Metric}} &{\textbf{MAE}} &{\textbf{MSE}} &{\textbf{MAE}} &{\textbf{MSE}} &{\textbf{MAE}} &{\textbf{MSE}} &{\textbf{MAE}} &{\textbf{MSE}}&{\textbf{MAE}} &{\textbf{MSE}} &{\textbf{MAE}} &{\textbf{MSE}} &{\textbf{MAE}} &{\textbf{MSE}} &{\textbf{MAE}} &{\textbf{MSE}}\\
\midrule
{\textbf{ETTh1}}
& {\cellcolor{high}{0.443}} & {\cellcolor{high}{0.426}} & {\cellcolor{high}{0.444}} & {\cellcolor{high}{0.430}} & {\cellcolor{high}{0.431}} & {\cellcolor{high}{0.436}} & {\cellcolor{high}{0.434}} & {\cellcolor{high}{0.446}} & {\cellcolor{high}{0.446}} & {\cellcolor{high}{0.431}} & {\cellcolor{high}{0.459}} & {\cellcolor{high}{0.448}} & {\cellcolor{lower}{0.435}} & {\cellcolor{high}{0.429}}  & {\cellcolor{lower}{0.434}} & {\cellcolor{high}{0.430}} \\
\midrule
{\textbf{ETTh2}}
& {\cellcolor{blank}{0.390}} & {\cellcolor{lower}{0.349}} & {\cellcolor{blank}{0.390}} & {\cellcolor{lower}{0.348}} & {\cellcolor{high}{0.391}} & {\cellcolor{high}{0.362}} & {\cellcolor{high}{0.395}} & {\cellcolor{high}{0.373}} & {\cellcolor{high}{0.408}} & {\cellcolor{high}{0.368}} & {\cellcolor{high}{0.410}} & {\cellcolor{high}{0.370}} & {\cellcolor{blank}{0.403}} & {\cellcolor{blank}{0.366}}  & {\cellcolor{blank}{0.403}} & {\cellcolor{blank}{0.366}} \\
\midrule
{\textbf{ETTm1}}
& {\cellcolor{high}{0.385}} & {\cellcolor{high}{0.350}} & {\cellcolor{high}{0.387}} & {\cellcolor{high}{0.352}} & {\cellcolor{high}{0.384}} & {\cellcolor{high}{0.387}} & {\cellcolor{high}{0.387}} & {\cellcolor{high}{0.391}} & {\cellcolor{lower}{0.390}} & {\cellcolor{high}{0.358}} & {\cellcolor{lower}{0.389}} & {\cellcolor{high}{0.359}} & {\cellcolor{lower}{0.388}} & {\cellcolor{lower}{0.357}}  & {\cellcolor{lower}{0.385}} & {\cellcolor{lower}{0.355}} \\
\midrule
{\textbf{ETTm2}}
& {\cellcolor{high}{0.319}} & {\cellcolor{high}{0.258}} & {\cellcolor{high}{0.320}} & {\cellcolor{high}{0.259}} & {\cellcolor{lower}{0.320}} & {\cellcolor{high}{0.268}} & {\cellcolor{lower}{0.318}} & {\cellcolor{high}{0.277}} & {\cellcolor{high}{0.327}} & {\cellcolor{high}{0.268}} & {\cellcolor{high}{0.330}} & {\cellcolor{high}{0.274}} & {\cellcolor{blank}{0.325}} & {\cellcolor{blank}{0.265}}  & {\cellcolor{blank}{0.325}} & {\cellcolor{blank}{0.265}} \\
\midrule
{\textbf{ECL}}
& {\cellcolor{high}{0.266}} & {\cellcolor{high}{0.165}} & {\cellcolor{high}{0.267}} & {\cellcolor{high}{0.166}} & {\cellcolor{high}{0.269}} & {\cellcolor{high}{0.184}} & {\cellcolor{high}{0.272}} & {\cellcolor{high}{0.189}} & {\cellcolor{high}{0.263}} & {\cellcolor{high}{0.163}} & {\cellcolor{high}{0.271}} & {\cellcolor{high}{0.169}} & {\cellcolor{high}{0.259}} & {\cellcolor{high}{0.166}}  & {\cellcolor{high}{0.265}} & {\cellcolor{high}{0.169}} \\
\midrule
{\textbf{Weather}}
& {\cellcolor{lower}{0.269}} & {\cellcolor{lower}{0.233}} & {\cellcolor{lower}{0.268}} & {\cellcolor{lower}{0.229}} & {\cellcolor{high}{0.271}} & {\cellcolor{high}{0.255}} & {\cellcolor{high}{0.277}} & {\cellcolor{high}{0.259}} & {\cellcolor{blank}{0.266}} & {\cellcolor{blank}{0.228}} & {\cellcolor{blank}{0.266}} & {\cellcolor{blank}{0.228}} & {\cellcolor{high}{0.257}} & {\cellcolor{high}{0.230}}  & {\cellcolor{high}{0.268}} & {\cellcolor{high}{0.232}} \\
\midrule
{\textbf{Exchange}}
& {\cellcolor{high}{0.434}} & {\cellcolor{high}{0.436}} & {\cellcolor{high}{0.449}} & {\cellcolor{high}{0.450}} & {\cellcolor{lower}{0.413}} & {\cellcolor{lower}{0.378}} & {\cellcolor{lower}{0.408}} & {\cellcolor{lower}{0.370}} & {\cellcolor{blank}{0.448}} & {\cellcolor{high}{0.436}} & {\cellcolor{blank}{0.448}} & {\cellcolor{high}{0.441}} & {\cellcolor{high}{0.408}} & {\cellcolor{high}{0.384}}  & {\cellcolor{high}{0.428}} & {\cellcolor{high}{0.410}} \\
\midrule
{\textbf{Solar}}
& {\cellcolor{high}{0.264}} & {\cellcolor{high}{0.198}} & {\cellcolor{high}{0.274}} & {\cellcolor{high}{0.210}} & {\cellcolor{high}{0.260}} & {\cellcolor{high}{0.227}} & {\cellcolor{high}{0.268}} & {\cellcolor{high}{0.251}} & {\cellcolor{blank}{0.257}} & {\cellcolor{lower}{0.190}} & {\cellcolor{blank}{0.257}} & {\cellcolor{lower}{0.189}} & {\cellcolor{high}{0.264}} & {\cellcolor{high}{0.211}}  & {\cellcolor{high}{0.276}} & {\cellcolor{high}{0.216}} \\
\midrule
{\textbf{Traffic}}
& {\cellcolor{high}{0.277}} & {\cellcolor{high}{0.390}} & {\cellcolor{high}{0.284}} & {\cellcolor{high}{0.397}} & {\cellcolor{high}{0.284}} & {\cellcolor{high}{0.424}} & {\cellcolor{high}{0.286}} & {\cellcolor{high}{0.465}} & {\cellcolor{high}{0.279}} & {\cellcolor{high}{0.395}} & {\cellcolor{high}{0.284}} & {\cellcolor{high}{0.398}} & {\cellcolor{high}{0.275}} & {\cellcolor{high}{0.404}}  & {\cellcolor{high}{0.298}} & {\cellcolor{high}{0.422}} \\
\midrule
{\textbf{PEMS03}}
& {\cellcolor{high}{0.264}} & {\cellcolor{blank}{0.171}} & {\cellcolor{high}{0.270}} & {\cellcolor{blank}{0.171}} & {\cellcolor{high}{0.401}} & {\cellcolor{high}{0.363}} & {\cellcolor{high}{0.406}} & {\cellcolor{high}{0.373}} & {\cellcolor{high}{0.282}} & {\cellcolor{high}{0.183}} & {\cellcolor{high}{0.284}} & {\cellcolor{high}{0.185}} & {\cellcolor{high}{0.268}} & {\cellcolor{high}{0.178}}  & {\cellcolor{high}{0.281}} & {\cellcolor{high}{0.183}} \\
\midrule
{\textbf{PEMS04}}
& {\cellcolor{high}{0.346}} & {\cellcolor{high}{0.458}} & {\cellcolor{high}{0.358}} & {\cellcolor{high}{0.463}} & {\cellcolor{high}{0.508}} & {\cellcolor{high}{0.772}} & {\cellcolor{high}{0.513}} & {\cellcolor{high}{0.778}} & {\cellcolor{high}{0.376}} & {\cellcolor{high}{0.482}} & {\cellcolor{high}{0.379}} & {\cellcolor{high}{0.484}} & {\cellcolor{high}{0.362}} & {\cellcolor{high}{0.477}}  & {\cellcolor{high}{0.375}} & {\cellcolor{high}{0.488}} \\
\midrule
{\textbf{PEMS07}}
& {\cellcolor{high}{0.221}} & {\cellcolor{high}{0.115}} & {\cellcolor{high}{0.242}} & {\cellcolor{high}{0.130}} & {\cellcolor{high}{0.394}} & {\cellcolor{high}{0.384}} & {\cellcolor{high}{0.405}} & {\cellcolor{high}{0.393}} & {\cellcolor{high}{0.250}} & {\cellcolor{high}{0.137}} & {\cellcolor{high}{0.253}} & {\cellcolor{high}{0.142}} & {\cellcolor{high}{0.235}}  & {\cellcolor{high}{0.136}} & {\cellcolor{high}{0.252}} & {\cellcolor{high}{0.145}}\\
\midrule
{\textbf{PEMS08}}
& {\cellcolor{high}{0.345}} & {\cellcolor{lower}{0.510}} & {\cellcolor{high}{0.353}} & {\cellcolor{lower}{0.509}} & {\cellcolor{high}{0.488}} & {\cellcolor{high}{0.770}} & {\cellcolor{high}{0.506}} & {\cellcolor{high}{0.837}} & {\cellcolor{high}{0.377}} & {\cellcolor{high}{0.527}} & {\cellcolor{high}{0.386}} & {\cellcolor{high}{0.544}} & {\cellcolor{high}{0.359}} & {\cellcolor{high}{0.510}}  & {\cellcolor{high}{0.387}} & {\cellcolor{high}{0.557}} \\
\bottomrule
{}& \multicolumn{4}{c|}{\textbf{WR: \textcolor{red}{77\%}, PR: \textcolor{blue}{69.7\%}, SP: \textcolor{red!50}{1.53x}}} & \multicolumn{4}{c|}{\textbf{WR: \textcolor{red}{88\%}, PR: \textcolor{blue}{63.2\%}, SP: \textcolor{red!50}{1.48x}}}& \multicolumn{4}{c|}{\textbf{WR: \textcolor{red}{77\%}, PR: \textcolor{blue}{63.2\%}, SP: \textcolor{red!50}{1.57x}}}& \multicolumn{4}{c}{\textbf{WR: \textcolor{red}{73\%}, PR: \textcolor{blue}{60.2\%}, SP: \textcolor{red!50}{1.51x}}}\\
\bottomrule
\end{tabular}

}
 \begin{tablenotes}
 \footnotesize
\item[*] WR: Win Rate; PR: Average Pruning Layer Ratio; SP: Average Speedup Ratio.
\end{tablenotes}
\end{threeparttable}
\end{table}

\begin{table}[h]
\centering
\vspace{-0.6cm}
\captionsetup{justification=justified,singlelinecheck=false}
\caption{Comparison of pre- and post-pruning metrics in four TSFMs. Results are averaged across horizons $\in \{96,192,336,720\}$. FT: fine-tuning on new data distributions. Pruning + FT: Prune-then-finetune. \colorbox[HTML]{FFE5C6}{\strut Performance gain} and \colorbox[HTML]{FF8787}{\strut Performance loss} denote improved and degraded metrics.} 
\label{Method adaptation for tsfms}
\begin{threeparttable}
\vspace{-0.2cm}
\renewcommand{\arraystretch}{1.4}
\resizebox{1\textwidth}{!}{\small  
\begin{tabular}{c|cccc|cccc|cccc|cccc}
\toprule
\multirow{3}{*}{\textbf{Models}} & \multicolumn{16}{c}{\cellcolor{gray!40}\textbf{TSFMs}} \\
 &\multicolumn{4}{c|}{\textbf{Sundial$_{\text{Large}}$ (\citeyear{liu2025sundialfamilyhighlycapable})}} & \multicolumn{4}{c|}{\textbf{Chronos$_{\text{Base}}$ (\citeyear{ansari2024chronos})}}& \multicolumn{4}{c|}{\textbf{Moirai$_{\text{Large}}$ (\citeyear{moirai})}}& \multicolumn{4}{c}{\textbf{TimesFM (\citeyear{das2024decoder})}} \\
& \multicolumn{2}{c}{\textbf{Pruning + FT}} & \multicolumn{2}{c|}{\textbf{FT}}& \multicolumn{2}{c}{\textbf{Pruning + FT}} & \multicolumn{2}{c|}{\textbf{FT}}& \multicolumn{2}{c}{\textbf{Pruning + FT}} & \multicolumn{2}{c|}{\textbf{FT}}& \multicolumn{2}{c}{\textbf{Pruning + FT}} & \multicolumn{2}{c}{\textbf{FT}}\\
\midrule
{\textbf{Metric}} &{\textbf{MAE}} &{\textbf{MSE}} &{\textbf{MAE}} &{\textbf{MSE}} &{\textbf{MAE}} &{\textbf{MSE}} &{\textbf{MAE}} &{\textbf{MSE}}&{\textbf{MAE}} &{\textbf{MSE}} &{\textbf{MAE}} &{\textbf{MSE}} &{\textbf{MAE}} &{\textbf{MSE}} &{\textbf{MAE}} &{\textbf{MSE}}\\
\midrule
{\textbf{ETTh1}}
& {\cellcolor{high}{0.408}} & {\cellcolor{high}{0.387}} & {\cellcolor{high}{0.418}} & {\cellcolor{high}{0.398}} & {\cellcolor{high}{0.405}} & {\cellcolor{high}{0.416}} & {\cellcolor{high}{0.411}} & {\cellcolor{high}{0.424}} & {\cellcolor{high}{0.422}} & {\cellcolor{high}{0.425}} & {\cellcolor{high}{0.429}} & {\cellcolor{high}{0.433}} & {\cellcolor{high}{0.411}} & {\cellcolor{high}{0.410}}  & {\cellcolor{high}{0.416}} & {\cellcolor{high}{0.415}} \\
\midrule
{\textbf{ETTh2}}
& {\cellcolor{high}{0.382}} & {\cellcolor{high}{0.327}} & {\cellcolor{high}{0.390}} & {\cellcolor{high}{0.334}} & {\cellcolor{high}{0.366}} & {\cellcolor{high}{0.337}} & {\cellcolor{high}{0.370}} & {\cellcolor{high}{0.344}} & {\cellcolor{lower}{0.392}} & {\cellcolor{high}{0.348}} & {\cellcolor{lower}{0.391}} & {\cellcolor{high}{0.351}} & {\cellcolor{high}{0.369}} & {\cellcolor{high}{0.332}}  & {\cellcolor{high}{0.376}} & {\cellcolor{high}{0.337}} \\
\midrule
{\textbf{ETTm1}}
& {\cellcolor{high}{0.367}} & {\cellcolor{high}{0.329}} & {\cellcolor{high}{0.374}} & {\cellcolor{high}{0.334}} & {\cellcolor{high}{0.356}} & {\cellcolor{high}{0.347}} & {\cellcolor{high}{0.358}} & {\cellcolor{high}{0.354}} & {\cellcolor{high}{0.374}} & {\cellcolor{high}{0.352}} & {\cellcolor{high}{0.381}} & {\cellcolor{high}{0.357}} & {\cellcolor{high}{0.372}} & {\cellcolor{high}{0.366}}  & {\cellcolor{high}{0.379}} & {\cellcolor{high}{0.380}} \\
\midrule
{\textbf{ETTm2}}
& {\cellcolor{blank}{0.315}} & {\cellcolor{high}{0.250}} & {\cellcolor{blank}{0.315}} & {\cellcolor{high}{0.255}} & {\cellcolor{high}{0.298}} & {\cellcolor{high}{0.255}} & {\cellcolor{high}{302}} & {\cellcolor{high}{0.268}} & {\cellcolor{high}{0.313}} & {\cellcolor{high}{0.260}} & {\cellcolor{high}{0.317}} & {\cellcolor{high}{0.263}} & {\cellcolor{high}{0.301}} & {\cellcolor{high}{0.254}}  & {\cellcolor{high}{0.307}} & {\cellcolor{high}{0.259}} \\
\midrule
{\textbf{ECL}}
& {\cellcolor{high}{0.260}} & {\cellcolor{high}{0.161}} & {\cellcolor{high}{0.262}} & {\cellcolor{high}{0.163}} & {\cellcolor{high}{0.230}} & {\cellcolor{high}{0.150}} & {\cellcolor{high}{0.244}} & {\cellcolor{high}{0.156}} & {\cellcolor{high}{0.269}} & {\cellcolor{high}{0.174}} & {\cellcolor{high}{0.274}} & {\cellcolor{high}{0.179}} & -& -& - & - \\
\midrule
{\textbf{Weather}}
& {\cellcolor{high}{0.227}} & {\cellcolor{high}{0.265}} & {\cellcolor{high}{0.233}} & {\cellcolor{high}{0.271}} & {\cellcolor{high}{0.246}} & {\cellcolor{high}{0.222}} & {\cellcolor{high}{0.267}} & {\cellcolor{high}{0.257}} & {\cellcolor{high}{0.282}} & {\cellcolor{high}{0.238}} & {\cellcolor{high}{0.292}} & {\cellcolor{high}{0.253}} & -& -& - & - \\
\bottomrule
{}& \multicolumn{4}{c|}{\textbf{WR: \textcolor{red}{92\%}, PR: \textcolor{blue}{72.9\%}, SP: \textcolor{red!50}{2.87x}}} & \multicolumn{4}{c|}{\textbf{WR: \textcolor{red}{100\%}, PR: \textcolor{blue}{71.3\%}, SP: \textcolor{red!50}{2.64x}}}& \multicolumn{4}{c|}{\textbf{WR: \textcolor{red}{92\%}, PR: \textcolor{blue}{67.1\%}, SP: \textcolor{red!50}{2.33x}}}& \multicolumn{4}{c}{\textbf{WR: \textcolor{red}{100\%}, PR: \textcolor{blue}{81.2\%}, SP: \textcolor{red!50}{3.09x}}}\\
\bottomrule
\end{tabular}

}
 \begin{tablenotes}
 \footnotesize
    \item *WR: Win Rate, PR: Average Pruning Layer Ratio, SP: Average Speedup Ratio.
    \item *Other datasets that have already been involved in pre-training are excluded from evaluation, and “ECL \& \\Weather” in TimesFM are denoted by a dash (“–”).
\end{tablenotes}
\end{threeparttable}
\vspace{-0.2cm}
\end{table}

\vspace{-0.2cm}
\paragraph{Results.} As shown in Tab. \ref{Method adaptation for baselines}, for the four LLM4TS models, we retain only one-third of the layers instead of the entire LLM, resulting in a 1.5× speedup. This pruning strategy surpasses the baseline truncation method in over 80\% of scenarios, with only a minor performance drop (below 1\% on average) in rare cases. As shown in Tab. \ref{Method adaptation for tsfms}, for the four TSFMs models, we retain only about 26.9\% of the layers on average, achieving over a 2.7× speedup. By preserving only the key layers, the performance remains virtually lossless. For various famous LLM4TS and TSFMs in the TS domain, our method demonstrates that retaining only the critical layers not only yields a more lightweight model, but also achieves comparable or even superior forecasting accuracy.

\subsection{Less Is More: Full LLMs Are Not Always Beneficial}
\label{Full vs. Pruned}
To balance the use of LLMs’ prior knowledge and their vast parameter scale, \textsc{Fsca}, \textsc{Ofa}, and \textsc{Calf} adopt a coarse-grained strategy by directly truncating the early layers as the backbone. However, such pruning is not only ineffective in preserving the complete semantic of LLM, but also lacks precise control over the number of retained layers. In practice, an exhaustive search is usually used to find a suboptimal result, lacking any theoretical justification. Our method provides a more targeted and principled approach, leading to better performance. 

In this section, we show that our approach both reduces semantic interference introduced by redundant parameters, and outperforms coarse-grained pruning. Specifically, we explore the impact of 3 layer retention strategies on predictions. \textbf{Pruned}: our pruning method; \textbf{Full}: entire LLM used as backbone without pruning; \textbf{Random}: An equal number of layers to those retained by our pruning method are selected, but their positions are randomly chosen.

Tab. \ref{tab:Performance_of_Different_Layer_Backbones} shows that our pruned models not only outperform the models using the full backbone, but also surpass the randomly selected layers of equal number, regardless of their positions. Random leads to suboptimal performance, indicating layers contribute non-arbitrarily.
\begin{table}[h]
\centering
\vspace{-0.2cm}
\caption{MSE of different retained layers strategy. Our pruning method achieves the lowest MSE, outperforming both the Full and Random schemes. Horizons $\in \{96,192,336,720\}$.}
\vspace{-0.20cm}
\label{tab:Performance_of_Different_Layer_Backbones}
\resizebox{1\textwidth}{!}{ 
\begin{tabular}{c|ccc|ccc|ccc}
\toprule
\multirow{2}{*}{\textbf{Models}} & \multicolumn{3}{c|}{\textbf{FSCA}} & \multicolumn{3}{c|}{\textbf{CALF}} & \multicolumn{3}{c}{\textbf{OFA}}\\
& \textbf{Pruned} & \textbf{Full} & \textbf{Random} & \textbf{Pruned} & \textbf{Full} & \textbf{Random} & \textbf{Pruned} & \textbf{Full} & \textbf{Random} \\
\midrule
{\textbf{ETTh1}} & 0.426 & 0.442 & 0.432 & 0.436 & 0.445 & 0.461 & 0.429 & 0.461 & 0.451 \\
\midrule
{\textbf{ETTh2}} & 0.349 & 0.367 & 0.355 & 0.362 & 0.374 & 0.374 & 0.366 & 0.379 & 0.384 \\
\midrule
{\textbf{ETTm1}} & 0.350 & 0.368 & 0.360 & 0.387 & 0.392 & 0.397 & 0.357 & 0.365 & 0.366 \\
\midrule
{\textbf{ETTm2}} & 0.258 & 0.267 & 0.258 & 0.268 & 0.280 & 0.279 & 0.265 & 0.287 & 0.282 \\
\midrule
{\textbf{Weather}} & 0.233 & 0.237 & 0.244 & 0.255 & 0.267 & 0.278 & 0.230 & 0.248 & 0.231 \\
\midrule
\multirow{2}{*}{\textbf{avg.}} & 0.323
 & 0.336 & 0.330 & 0.342 & 0.352 & 0.358 & 0.329 & 0.348 & 0.343 \\
 & {\textcolor[HTML]{FFD43B}{\faMedal}}
 & {\textcolor[HTML]{E67700}{\faMedal}} & {\textcolor[HTML]{CED4DA}{\faMedal}} & {\textcolor[HTML]{FFD43B}{\faMedal}} & {\textcolor[HTML]{CED4DA}{\faMedal}} & {\textcolor[HTML]{E67700}{\faMedal}} & {\textcolor[HTML]{FFD43B}{\faMedal}} & {\textcolor[HTML]{E67700}{\faMedal}} & {\textcolor[HTML]{CED4DA}{\faMedal}} \\
\bottomrule
\end{tabular}}
\end{table}

\vspace{-.2cm}
\subsection{Ablations}
\paragraph{Pruning Without Fine-tuning.} In the previous experiments, we fine-tune the model after removing redundant layers. Here, we analyze the negative effects when fine-tuning is omitted. As shown in Tab. \ref{tab:Different strategies}, omitting fine-tuning leads to a failure in bridging the gap between the pruned structure and the parameter distribution. Hence, fine-tuning after pruning is indispensable.

\begin{figure}[h]
\vspace{+0.3cm}
    \centering
    \begin{minipage}[t]{0.46\linewidth}
        \vspace{-0.1cm}
        \centering
        \includegraphics[width=\linewidth]{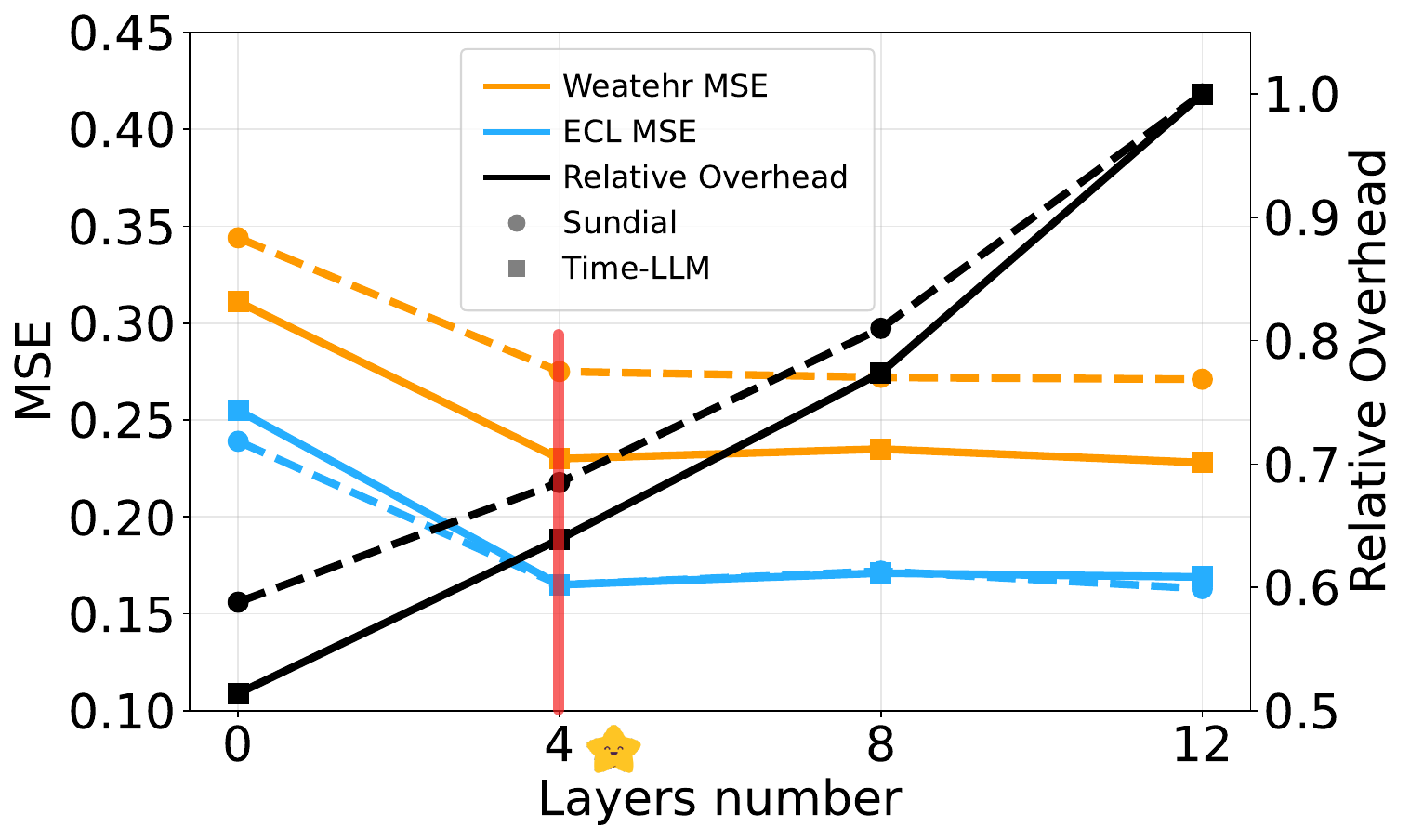}
        \vspace{-0.5cm}
        \captionsetup{singlelinecheck=true}
        \caption{Overhead–Accuracy Trade-off.}
        \label{fig:Trade-off between efficiency and accuracy}
    \end{minipage}
    \hspace{0.005\linewidth}
    \begin{minipage}[t]{0.52\linewidth}
        \vspace{-0.5cm}
        \centering
        \renewcommand{\arraystretch}{2.3}
        \setlength{\tabcolsep}{5pt}
        \begin{table}[H]
        \caption{Accuracy of our method on TS classification tasks, achieving better performance with fewer parameters.}
        \label{tab:Results for classification task}
        \vspace{-0.2cm}
        \resizebox{\linewidth}{!}{
        \begin{tabular}{c|cc|cc|cc}
        \toprule
        \multirow{2}{*}{\textbf{Dataset}} & \multicolumn{2}{c|}{\textbf{FSCA}} & \multicolumn{2}{c|}{\textbf{Time-LLM(G)}} & \multicolumn{2}{c}{\textbf{OFA}} \\
        & \textbf{Pruned} & \textbf{Original} & \textbf{Pruned} & \textbf{Original}& \textbf{Pruned} & \textbf{Original}\\
        \midrule
        \textbf{Handwriting}     & {\cellcolor{high}{38.8}} & {\cellcolor{high}{38.6}} & {\cellcolor{lower}{33.4}} & {\cellcolor{lower}{33.6}} & {\cellcolor{high}{32.9}} & {\cellcolor{high}{32.7}} \\
        \textbf{Heartbeat}       & {\cellcolor{high}{79.1}} & {\cellcolor{high}{78.7}} & {\cellcolor{lower}{78.4}} & {\cellcolor{lower}{78.6}} & {\cellcolor{blank}{77.2}} & {\cellcolor{blank}{77.2}} \\
        \textbf{JapaneseVowels}  & {\cellcolor{lower}{98.6}} & {\cellcolor{lower}{98.9}} & {\cellcolor{high}{98.3}} & {\cellcolor{high}{97.8}} & {\cellcolor{high}{98.7}} & {\cellcolor{high}{98.5}} \\
        \bottomrule
        \end{tabular}
        }
        \end{table}
    \end{minipage}
    \vspace{-0.3cm}
\end{figure}

\begin{table}[t]
\centering
\vspace{-0.5cm}
\caption{Results of pruning w \& w/o fine-tuning. \textbf{Vanilla} refers to fine-tuning after pruning, while \textbf{w/o} indicates that fine-tuning is omitted. {$\triangle$} indicates the difference between the latter and the formerr.}
\vspace{-0.20cm}
\label{tab:Different strategies}
\resizebox{0.7\textwidth}{!}{ 
\begin{tabular}{c|c|ccc|ccc}
\toprule
\multirow{2}{*}{\textbf{Dataset}} & \multirow{2}{*}{\textbf{Metric}} & \multicolumn{3}{c|}{\textbf{OFA}} & \multicolumn{3}{c}{\textbf{Sundial$_{\text{Large}}$}} \\
 &  & vanilla & w/o & {$\triangle$}& vanilla & w/o & {$\triangle$}\\
\midrule
\multirow{2}{*}{\textbf{ETTh1}} & MAE & 0.435 & 0.554 &{\cellcolor{blank1}{0.121}}\includegraphics[width=0.02\linewidth]{Figure/increase.png}& 0.408 & 0.512 &{\cellcolor{blank1}{0.104}}\includegraphics[width=0.02\linewidth]{Figure/increase.png}\\  
 & MSE & 0.429 & 0.627 &{\cellcolor{blank1}{0.198}}\includegraphics[width=0.02\linewidth]{Figure/increase.png}& 0.387 & 0.455 &{\cellcolor{blank1}{0.068}}\includegraphics[width=0.02\linewidth]{Figure/increase.png}\\                                  
\midrule
\multirow{2}{*}{\textbf{ECL}} & MAE & 0.259 & 0.318 &{\cellcolor{blank1}{0.059}}\includegraphics[width=0.02\linewidth]{Figure/increase.png}& 0.260 & 0.334 &{\cellcolor{blank1}{0.074}}\includegraphics[width=0.02\linewidth]{Figure/increase.png}\\
 & MSE & 0.166 & 0.223 &{\cellcolor{blank1}{0.057}}\includegraphics[width=0.02\linewidth]{Figure/increase.png}& 0.161 & 0.217 &{\cellcolor{blank1}{0.056}}\includegraphics[width=0.02\linewidth]{Figure/increase.png}\\
\bottomrule
\end{tabular}}
\end{table}
\vspace{-0.2cm}
\paragraph{Trade-off Between Efficiency and Accuracy.}
\label{subsec:Inference Overhead and Performance}
We retain different numbers of layers in descending order of importance to explore the trade-off between efficiency and accuracy. Relative overhead is defined as $ {T_{\text{pruned model}}}/{T_{\text{original model}}}$, and $T$ is inference time. Fig. \ref{fig:Trade-off between efficiency and accuracy} shows that just keep top 4 critical layers perform well, yet eliminating all layers results in a notably worse MSE. Retaining 1/3 of the layers achieves an optimal balance, as the forecasting performance is almost saturated while requiring merely 65\% of the inference time compared to the full model.

\vspace{-0.25cm}
\paragraph{Potential of  Method on TS Classification.} To emphasize the broad applicability of our approach, we evaluate it against three baseline models on three distinct datasets: Handwriting, Heartbeat, and JapaneseVowels \citep{ICLR2025_e1de63ec}. As shown in Tab. \ref{tab:Results for classification task}, our method shows strong performance on TS forecasting tasks, and we extend it to TS classification to examine its potential.

\section{Discussion }
\begin{figure}[h]
\vspace{-0.0cm}
    \centering
    \includegraphics[width=1\linewidth]{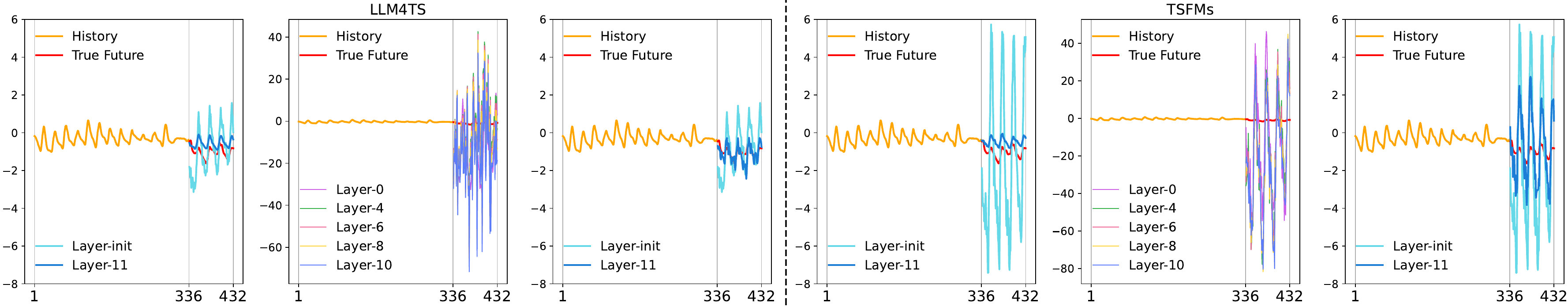}
        \vspace{-.4cm}
    \caption{Visualization of hidden states projected into the TS domain. Cols. 1\textasciitilde3 show results from LLM4TS models, and Cols. 4\textasciitilde6 present those from TSFMs.}
    \vspace{-.2cm}
    \label{Projecting hidden states into the TS}
\end{figure}

\paragraph{Discussion 1: From hidden states to prediction.} \label{subsection:From Representation to Prediction: A Progressive Perspective} Building on our finding that model still performs well after pruning most layers with fine-tuning, we further analyze how temporal hidden states evolve across layers. The hidden states from each layer are fed into Prediction head to generate forecasts, as shown in Fig. \ref{Projecting hidden states into the TS}. 

For reference, (Cols. 1 \& 4) correspond to the unpruned model, where Prediction head consumes hidden states from  the initial state (input embedding with positional encoding only) and final layer. By preserving only the first and last layers (Cols. 3 \& 6) without any fine-tuning, the model still delivers competitive performance. In contrast, projecting intermediate layers back to the TS yields outputs that diverge from ground-truth distribution (Cols. 2 \& 5). 

\begin{figure}[h]
    \centering
    \begin{subfigure}[b]{1\linewidth}
        \centering
        \includegraphics[width=\linewidth]{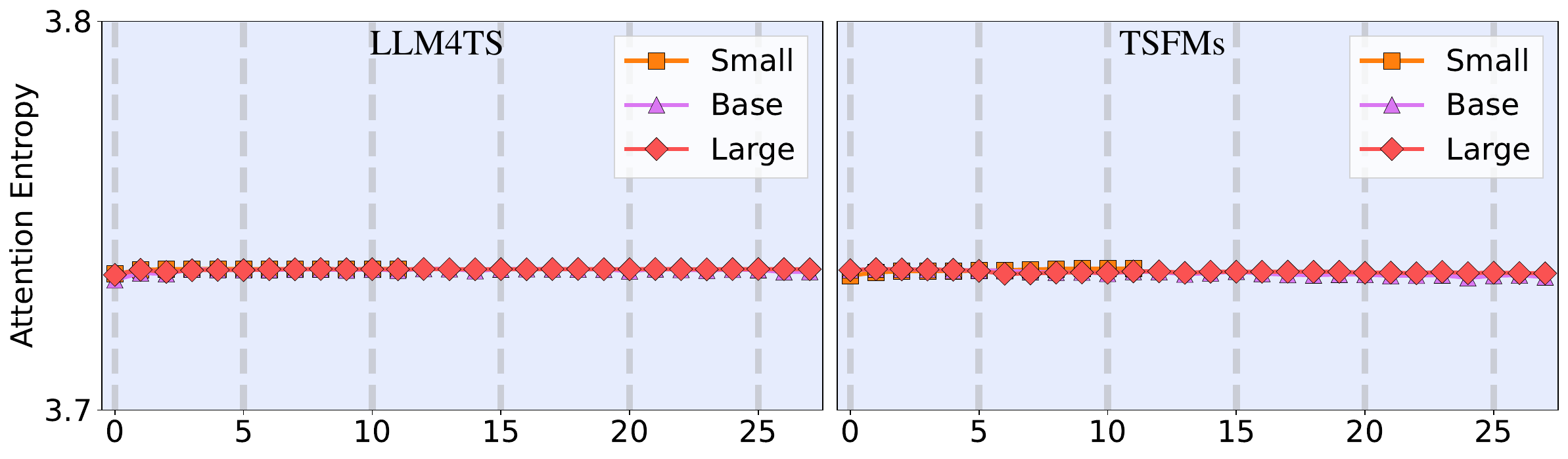}
        \caption{Three model scales under single-dataset learning.}
    \end{subfigure}
    
    \begin{subfigure}[b]{1\linewidth}
        \centering
        \includegraphics[width=\linewidth]{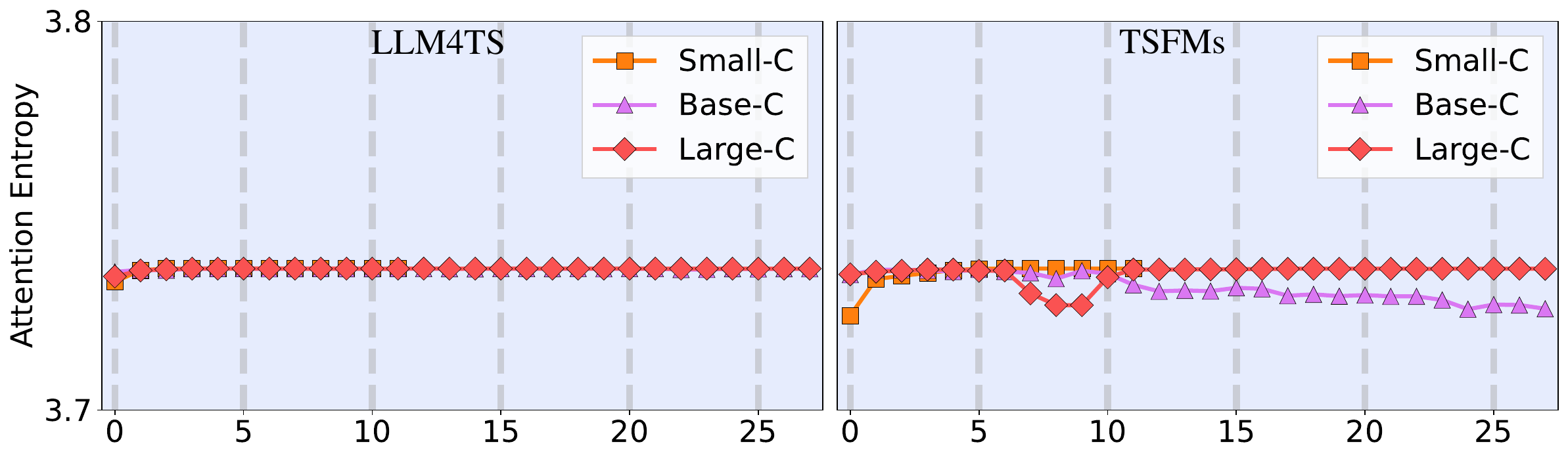}
        \caption{Three model scales under cross-dataset learning.}
    \end{subfigure}
        \vspace{-.3cm}
    \caption{Attention entropy of all layers across twelve models. All layers exhibit nearly identical entropy values (3.7\textasciitilde3.8), suggesting minimal diversity in attention distribution.}
    \vspace{-.1cm}
    \label{fig:Attention entropy}
\end{figure}

\begin{figure}[h]
    \centering
    \vspace{-0.5em}
    \begin{subfigure}[b]{0.48\linewidth}
        \centering
        \includegraphics[width=\linewidth]{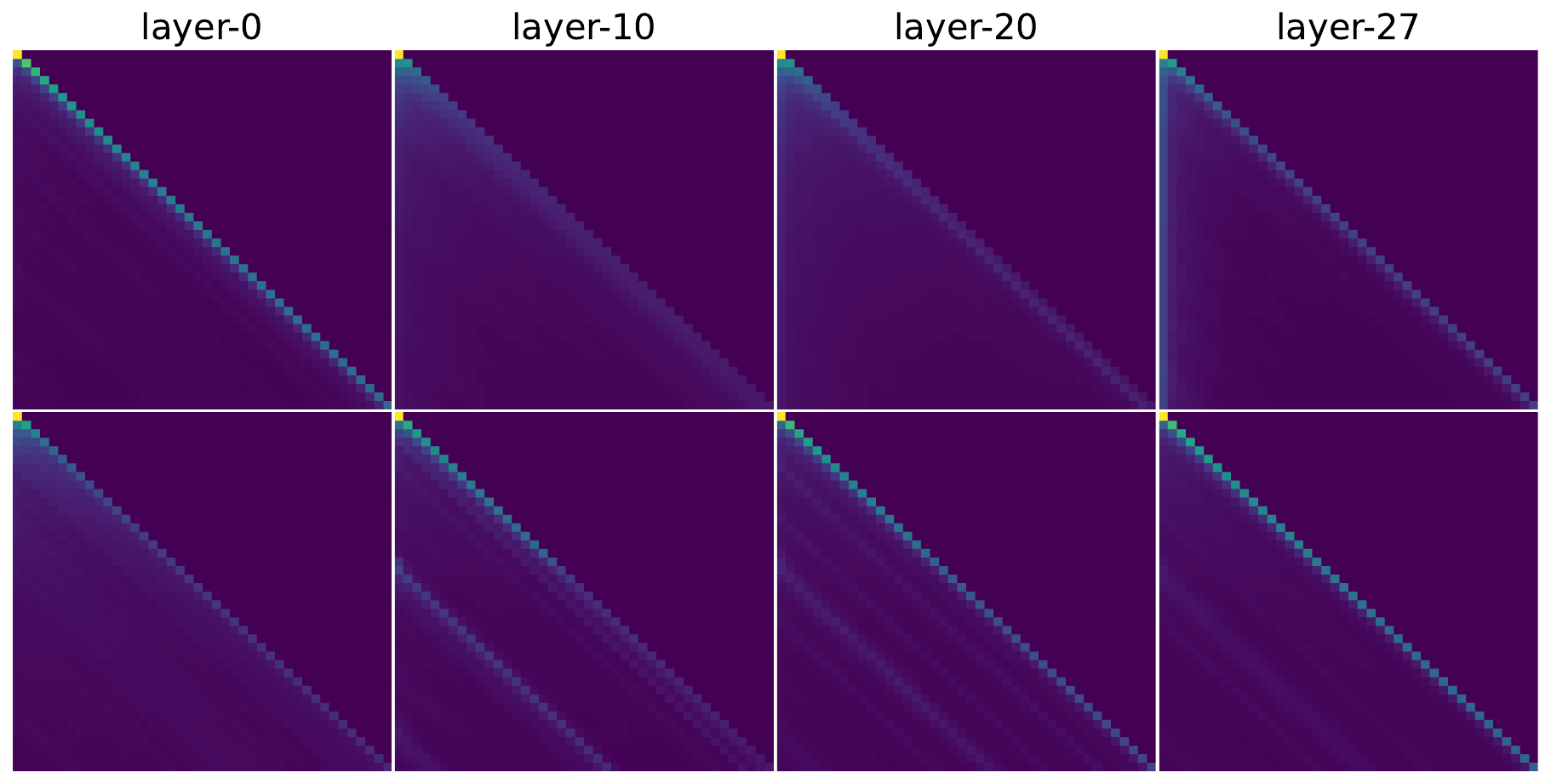}
    \end{subfigure}
    \begin{subfigure}[b]{0.48\linewidth}
        \centering
        \includegraphics[width=\linewidth]{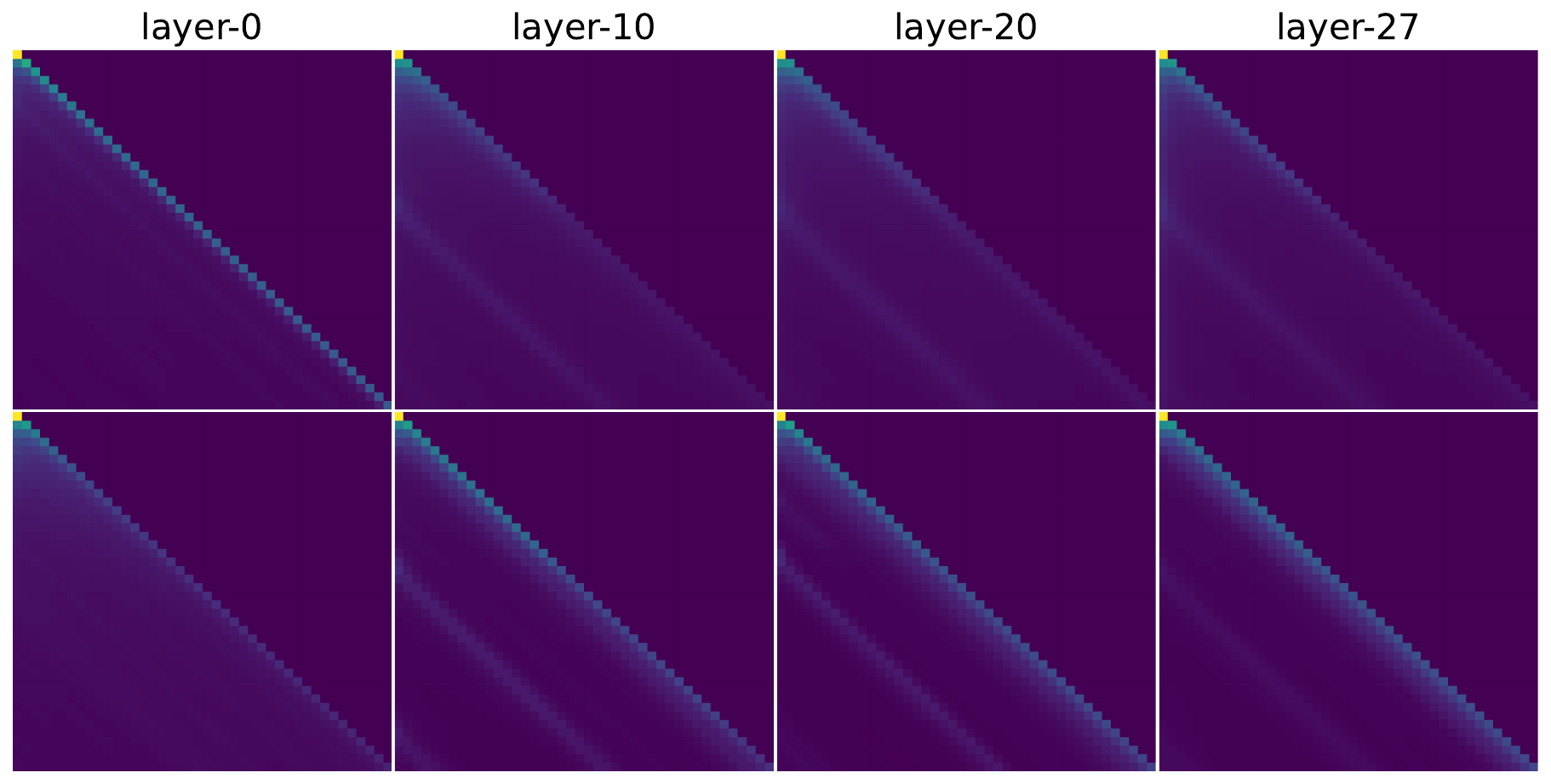}
    \end{subfigure}
    \caption{Mean attention weights across all heads for four models. The attention is primarily focused on the diagonal, with only a few tokens being critical for prediction. \textbf{(\emph{Upper-left})} attention weights of LLM4TS-Base, \textbf{(\emph{Lower-left})} attention weights of TSFMs-Base, \textbf{(\emph{Upper-right})} attention weights of LLM4TS-Large, \textbf{(\emph{Lower-right})} attention weights of TSFMs-Large.}
    \label{fig:attention weiht}
\end{figure}

The representations in layers 2\textasciitilde10 exhibit substantial deviation from the TS domain, whereas the first and last layers primarily function as domain adapters between the TS input and the model’s internal representations. Consequently, even when only layers 0 and 11 are retained and all others are pruned, the pruned model still delivers satisfactory performance without any fine-tuning.
\vspace{-.2cm}
\paragraph{Discussion 2: Attention weights as a window into pattern learning for TS.} \label{subsection:Attention Weights as a Window into Pattern Learning for TS}
Attention entropy $\mathcal{H}^{l}$ measures the attention distribution of each head across tokens \citep{zhang2025attentionentropykeyfactor}. Interestingly, we find that $\mathcal{H}^{l}$ remains stable across layers in Fig. \ref{fig:Attention entropy}, meaning limited change in attention dispersion. Moreover, heads often focus on diagonal positions, which exert dominant influence on prediction in Fig. \ref{fig:attention weiht}. If very few tokens determine the prediction, it is understandable that a small set of informative token features may suffice without extensive parameters \citep{kim2022learned,wen2025token}.

\section{Conclusion}
In this work, we conduct evaluations of large-scale TS models across architectures, scales, data volumes, distributions, and learning strategies to investigate their effects on the scaling paradox, provide the first in-depth analysis of inter- and intra-layer representations, reveal the phenomenon of few-layer dominance whereby only a small subset of layers are critical, and further propose a critical-layer identification method that preserves forecasting accuracy while improving inference efficiency, with extensive validation on eight representative models confirming its universality and effectiveness. In future work, we will incorporate the discovered few-layer dominance phenomenon into the architectural design of large-scale TS models, aiming to achieve more efficient and high-performing forecasting systems.
\clearpage

\bibliography{iclr2026_conference}

@misc{zhang2025attentionentropykeyfactor,
      title={Attention Entropy is a Key Factor: An Analysis of Parallel Context Encoding with Full-attention-based Pre-trained Language Models}, 
      author={Zhisong Zhang and Yan Wang and Xinting Huang and Tianqing Fang and Hongming Zhang and Chenlong Deng and Shuaiyi Li and Dong Yu},
      year={2025},
      eprint={2412.16545},
      archivePrefix={arXiv},
      primaryClass={cs.CL},
      url={https://arxiv.org/abs/2412.16545}, 
}

@misc{yang2025qwen3technicalreport,
      title={Qwen3 Technical Report}, 
      author={An Yang and Anfeng Li and Baosong Yang and Beichen Zhang and Binyuan Hui and Bo Zheng and Bowen Yu and Chang Gao and Chengen Huang and Chenxu Lv and Chujie Zheng and Dayiheng Liu and Fan Zhou and Fei Huang and Feng Hu and Hao Ge and Haoran Wei and Huan Lin and Jialong Tang and Jian Yang and Jianhong Tu and Jianwei Zhang and Jianxin Yang and Jiaxi Yang and Jing Zhou and Jingren Zhou and Junyang Lin and Kai Dang and Keqin Bao and Kexin Yang and Le Yu and Lianghao Deng and Mei Li and Mingfeng Xue and Mingze Li and Pei Zhang and Peng Wang and Qin Zhu and Rui Men and Ruize Gao and Shixuan Liu and Shuang Luo and Tianhao Li and Tianyi Tang and Wenbiao Yin and Xingzhang Ren and Xinyu Wang and Xinyu Zhang and Xuancheng Ren and Yang Fan and Yang Su and Yichang Zhang and Yinger Zhang and Yu Wan and Yuqiong Liu and Zekun Wang and Zeyu Cui and Zhenru Zhang and Zhipeng Zhou and Zihan Qiu},
      year={2025},
      eprint={2505.09388},
      archivePrefix={arXiv},
      primaryClass={cs.CL},
      url={https://arxiv.org/abs/2505.09388}, 
}

@article{radford2018improving,
  title={Improving language understanding by generative pre-training},
  author={Radford, Alec and Narasimhan, Karthik and Salimans, Tim and Sutskever, Ilya and others},
  year={2018},
  publisher={San Francisco, CA, USA}
}

@misc{touvron2023llamaopenefficientfoundation,
      title={LLaMA: Open and Efficient Foundation Language Models}, 
      author={Hugo Touvron and Thibaut Lavril and Gautier Izacard and Xavier Martinet and Marie-Anne Lachaux and Timothée Lacroix and Baptiste Rozière and Naman Goyal and Eric Hambro and Faisal Azhar and Aurelien Rodriguez and Armand Joulin and Edouard Grave and Guillaume Lample},
      year={2023},
      eprint={2302.13971},
      archivePrefix={arXiv},
      primaryClass={cs.CL},
      url={https://arxiv.org/abs/2302.13971}, 
}

@article{zhou2023one,
  title={One fits all: Power general time series analysis by pretrained lm},
  author={Zhou, Tian and Niu, Peisong and Sun, Liang and Jin, Rong and others},
  journal={Advances in neural information processing systems},
  year={2023}
}

@inproceedings{xiaoming2025time,
  title={Time-MoE: Billion-Scale Time Series Foundation Models with Mixture of Experts},
  author={Xiaoming, Shi and Shiyu, Wang and Yuqi, Nie and Dianqi, Li and Zhou, Ye and Qingsong, Wen and Jin, Ming},
  booktitle={ICLR 2025: The Thirteenth International Conference on Learning Representations},
  year={2025},
  organization={International Conference on Learning Representations}
}

@article{wilinski2024exploring,
  title={Exploring representations and interventions in time series foundation models},
  author={Wili{\'n}ski, Micha{\l} and Goswami, Mononito and Potosnak, Willa and {\.Z}ukowska, Nina and Dubrawski, Artur},
  journal={arXiv preprint arXiv:2409.12915},
  year={2024}
}

@inproceedings{das2024decoder,
  title={A decoder-only foundation model for time-series forecasting},
  author={Das, Abhimanyu and Kong, Weihao and Sen, Rajat and Zhou, Yichen},
  booktitle={Forty-first International Conference on Machine Learning},
  year={2024}
}

@article{chang2025llm4ts,
  title={Llm4ts: Aligning pre-trained llms as data-efficient time-series forecasters},
  author={Chang, Ching and Wang, Wei-Yao and Peng, Wen-Chih and Chen, Tien-Fu},
  journal={ACM Transactions on Intelligent Systems and Technology},
  volume={16},
  number={3},
  pages={1--20},
  year={2025},
  publisher={ACM New York, NY}
}

@article{zhao2025less,
  title={Less is More: Unlocking Specialization of Time Series Foundation Models via Structured Pruning},
  author={Zhao, Lifan and Shen, Yanyan and Liu, Zhaoyang and Wang, Xue and Deng, Jiaji},
  journal={arXiv preprint arXiv:2505.23195},
  year={2025}
}

@article{jin2023time,
  title={Time-llm: Time series forecasting by reprogramming large language models},
  author={Jin, Ming and Wang, Shiyu and Ma, Lintao and Chu, Zhixuan and Zhang, James Y and Shi, Xiaoming and Chen, Pin-Yu and Liang, Yuxuan and Li, Yuan-Fang and Pan, Shirui and others},
  journal={arXiv preprint arXiv:2310.01728},
  year={2023}
}

@inproceedings{liu2025calf,
  title={Calf: Aligning llms for time series forecasting via cross-modal fine-tuning},
  author={Liu, Peiyuan and Guo, Hang and Dai, Tao and Li, Naiqi and Bao, Jigang and Ren, Xudong and Jiang, Yong and Xia, Shu-Tao},
  booktitle={Proceedings of the AAAI Conference on Artificial Intelligence},
  year={2025}
}

@article{zhang2024large,
  title={Large language models for time series: A survey},
  author={Zhang, Xiyuan and Chowdhury, Ranak Roy and Gupta, Rajesh K and Shang, Jingbo},
  journal={arXiv preprint arXiv:2402.01801},
  year={2024}
}

@article{wang2025lightgts,
  title={LightGTS: A Lightweight General Time Series Forecasting Model},
  author={Wang, Yihang and Qiu, Yuying and Chen, Peng and Shu, Yang and Rao, Zhongwen and Pan, Lujia and Yang, Bin and Guo, Chenjuan},
  journal={arXiv preprint arXiv:2506.06005},
  year={2025}
}

@article{goswami2024moment,
  title={Moment: A family of open time-series foundation models},
  author={Goswami, Mononito and Szafer, Konrad and Choudhry, Arjun and Cai, Yifu and Li, Shuo and Dubrawski, Artur},
  journal={arXiv preprint arXiv:2402.03885},
  year={2024}
}

@article{ansari2024chronos,
  title={Chronos: Learning the language of time series},
  author={Ansari, Abdul Fatir and Stella, Lorenzo and Turkmen, Caner and Zhang, Xiyuan and Mercado, Pedro and Shen, Huibin and Shchur, Oleksandr and Rangapuram, Syama Sundar and Arango, Sebastian Pineda and Kapoor, Shubham and others},
  journal={arXiv preprint arXiv:2403.07815},
  year={2024}
}

@inproceedings{Yuqietal-2023-PatchTST,
  title     = {A Time Series is Worth 64 Words: Long-term Forecasting with Transformers},
  author    = {Nie, Yuqi and
               H. Nguyen, Nam and
               Sinthong, Phanwadee and 
               Kalagnanam, Jayant},
  booktitle = {International Conference on Learning Representations},
  year      = {2023}
}

@inproceedings{
zhang2023crossformer,
title={Crossformer: Transformer Utilizing Cross-Dimension Dependency for Multivariate Time Series Forecasting},
author={Yunhao Zhang and Junchi Yan},
booktitle={International Conference on Learning Representations},
year={2023},
}

@inproceedings{ICLR2024_2ea18fdc,
 author = {Liu, Yong and Hu, Tengge and Zhang, Haoran and Wu, Haixu and Wang, Shiyu and Ma, Lintao and Long, Mingsheng},
 booktitle = {International Conference on Representation Learning},
 editor = {B. Kim and Y. Yue and S. Chaudhuri and K. Fragkiadaki and M. Khan and Y. Sun},
 title = {iTransformer: Inverted Transformers Are Effective for Time Series Forecasting},
 url = {https://proceedings.iclr.cc/paper_files/paper/2024/file/2ea18fdc667e0ef2ad82b2b4d65147ad-Paper-Conference.pdf},
 year = {2024}
}

@misc{liu2025sundialfamilyhighlycapable,
      title={Sundial: A Family of Highly Capable Time Series Foundation Models}, 
      author={Yong Liu and Guo Qin and Zhiyuan Shi and Zhi Chen and Caiyin Yang and Xiangdong Huang and Jianmin Wang and Mingsheng Long},
      year={2025},
      eprint={2502.00816},
      archivePrefix={arXiv},
      primaryClass={cs.LG},
      url={https://arxiv.org/abs/2502.00816},
}

@inproceedings{zeng2023transformers,
  title={Are transformers effective for time series forecasting?},
  author={Zeng, Ailing and Chen, Muxi and Zhang, Lei and Xu, Qiang},
  booktitle={Proceedings of the AAAI conference on artificial intelligence},
  year={2023}
}

@inproceedings{ICLR2025_e1de63ec,
 author = {Hu, Yuxiao and Li, Qian and Zhang, Dongxiao and Yan, Jinyue and Chen, Yuntian},
 booktitle = {International Conference on Representation Learning},
 editor = {Y. Yue and A. Garg and N. Peng and F. Sha and R. Yu},
 pages = {90696--90722},
 title = {Context-Alignment: Activating and Enhancing LLMs Capabilities in Time Series},
 url = {https://proceedings.iclr.cc/paper_files/paper/2025/file/e1de63ec74f40d3234c4e053f3528e18-Paper-Conference.pdf},
 volume = {2025},
 year = {2025}
}

@inproceedings{zhuocheng-etal-2023-scaling,
    title = "Scaling Law for Document Neural Machine Translation",
    author = "Zhuocheng, Zhang  and
      Gu, Shuhao  and
      Zhang, Min  and
      Feng, Yang",
    editor = "Bouamor, Houda  and
      Pino, Juan  and
      Bali, Kalika",
    booktitle = "Findings of the Association for Computational Linguistics: EMNLP 2023",
    month = dec,
    year = "2023",
    address = "Singapore",
    publisher = "Association for Computational Linguistics",
    url = "https://aclanthology.org/2023.findings-emnlp.556/",
    doi = "10.18653/v1/2023.findings-emnlp.556",
    pages = "8290--8303"
}

@article{isik2024scaling,
  title={Scaling laws for downstream task performance in machine translation},
  author={Isik, Berivan and Ponomareva, Natalia and Hazimeh, Hussein and Paparas, Dimitris and Vassilvitskii, Sergei and Koyejo, Sanmi},
  journal={arXiv preprint arXiv:2402.04177},
  year={2024}
}

@inproceedings{Chowdhery2022PaLMSL,
    title   = {PaLM: Scaling Language Modeling with Pathways},
    author  = {Aakanksha Chowdhery and Sharan Narang and Jacob Devlin and Maarten Bosma and Gaurav Mishra and Adam Roberts and Paul Barham and Hyung Won Chung and Charles Sutton and Sebastian Gehrmann and Parker Schuh and Kensen Shi and Sasha Tsvyashchenko and Joshua Maynez and Abhishek Rao and Parker Barnes and Yi Tay and Noam M. Shazeer and Vinodkumar Prabhakaran and Emily Reif and Nan Du and Benton C. Hutchinson and Reiner Pope and James Bradbury and Jacob Austin and Michael Isard and Guy Gur-Ari and Pengcheng Yin and Toju Duke and Anselm Levskaya and Sanjay Ghemawat and Sunipa Dev and Henryk Michalewski and Xavier Garc{\'i}a and Vedant Misra and Kevin Robinson and Liam Fedus and Denny Zhou and Daphne Ippolito and David Luan and Hyeontaek Lim and Barret Zoph and Alexander Spiridonov and Ryan Sepassi and David Dohan and Shivani Agrawal and Mark Omernick and Andrew M. Dai and Thanumalayan Sankaranarayana Pillai and Marie Pellat and Aitor Lewkowycz and Erica Oliveira Moreira and Rewon Child and Oleksandr Polozov and Katherine Lee and Zongwei Zhou and Xuezhi Wang and Brennan Saeta and Mark Diaz and Orhan Firat and Michele Catasta and Jason Wei and Kathleen S. Meier-Hellstern and Douglas Eck and Jeff Dean and Slav Petrov and Noah Fiedel},
    year    = {2022}
}

@article{woo2024unified,
  title={Unified training of universal time series forecasting transformers},
  author={Woo, Gerald and Liu, Chenghao and Kumar, Akshat and Xiong, Caiming and Savarese, Silvio and Sahoo, Doyen},
  year={2024},
  publisher={PMLR}
}

@inproceedings{huang2025exploiting,
  title={Exploiting Language Power for Time Series Forecasting with Exogenous Variables},
  author={Huang, Qihe and Zhou, Zhengyang and Yang, Kuo and Wang, Yang},
  booktitle={Proceedings of the ACM on Web Conference 2025},
  pages={4043--4052},
  year={2025}
}

@inproceedings{pmlr-v97-kornblith19a,
  title = {Similarity of Neural Network Representations Revisited},
  author = {Kornblith, Simon and Norouzi, Mohammad and Lee, Honglak and Hinton, Geoffrey},
  booktitle = {Proceedings of the 36th International Conference on Machine Learning},
  pages = {3519--3529},
  year = {2019},
  volume = {97},
  month = {09--15 Jun},
  publisher = {PMLR}
}

@inproceedings{liu2021attention,
  title={Attention as relation: learning supervised multi-head self-attention for relation extraction},
  author={Liu, Jie and Chen, Shaowei and Wang, Bingquan and Zhang, Jiaxin and Li, Na and Xu, Tong},
  booktitle={Proceedings of the twenty-ninth international conference on international joint conferences on artificial intelligence},
  pages={3787--3793},
  year={2021}
}

@inproceedings{kim2022learned,
  title={Learned token pruning for transformers},
  author={Kim, Sehoon and Shen, Sheng and Thorsley, David and Gholami, Amir and Kwon, Woosuk and Hassoun, Joseph and Keutzer, Kurt},
  booktitle={Proceedings of the 28th ACM SIGKDD conference on knowledge discovery and data mining},
  pages={784--794},
  year={2022}
}

@article{wen2025token,
  title={Token Pruning in Multimodal Large Language Models: Are We Solving the Right Problem?},
  author={Wen, Zichen and Gao, Yifeng and Li, Weijia and He, Conghui and Zhang, Linfeng},
  journal={arXiv preprint arXiv:2502.11501},
  year={2025}
}

@inproceedings{pan2024s,
  title={SSIP-LLM: Semantic space informed prompt learning with LLM for time series forecasting},
  author={Pan, Zijie and Jiang, Yushan and Garg, Sahil and Schneider, Anderson and Nevmyvaka, Yuriy and Song, Dongjin},
  booktitle={Forty-first International Conference on Machine Learning},
  year={2024}
}

@inproceedings{liu2025timecma,
  title={Timecma: Towards llm-empowered multivariate time series forecasting via cross-modality alignment},
  author={Liu, Chenxi and Xu, Qianxiong and Miao, Hao and Yang, Sun and Zhang, Lingzheng and Long, Cheng and Li, Ziyue and Zhao, Rui},
  booktitle={Proceedings of the AAAI Conference on Artificial Intelligence},
  year={2025}
}

@inproceedings{liu2024unitime,
  title={Unitime: A language-empowered unified model for cross-domain time series forecasting},
  author={Liu, Xu and Hu, Junfeng and Li, Yuan and Diao, Shizhe and Liang, Yuxuan and Hooi, Bryan and Zimmermann, Roger},
  booktitle={Proceedings of the ACM Web Conference 2024},
  pages={4095--4106},
  year={2024}
}

@article{liu2024timer,
  title={Timer: Transformers for Time Series Analysis at Scale},
  author={Liu, Yong and Zhang, Haoran and Li, Chenyu and Huang, Xiangdong and Wang, Jianmin and Long, Mingsheng},
  journal={arXiv preprint arXiv:2402.02368},
  year={2024} 
}

@inproceedings{liu2022pyraformer,
title={Pyraformer: Low-Complexity Pyramidal Attention for Long-Range Time Series Modeling and Forecasting},
author={Liu, Shizhan and Yu, Hang and Liao, Cong and Li, Jianguo and Lin, Weiyao and Liu, Alex X and Dustdar, Schahram},
booktitle={International Conference on Learning Representations},
year={2022}
}

@inproceedings{wu2021autoformer,
  title={Autoformer: Decomposition Transformers with {Auto-Correlation} for Long-Term Series Forecasting},
  author={Haixu Wu and Jiehui Xu and Jianmin Wang and Mingsheng Long},
  booktitle={Advances in Neural Information Processing Systems},
  year={2021}
}

@inproceedings{zhou2022fedformer,
  title={Fedformer: Frequency enhanced decomposed transformer for long-term series forecasting},
  author={Zhou, Tian and Ma, Ziqing and Wen, Qingsong and Wang, Xue and Sun, Liang and Jin, Rong},
  booktitle={International conference on machine learning},
  pages={27268--27286},
  year={2022},
  organization={PMLR}
}

@article{xue2023promptcast,
  title={Promptcast: A new prompt-based learning paradigm for time series forecasting},
  author={Xue, Hao and Salim, Flora D},
  journal={IEEE Transactions on Knowledge and Data Engineering},
  volume={36},
  number={11},
  pages={6851--6864},
  year={2023},
  publisher={IEEE}
}

@article{cao2023tempo,
  title={Tempo: Prompt-based generative pre-trained transformer for time series forecasting},
  author={Cao, Defu and Jia, Furong and Arik, Sercan O and Pfister, Tomas and Zheng, Yixiang and Ye, Wen and Liu, Yan},
  journal={arXiv preprint arXiv:2310.04948},
  year={2023}
}

@article{kitaev2020reformer,
  title={Reformer: The efficient transformer},
  author={Kitaev, Nikita and Kaiser, {\L}ukasz and Levskaya, Anselm},
  journal={arXiv preprint arXiv:2001.04451},
  year={2020}
}

@inproceedings{moirai,
author = {Woo, Gerald and Liu, Chenghao and Kumar, Akshat and Xiong, Caiming and Savarese, Silvio and Sahoo, Doyen},
title = {Unified training of universal time series forecasting transformers},
year = {2024},
publisher = {JMLR.org},
booktitle = {Proceedings of the 41st International Conference on Machine Learning},
articleno = {2178},
numpages = {25},
location = {Vienna, Austria},
series = {ICML'24}
}

@misc{loshchilov2019decoupledweightdecayregularization,
      title={Decoupled Weight Decay Regularization}, 
      author={Ilya Loshchilov and Frank Hutter},
      year={2019},
      eprint={1711.05101},
      archivePrefix={arXiv},
      primaryClass={cs.LG},
      url={https://arxiv.org/abs/1711.05101}, 
}

@inproceedings{wolf-etal-2020-transformers,
    title = "Transformers: State-of-the-Art Natural Language Processing",
    author = "Wolf, Thomas  and
      Debut, Lysandre  and
      Sanh, Victor  and
      Chaumond, Julien  and
      Delangue, Clement  and
      Moi, Anthony  and
      Cistac, Pierric  and
      Rault, Tim  and
      Louf, Remi  and
      Funtowicz, Morgan  and
      Davison, Joe  and
      Shleifer, Sam  and
      von Platen, Patrick  and
      Ma, Clara  and
      Jernite, Yacine  and
      Plu, Julien  and
      Xu, Canwen  and
      Le Scao, Teven  and
      Gugger, Sylvain  and
      Drame, Mariama  and
      Lhoest, Quentin  and
      Rush, Alexander",
    editor = "Liu, Qun  and
      Schlangen, David",
    booktitle = "Proceedings of the 2020 Conference on Empirical Methods in Natural Language Processing: System Demonstrations",
    month = oct,
    year = "2020",
    address = "Online",
    publisher = "Association for Computational Linguistics",
    url = "https://aclanthology.org/2020.emnlp-demos.6/",
    doi = "10.18653/v1/2020.emnlp-demos.6",
    pages = "38--45"
}

@inproceedings{chencloser,
  title={A Closer Look at Transformers for Time Series Forecasting: Understanding Why They Work and Where They Struggle},
  author={Chen, Yu and C{\'e}spedes, Nathalia and Barnaghi, Payam},
year={2025},
  booktitle={Forty-second International Conference on Machine Learning}
}

@article{lim2021time,
  title={Time-series forecasting with deep learning: a survey},
  author={Lim, Bryan and Zohren, Stefan},
  journal={Philosophical Transactions of the Royal Society A},
  volume={379},
  number={2194},
  pages={20200209},
  year={2021},
  publisher={The Royal Society Publishing}
}

@article{chen2023long,
  title={Long sequence time-series forecasting with deep learning: A survey},
  author={Chen, Zonglei and Ma, Minbo and Li, Tianrui and Wang, Hongjun and Li, Chongshou},
  journal={Information Fusion},
  volume={97},
  pages={101819},
  year={2023},
  publisher={Elsevier}
}

@inproceedings{tong2023probabilistic,
  title={Probabilistic decomposition transformer for time series forecasting},
  author={Tong, Junlong and Xie, Liping and Zhang, Kanjian},
  booktitle={Proceedings of the 2023 SIAM International Conference on Data Mining (SDM)},
  pages={478--486},
  year={2023},
  organization={SIAM}
}

@article{Zhou_Zhang_Peng_Zhang_Li_Xiong_Zhang_2021, title={Informer: Beyond Efficient Transformer for Long Sequence Time-Series Forecasting}, volume={35}, url={https://ojs.aaai.org/index.php/AAAI/article/view/17325}, DOI={10.1609/aaai.v35i12.17325}, abstractNote={Many real-world applications require the prediction of long sequence time-series, such as electricity consumption planning. Long sequence time-series forecasting (LSTF) demands a high prediction capacity of the model, which is the ability to capture precise long-range dependency coupling between output and input efficiently. Recent studies have shown the potential of Transformer to increase the prediction capacity. However, there are several severe issues with Transformer that prevent it from being directly applicable to LSTF, including quadratic time complexity, high memory usage, and inherent limitation of the encoder-decoder architecture. To address these issues, we design an efficient transformer-based model for LSTF, named Informer, with three distinctive characteristics: (i) a ProbSparse self-attention mechanism, which achieves O(L log L) in time complexity and memory usage, and has comparable performance on sequences’ dependency alignment. (ii) the self-attention distilling highlights dominating attention by halving cascading layer input, and efficiently handles extreme long input sequences. (iii) the generative style decoder, while conceptually simple, predicts the long time-series sequences at one forward operation rather than a step-by-step way, which drastically improves the inference speed of long-sequence predictions. Extensive experiments on four large-scale datasets demonstrate that Informer significantly outperforms existing methods and provides a new solution to the LSTF problem.}, number={12}, journal={Proceedings of the AAAI Conference on Artificial Intelligence}, author={Zhou, Haoyi and Zhang, Shanghang and Peng, Jieqi and Zhang, Shuai and Li, Jianxin and Xiong, Hui and Zhang, Wancai}, year={2021}, month={May}, pages={11106-11115} }

@article{ma2023llm,
  title={Llm-pruner: On the structural pruning of large language models},
  author={Ma, Xinyin and Fang, Gongfan and Wang, Xinchao},
  journal={Advances in neural information processing systems},
  volume={36},
  pages={21702--21720},
  year={2023}
}

@article{men2024shortgpt,
  title={Shortgpt: Layers in large language models are more redundant than you expect},
  author={Men, Xin and Xu, Mingyu and Zhang, Qingyu and Wang, Bingning and Lin, Hongyu and Lu, Yaojie and Han, Xianpei and Chen, Weipeng},
  journal={arXiv preprint arXiv:2403.03853},
  year={2024}
}

@article{zhao2025skipgpt,
  title={SkipGPT: Dynamic Layer Pruning Reinvented with Token Awareness and Module Decoupling},
  author={Zhao, Anhao and Ye, Fanghua and Fan, Yingqi and Tong, Junlong and Fei, Zhiwei and Su, Hui and Shen, Xiaoyu},
  journal={arXiv preprint arXiv:2506.04179},
  year={2025}
}

@article{yao2022zeroquant,
  title={Zeroquant: Efficient and affordable post-training quantization for large-scale transformers},
  author={Yao, Zhewei and Yazdani Aminabadi, Reza and Zhang, Minjia and Wu, Xiaoxia and Li, Conglong and He, Yuxiong},
  journal={Advances in neural information processing systems},
  volume={35},
  pages={27168--27183},
  year={2022}
}

@article{sung2023ecoflap,
  title={Ecoflap: Efficient coarse-to-fine layer-wise pruning for vision-language models},
  author={Sung, Yi-Lin and Yoon, Jaehong and Bansal, Mohit},
  journal={arXiv preprint arXiv:2310.02998},
  year={2023}
}

@inproceedings{liang2025efficientllava,
  title={Efficientllava: Generalizable auto-pruning for large vision-language models},
  author={Liang, Yinan and Wang, Ziwei and Xu, Xiuwei and Zhou, Jie and Lu, Jiwen},
  booktitle={Proceedings of the Computer Vision and Pattern Recognition Conference},
  pages={9445--9454},
  year={2025}
}

@article{fan2025visipruner,
  title={VisiPruner: Decoding Discontinuous Cross-Modal Dynamics for Efficient Multimoda},
  author={Fan, Yingqi and Zhao, Anhao and Fu, Jinlan and Tong, Junlong and Su, Hui and Pan, Yijie and Zhang, Wei and Shen, Xiaoyu},
  journal={arXiv preprint arXiv:2510.17205},
  year={2025}
}

@article{kaplan2020scaling,
  title={Scaling laws for neural language models},
  author={Kaplan, Jared and McCandlish, Sam and Henighan, Tom and Brown, Tom B and Chess, Benjamin and Child, Rewon and Gray, Scott and Radford, Alec and Wu, Jeffrey and Amodei, Dario},
  journal={arXiv preprint arXiv:2001.08361},
  year={2020}
}

@inproceedings{aghajanyan2023scaling,
  title={Scaling laws for generative mixed-modal language models},
  author={Aghajanyan, Armen and Yu, Lili and Conneau, Alexis and Hsu, Wei-Ning and Hambardzumyan, Karen and Zhang, Susan and Roller, Stephen and Goyal, Naman and Levy, Omer and Zettlemoyer, Luke},
  booktitle={International Conference on Machine Learning},
  pages={265--279},
  year={2023},
  organization={PMLR}
}

@inproceedings{chen2024internvl,
  title={Internvl: Scaling up vision foundation models and aligning for generic visual-linguistic tasks},
  author={Chen, Zhe and Wu, Jiannan and Wang, Wenhai and Su, Weijie and Chen, Guo and Xing, Sen and Zhong, Muyan and Zhang, Qinglong and Zhu, Xizhou and Lu, Lewei and others},
  booktitle={Proceedings of the IEEE/CVF conference on computer vision and pattern recognition},
  pages={24185--24198},
  year={2024}
}
\bibliographystyle{iclr2026_conference}

\clearpage

\appendix

\section{More Related Work}\label{subsection_appendix:More Related Work}

\subsection{Large Language Models for Time Series} 
Applying fully pre-trained LLMs to TS tasks poses the central challenge of achieving effective modality alignment . Existing efforts can be broadly categorized into two directions: TS embedding and latent space alignment \citep{zhang2024large,pan2024s,liu2025calf}. TS embedding maps raw TS into representations that are more compatible with the pre-trained LLMs space, simplifying knowledge transfer \citep{jin2023time,liu2024unitime}. When combined with prompt-based mechanisms, the strategy leverages intrinsic capacity of LLMs to process TS tasks in a form that preserves generalization abilities \citep{liu2025timecma}. By contrast, latent-space alignment seeks to directly bridge representational gap \citep{zhou2023one,ICLR2025_e1de63ec,liu2025calf}. It is typically achieved through joint training, contrastive learning, or other optimization strategies that enforce deeper cross-modal alignment, enhancing feature sharing and transferability across modalities. One line of work retains original structure of LLMs, mapping TS to input layer \citep{jin2023time}. It maximizes preservation of pretrained parameters and prior knowledge learned from corpora, avoiding instability and retraining overhead that structural alterations may incur. However, such direct adoption often overlooks issues of parameter compatibility, which may compromise efficiency and adaptability in applications. In contrast, another line of work employs only a few shallow layers, which reduces computational complexity and parameter scale, improve inference efficiency and alleviate deployment costs \citep{pan2024s,chang2025llm4ts}. However, it's drawback lies in the loss of pretrained knowledge, diminishing ability to capture local patterns and long-range dependencies. 

\subsection{Time Series Foundation Models} 
TSFMs aim to construct large-scale, general-purpose models for TS tasks by learning on massive heterogeneous datasets, enabling generalized capabilities across tasks \citep{ansari2024chronos,das2024decoder,liu2025sundialfamilyhighlycapable}. TSFMs have demonstrated remarkable potential in zero-shot scenarios, where they can deliver accurate predictions on unseen domains, variable types and temporal granularities, reducing the reliance on task-specific training data that traditional methods typically require. To better accommodate inherent properties of TS—such as continuity, trend, and multi-scale dynamics \citep{liu2024timer}, TSFMs typically incorporate customized adaptations of Transformer architecture. At the same time, they rely on large-scale, cross-domain pretraining to capture diverse patterns from multi-domains. However, most current TSFMs either directly adopt or only slightly modify LLMs architectures, overlooking  fundamental discrepancies between TS and text in terms of statistical properties, dependency structures, and semantic representations \citep{das2024decoder,goswami2024moment,liu2025sundialfamilyhighlycapable}. Consequently, designing architectures and training strategies that explicitly account for statistical and structural uniqueness of TS remains an open challenge for this field. Although current TSFMs have achieved notable performance, they still exhibit limitations in parameter and efficiency, deployment feasibility, inference latency, and cross-task adaptability. 

\subsection{Time Series Representation Learning} 
Representation learning is an indispensable research direction in TS analysis. Effective representations can capture important patterns of TS, reduce redundancy, and enhance performance of downstream tasks. Among current approaches, multi-scale pyramid structures decompose the original TS into hierarchical features at different resolutions, reducing computational overhead \citep{liu2022pyraformer}. Frequency-domain decomposition and frequency-enhanced attention mechanisms, leveraging Fourier or wavelet transforms, optimize seasonal pattern modeling and enable low-complexity representation learning \citep{zhou2022fedformer}. TS can thus be decomposed into seasonal, trend, and residual components to capture multi-scale dynamic patterns \citep{wu2021autoformer}. Next-token prediction employ patch-based inputs, discretization, or modality alignment to pre-train models, enhancing numerical reasoning and inference capabilities \citep{cao2023tempo,xue2023promptcast}. Multi-task learning further integrates reconstruction and prediction errors to improve robustness and expressiveness of learned representations \citep{liu2024unitime}. Despite these advances, investigations into hidden representations and underlying mechanisms of TS models remain limited, which constrains a deeper understanding of their semantic modeling capacity and reasoning process, and affects interpretability and transferability in complex downstream tasks.

\section{Preliminary}
\subsection{Problem Definition}
\label{subsection:Problem Definition}
\noindent TS forecasting constitutes one of the most challenging tasks in TS analysis, due to the inherent non-stationarity, complex temporal dependencies, potential presence of exogenous variables, and often limited availability of high-quality labeled future data for supervision. The core objective is to learn a temporal extrapolation operator that maps a historical observation into a future horizon.

Let $\boldsymbol{X} \in \mathbb{R}^{T \times V}$ denote the observed historical sequence, where $T \in \mathbb{N}^+$ is the input sequence length and $V \in \mathbb{N}^+$ is the number of variables. The goal is to learn a parametric mapping function:
\vspace{+0.3cm}
\[
\boldsymbol{F}_{\boldsymbol{\Theta_F}}: \mathbb{R}^{T \times V} \to \mathbb{R}^{T' \times V}
\]
\vspace{-0.2cm}

It is parameterized by $\boldsymbol{\Theta_F} \in \mathbb{R}^d$, and  predicts the future trajectory $\hat{\boldsymbol{X}}_{[T:T+T')} \in \mathbb{R}^{T' \times V}$ over the next $T' \in \mathbb{N}^+$ time steps. Therefore, given a collection of $K$ distinct time series distributions $\{\mathcal{D}_1, \mathcal{D}_2, \dots, \mathcal{D}_K\}$, and a differentiable loss function $\mathcal{L}: \mathbb{R}^{T' \times V} \times \mathbb{R}^{T' \times V} \to \mathbb{R}_{\geq 0}$, the optimal parameters $\boldsymbol{\Theta_F}^*$ are obtained by minimizing the expected risk over both task distributions. To  evaluate the performance, several standard metrics are employed. Let $\boldsymbol{Y} = \boldsymbol{X}_{[T:T+T')} \in \mathbb{R}^{T' \times V}$ denote the ground truth future values, and $\hat{\boldsymbol{Y}} = \hat{\boldsymbol{X}}_{[T:T+T')} \in \mathbb{R}^{T' \times V}$ the corresponding predictions. Define the element-wise error matrix $\boldsymbol{E} = \hat{\boldsymbol{Y}} - \boldsymbol{Y} \in \mathbb{R}^{T' \times V}$. 
\subsection{Evaluation Metrics}
\label{subsection_appendix:Evaluation Metrics}
\textbf{Mean Absolute Error (MAE)}:
    \[
    \mathrm{MAE} = \frac{1}{T' V} \sum_{t=1}^{T'} \sum_{v=1}^{V} \left| \hat{Y}_{t,v} - Y_{t,v} \right|
    \]

\textbf{Mean Squared Error (MSE)}:
    \[
    \mathrm{MSE} = \frac{1}{T' V} \sum_{t=1}^{T'} \sum_{v=1}^{V} \left( \hat{Y}_{t,v} - Y_{t,v} \right)^2
    \]

\textbf{Mean Absolute Percentage Error (MAPE)} (assuming $Y_{t,v} \neq 0$):
    \[
    \mathrm{MAPE} = \frac{100\%}{T' V} \sum_{t=1}^{T'} \sum_{v=1}^{V} \left| \frac{\hat{Y}_{t,v} - Y_{t,v}}{Y_{t,v}} \right|
    \]

While MAE and MSE emphasize absolute and squared deviations respectively, and MAPE captures relative errors across scales.  In this paper, we focus on MAE and MSE due to their stability and robustness, particularly in scenarios where values may approach zero.

\subsection{Diverse Forecasting Scenarios}

\noindent The practical utility of a forecasting model is critically determined by its data efficiency — its ability to deliver accurate predictions under varying levels of access to target-domain supervision. We formalize three canonical regimes along this spectrum: \emph{full-shot}, \emph{few-shot}, and \emph{zero-shot} learning.

\paragraph{Full-shot learning} 
In the full-shot regime, the model is trained and evaluated on abundant samples from the same distribution $\mathcal{D}_{\text{target}}$. Let $\mathcal{S}_{\text{train}} = \{ (\boldsymbol{X}^{(i)}, \boldsymbol{Y}^{(i)}) \}_{i=1}^N$ with $N \to \infty$ (or sufficiently large) denote the training set drawn from $\mathcal{D}_{\text{target}}$. The objective is to learn parameters $\boldsymbol{\Theta_F}$ that minimize the empirical risk:
\[
\boldsymbol{\Theta_F}^* = \arg\min_{\boldsymbol{\Theta_F}} \frac{1}{N} \sum_{i=1}^N \mathcal{L} \big( \boldsymbol{F}_{\boldsymbol{\Theta_F}}(\boldsymbol{X}^{(i)}), \boldsymbol{Y}^{(i)} \big)
\]
Full-shot learning is the classical supervised learning setup and serves as the performance upper bound for data-scarce regimes.

\paragraph{Few-shot learning} 
In few-shot forecasting, only a small support set $\mathcal{S}_{\text{support}} = \{ (\boldsymbol{X}^{(i)}, \boldsymbol{Y}^{(i)}) \}_{i=1}^M$ from $\mathcal{D}_{\text{target}}$ is available for adaptation. The model, pre-trained on auxiliary distributions $\{\mathcal{D}_1, \dots, \mathcal{D}_K\}$, must rapidly adapt its parameters to minimize the target risk:
\vspace{+0.3cm}
\[
\mathcal{R}_{\text{target}} = \mathbb{E}_{(\boldsymbol{X}, \boldsymbol{Y}) \sim \mathcal{D}_{\text{target}}} \left[ \mathcal{L}\big(\boldsymbol{F}_{\boldsymbol{\Theta_F}}(\boldsymbol{X}), \boldsymbol{Y} \big) \right]
\]
\vspace{-0.1cm}

\paragraph{Zero-shot learning} 
In the most challenging zero-shot regime, \emph{no samples} from $\mathcal{D}_{\text{target}}$ are available during training or adaptation ($\mathcal{S}_{\text{support}} = \emptyset$). The model must generalize purely from inductive biases learned across $\{\mathcal{D}_1, \dots, \mathcal{D}_K\}$. we require:
\vspace{+0.2cm}
\[
\boldsymbol{F}_{\boldsymbol{\Theta_F}} \text{ trained on } \bigcup_{k=1}^K \mathcal{D}_k \Rightarrow \text{low } \mathcal{R}_{\text{target}} \text{ on unseen } \mathcal{D}_{\text{target}}
\]
\vspace{-0.2cm}

\noindent These three regimes form a hierarchy of practical difficulty: full-shot provides the ideal baseline; few-shot tests rapid adaptability; zero-shot evaluates true generalization.

\section{Time Series Modeling}
\label{section:Time Series Modeling}
\subsection{Time Series Embedding}
\label{appendix_subsection:Time Series Embed}
\noindent In TS forecasting, observations $\boldsymbol{X} \in \mathbb{R}^{T \times V}$ are rarely fed directly into  architectures. Instead, they are first projected into a latent embedding space $\mathbb{R}^d$ via a learnable or fixed encoding function $\Phi_{\text{embed}}$, producing a tokenized representation $\boldsymbol{E} \in \mathbb{R}^{N \times d}$. Three principal embedding paradigms have emerged: \textbf{point-wise}, \textbf{patch-wise}, and \textbf{variable-wise embeddings} , each offering distinct inductive biases and trade-offs  \citep{kitaev2020reformer,Yuqietal-2023-PatchTST,ICLR2024_2ea18fdc}.

\paragraph{Point-wise Embedding.} 
In this classical approach, each time step $t$ and each variable $v$ is embedded into a $d$-dimensional vector. For each $(t,v)$, a projection $\phi_{\text{point}}: \mathbb{R} \to \mathbb{R}^d$ is:
\vspace{+0.3cm}
\[
\boldsymbol{e}_{t,v} = \phi_{\text{point}}(X_{t,v}) = \boldsymbol{W}_{\text{emb}} \cdot X_{t,v} + \boldsymbol{b}_{\text{emb}}
\]
\vspace{-0.2cm}

While simple and expressive, point-wise embedding suffers from high token count ($N = T \cdot V$), leading to quadratic attention complexity and difficulty capturing temporal locality.

\paragraph{Variable-wise Embedding.} 
In multivariate settings, variable-wise embedding treats each channel as an independent “token sequence” across time. For variable $v$, the temporal slice $\boldsymbol{X}_{:, v} \in \mathbb{R}^T$ is embedded into a sequence of $T$ tokens via a variable-specific projection:
\vspace{+0.3cm}
\[
\boldsymbol{e}_{:, v} = \phi_{\text{var}}^{(v)}(\boldsymbol{X}_{:, v}) = \boldsymbol{W}_{\text{var}}^{(v)} \cdot \boldsymbol{X}_{:, v} + \boldsymbol{b}_{\text{var}}^{(v)},
\]
\vspace{-0.2cm}

This paradigm, used in models like iTransformer \citep{ICLR2024_2ea18fdc} and Crossformer \citep{zhang2023crossformer}, enables modeling inter-variable dependencies while preserving temporal structure per channel.

\paragraph{Patch-wise Embedding.} 
To enhance local context modeling, patch-wise embedding groups consecutive time steps into “patches” before embedding. Let $P$ denote the patch size, and assume $T$ is divisible by $P$ for simplicity. The sequence is partitioned into $T/P$ patches, each of size $P \times V$, and projected via a learnable linear layer:
\vspace{+0.3cm}
\[
\boldsymbol{e}_i = \phi_{\text{patch}}\left( \boldsymbol{X}_{[(i-1)P : i P), :} \right) = \boldsymbol{W}_{\text{patch}} \cdot \mathrm{vec}\left( \boldsymbol{X}_{[(i-1)P : i P), :} \right) + \boldsymbol{b}_{\text{patch}}
\]
\vspace{-0.2cm}

With patch-wise embedding currently dominating long-sequence forecasting due to its balance of efficiency and expressiveness, we adopt Patch-wise Embedding in the present work.

\subsection{Transformer Blocks}
\label{subsection_appendix:transformer-blocks}
The transformer architecture fundamentally relies on attention mechanisms to model dependencies within and across sequences. We formalize three canonical attention mechanisms that underpin modern transformer designs.

\paragraph{Vanilla Self-Attention} 
 Given an embedded sequence $\boldsymbol{E} \in \mathbb{R}^{N \times d}$, the scaled dot-product attention is defined as:
 \vspace{+0.3cm}
\[
    \mathrm{Attention}(\boldsymbol{Q}, \boldsymbol{K}, \boldsymbol{V}) = \mathrm{softmax}\left( \frac{\boldsymbol{Q} \boldsymbol{K}^\top}{\sqrt{d}} \right) \boldsymbol{V}
\]
\vspace{-0.2cm}

where $\boldsymbol{Q} = \boldsymbol{E} \boldsymbol{W}^Q$, $\boldsymbol{K} = \boldsymbol{E} \boldsymbol{W}^K$, $\boldsymbol{V} = \boldsymbol{E} \boldsymbol{W}^V$ are linear projections, and $\boldsymbol{W}^Q, \boldsymbol{W}^K, \boldsymbol{W}^V \in \mathbb{R}^{d \times d_k}$. It allows unrestricted information flow across all time steps, making it suitable for encoding historical context, but unsuitable for autoregressive generation, as it violates temporal causality.

\paragraph{Cross-Attention} 
In encoder-decoder architectures, cross-attention enables the decoder to attend to encoded representations of the input sequence \citep{ansari2024chronos}. Let $\boldsymbol{E}_{\text{enc}} \in \mathbb{R}^{N \times d}$ denote encoder outputs, and $\boldsymbol{E}_{\text{dec}} \in \mathbb{R}^{M \times d}$ denote decoder inputs:
 \vspace{+0.3cm}
\[
    \mathrm{CrossAttention}(\boldsymbol{E}_{\text{dec}}, \boldsymbol{E}_{\text{enc}}) = \mathrm{softmax}\left( \frac{\boldsymbol{Q}_{\text{dec}} \boldsymbol{K}_{\text{enc}}^\top}{\sqrt{d}} \right) \boldsymbol{V}_{\text{enc}}
\]
\vspace{-0.2cm}

where $\boldsymbol{Q}_{\text{dec}} = \boldsymbol{E}_{\text{dec}} \boldsymbol{W}^Q$, $\boldsymbol{K}_{\text{enc}} = \boldsymbol{E}_{\text{enc}} \boldsymbol{W}^K$, $\boldsymbol{V}_{\text{enc}} = \boldsymbol{E}_{\text{enc}} \boldsymbol{W}^V$. Cross-Attention enables conditional generation and is widely used in early transformer forecasting models, but introduces architectural complexity and often suffers from error propagation during auto-regressive decoding.

\paragraph{Causal Self-Attention.} 
To enable autoregressive modeling, where prediction at time $t$ depends only on $\{1, \dots, t-1\}$, a causal mask $\boldsymbol{M} \in \mathbb{R}^{N \times N}$ is applied:
 \vspace{+0.3cm}
\[
    \mathrm{CausalAttention}(\boldsymbol{Q}, \boldsymbol{K}, \boldsymbol{V}) = \mathrm{softmax}\left( \frac{\boldsymbol{Q} \boldsymbol{K}^\top}{\sqrt{d}} + \boldsymbol{M} \right) \boldsymbol{V}
\]
\vspace{-0.2cm}

Causal attention is the cornerstone of decoder-only transformers, which treat forecasting as a pure sequence completion task: the model is trained to predict future values conditioned on past context. Recent advances in large-scale modeling, particularly in language and multimodal domains, have demonstrated the superiority of decoder-only architectures with causal attention. Following the trend in large-scale modeling, we adopt a decoder-only transformer with causal self-attention for its simplicity, scalability, and strong empirical performance, enabling direct multi-step prediction without iterative error accumulation or encoder-decoder synchronization.

\subsection{Prediction Head}
\label{subsection_appendix:Prediction Head}
The hidden states from the final transformer layer $\boldsymbol{H}^{L-1} \in \mathbb{R}^{N \times d}$ are mapped to the target forecasting space through a linear Prediction Head. Let $\boldsymbol{W}_{\textrm{out}} \in \mathbb{R}^{d \times D}$ denote the learnable projection matrix, where $D = T' \cdot V$ is the total dimensionality of the prediction horizon, and $\boldsymbol{Y} \in \mathbb{R}^{T' \times V}$ represents the final predictions:
 \vspace{+0.3cm}
\[
    \boldsymbol{Y} = \mathrm{reshape}\left( \mathrm{LN}(\boldsymbol{H}^{L-1}) \boldsymbol{W}_{\textrm{out}} \right)
\]
\vspace{-0.2cm}

This direct multi-step mapping predicts all future time steps in a single forward pass, avoiding the error accumulation inherent in auto-regressive decoding. In our case, the use of a single-pass Linear further enhances efficiency by enabling lightweight computation and fast inference without sacrificing performance.

\section{Datasets}
\label{subsection_appendix:Datasets}
\subsection{Datasets of Single-Dataset Learning}
\label{subsection_appendix:single-dataset}

In the \emph{single-dataset learning} setting, models are trained, validated, and tested on temporally contiguous splits drawn from a single TS distribution. It evaluates the model’s in-domain \& full-shot learning capability, learn and extrapolate temporal patterns when abundant target-domain data is available during training. Datasets of single-Dataset learning:

\paragraph{ETT:} It includes four variants: ETTh1 and ETTh2 (hourly) and ETTm1 and ETTm2 (15-minutely). ETT are collected from two distinct regions over approximately two years. Each contains 7 variables, such as oil temperature, load, and equipment status and exhibit strong periodicity, trend drifts, and occasional abrupt shifts, making them standard benchmarks for long-term forecasting.

\vspace{-0.4cm}
\paragraph{ECL:} Hourly electricity consumption records from 321 clients over four years. It features high volatility, heterogeneous load profiles, and complex inter-client correlations, challenging models to capture both global trends and individual behavioral dynamics.

\vspace{-0.4cm}
\paragraph{Exchange:} Daily exchange rates of eight major currencies — Australia, UK, Canada, Switzerland, China, Japan, New Zealand, and Singapore, relative to the US dollar, spanning 16 years. It reflects slow macroeconomic trends, global financial coupling, and rare regime shifts, ideal for evaluating robustness under low-frequency non-stationarity.

\vspace{-0.4cm}
\paragraph{Solar:} 10-minute resolution solar power production from 137 stations in Alabama over one year. Dominated by strong diurnal cycles and weather-induced intermittency, it presents challenges in modeling fine-grained, spatially correlated renewable generation.

\vspace{-0.4cm}
\paragraph{Weather:} 21 meteorological variables, including air temperature, humidity, wind speed, and pressure, are recorded hourly at a German weather station over eight years. Exhibits rich multi-scale dynamics, suitable for evaluating hierarchical temporal modeling.

\vspace{-0.4cm}
\paragraph{Traffic:} It contains hourly road occupancy rates, the percentage of time a road segment is occupied by vehicles, and are recorded across 862 sensors on San Francisco Bay Area freeways over two years. It exhibits complex spatio-temporal correlations, strong weekly seasonality, and sharp rush-hour peaks, making it a challenging benchmark for modeling large-scale, real-world urban dynamics.

\paragraph{Data Splitting Protocol.} To ensure consistency and avoid temporal leakage, all datasets are partitioned chronologically. For the four ETT datasets, we follow the established 8:4:4 month-wise split: first 8 months for training, next 4 for validation, last 4 for testing. For all other datasets, we use a 7:1:2 ratio by sequence length: 70\% training, 10\% validation, 20\% testing. All variables are standardized using training-set statistics only. No external covariates or calendar features are used, meaning models rely solely on historical observations.

\subsection{Datasets of Cross-Dataset Learning}
\label{subsection_appendix:Datasets of Cross-Dataset Learning & Evaluation}

\noindent In the cross-dataset learning setting, models are trained on the union of training sets from multiple distinct TS distributions, validated on the union of validation sets, and finally evaluated separately on each dataset’s test set. This protocol evaluates a model’s ability to learn universal temporal representations that generalize across domains. Unlike single-dataset learning, this setup tests robustness to distribution shift, and parameter efficiency under heterogeneous data. We extend the dataset collection, resulting in a total of 41 datasets for cross-dataset experiments. During training, batches are sampled uniformly across datasets to ensure balanced exposure. At test time, metrics are computed individually per dataset to assess both in-domain and full-shot learning performance.

\noindent To further evaluate the out-of-domain generalization and zero-shot learning performance of models trained under the cross-dataset protocol, we introduce a held-out evaluation suite comprising eight additional TS datasets. These datasets are never seen during training or validation. Only their test sets are used for final evaluation, providing a strict zero-shot benchmark that measures how well learned temporal representations transfer to completely unseen domains.Datasets of cross-dataset learning:

\paragraph{NN5:} A collection of 111 daily TS representing sales records of non-food items from an anonymous retail chain in the UK, spanning 2 years. Characterized by intermittent demand, promotional spikes, and calendar-driven seasonality.

\vspace{-0.4cm}
\paragraph{PDB (Protein Data Bank):} Weekly counts of new protein structure submissions to the PDB from 2000 to 2022. Exhibits slow, science-policy-driven growth trends with occasional abrupt shifts due to technological or institutional changes.

\vspace{-0.4cm}
\paragraph{Sceaux:} Hourly measurements of temperature, humidity, and pressure recorded at a weather station in Sceaux, France, over 5 years. Features fine-grained meteorological dynamics with strong diurnal and seasonal cycles.

\vspace{-0.4cm}
\paragraph{Smart:} Electricity consumption readings from 30 smart meters in Irish households over 18 months at 30-minute intervals. Captures heterogeneous household behavior, appliance-level patterns, and weather-sensitive usage.

\vspace{-0.4cm}
\paragraph{Spanish:} Daily electricity demand across Spain from 2014 to 2019. Reflects national-scale consumption modulated by economic activity, holidays, temperature, and renewable penetration.

\vspace{-0.4cm}
\paragraph{Sunspot-rain:} A bivariate dataset combining monthly international sunspot numbers and regional rainfall in Eastern Asia from 1850 to 2020. Offers an ultra-long-term, low-frequency view of potential solar-terrestrial coupling.

\vspace{-0.4cm}
\paragraph{USbirths:} Daily counts of births in the United States from 1969 to 1988. Displays strong weekly periodicity, holiday dips, and long-term demographic trends.

\vspace{-0.4cm}
\paragraph{Wind-power:} Hourly wind power generation from 20 turbines in a European wind farm over one year. Highly volatile and non-stationary, governed by weather dynamics and turbine-specific efficiency curves.

\paragraph{Evaluation Protocol.} Critically, for all eight datasets above, only the test set is used to test and no training or validation samples are exposed to the model at any stage. It ensures a pure zero-shot evaluation: models must forecast these sequences based solely on representations learned from the original nine datasets. 

\subsection{Datasets for Baseline Representation Analysis and Pruning Studies}
\label{subsection_appendix:Datasets for Baseline Representation Analysis and Pruning Studies}

\noindent To evaluate the generality of our findings, particularly in representation learning, critical layers localization and model compression, we extend our datasets with 4 additional traffic flow datasets from the PEMS (Performance Measurement System) family: PEMS03, PEMS04, PEMS07, and PEMS08. These datasets provide large-scale,w real-world transportation dynamics and serve as stress tests for structural robustness and efficiency under distribution shift.

\paragraph{PEMS03:} Hourly traffic occupancy rates recorded by 358 sensors across the San Francisco Bay Area over 3 months in 2018. Exhibits strong spatial correlation and morning/evening rush-hour patterns.

\vspace{-0.4cm}
\paragraph{PEMS04:} Data from 307 sensors in the same region over 3 months in 2018, capturing similar dynamics to PEMS03 but with distinct sensor coverage and flow characteristics.

\vspace{-0.4cm}
\paragraph{PEMS07:} Records from 883 sensors in the Los Angeles metropolitan area over 6 months in 2017. Features higher spatial density and more complex congestion propagation patterns.

\vspace{-0.4cm}
\paragraph{PEMS08:} Measurements from 170 sensors in San Bernardino County over 6 months in 2016. Reflects suburban traffic dynamics with lower density but longer commute corridors.

\paragraph{Data Splitting Protocol.} All four PEMS datasets follow a 7:1:2, 70\% for training, 10\% for validation, and 20\% for testing. Variables are standardized per dataset using training-set statistics only. No external features are included, ensuring models rely solely on learned temporal-spatial representations. PEMS are not used in cross-dataset learning or zero-shot evaluation. 

\section{Implementation Details}
\subsection{Model Family Configurations}
\label{subsection_appendix:Model Family Configurations}
Important architectural parameters are reported :model layers, hidden channels, total learnable parameters, attention heads, positional encoding type, and embedding/prediction layers. All models use patch size 16 and stride 8 for time series embedding with linear input/output projections. We employ the commonly used setting of 336 input steps and 96 steps for future prediction.
\begin{table}[h]
  \centering
    \caption{Model Family Configurations. Model configurations across two families, LLM4TS and TSFMs, varying in scale from Tiny to Large. “–” indicates no pre-training (trained from scratch).
}
  \resizebox{\linewidth}{!}
  {
  \begin{tabular}{lccccccc}
    \toprule
    \textbf{Scale} & \textbf{Model family} & \textbf{Layers} & \textbf{Backbone} & \textbf{Channels} & \textbf{Learnable parameters} & \textbf{Patch Size}& \textbf{Stride}\\
    \midrule
    Tiny  & LLM4TS / TSFMs & 6 / 6  & GPT-2 / -& 768 / 768   & 3.92M / 85M & 16 &8\\
    Small & LLM4TS / TSFMs & 12 / 12 & GPT-2 / -& 768 / 768   & 3.93M / 128M & 16 &8\\
    Base & LLM4TS / TSFMs & 28 / 28 & Qwen-3 / -& 1,024 / 1,024 & 0.16B / 0.60B & 16 &8\\
    Large & LLM4TS / TSFMs & 28 / 28 & Qwen-3 / -& 2,048 / 2,048 & 0.32B / 1.73B & 16 &8\\
    \bottomrule
  \end{tabular}
}
  \label{fig:Model Family Configurations}
\end{table}

We select two representative LLMs as backbones for the LLM4TS family: GPT-2 \citep{radford2018improving} and Qwen-3 \citep{yang2025qwen3technicalreport}. GPT-2 is a decoder-only Transformer pretrained on large-scale web text, employing learnable absolute positional embeddings and layer normalization before attention . Its simplicity, widespread adoption, and availability in multiple scales make it an ideal baseline for studying LLM adaptation to TS. Qwen-3 is a state-of-the-art Chinese-English bilingual LLM that adopts RoPE and RMSNorm, demonstrating superior long-context modeling and instruction-following capabilities. We include Qwen-3 to examine whether modern architectural advances, particularly relative position awareness and better scaling, and translate into improved TS forecasting performance. By repurposing LLMs without modifying their core architecture, we establish a strong, reproducible baseline for evaluating how well pretrained linguistic representations can be transferred to temporal forecasting.

\subsection{Implementation Details for Sec. \ref{subsection:Does Backbone Scaling Improve Forecasting Performance? (RQ1)} \& \ref{subsection:Does Limited Data Volumn Hinder the Advantages of Backbone? (RQ2)}}
\label{subsection_appendix:Implementation Details for Sec. for 4.1 & 4.2}
In Sec. \ref{subsection:Does Backbone Scaling Improve Forecasting Performance? (RQ1)}, all experiments are implemented using the \texttt{HuggingFace Transformers} library (v4.51.3) and trained with \texttt{DeepSpeed} (v0.14.0) and \texttt{Accelerate} (v0.28.0) for distributed training \citep{wolf-etal-2020-transformers}. We use either 4$\times$ NVIDIA A100 40GB GPUs or 4$\times$ NVIDIA H100 80GB GPUs, depending on model scale. Training is conducted with a fixed random seed across all runs to ensure reproducibility. We employ the AdamW optimizer \citep{loshchilov2019decoupledweightdecayregularization} with learning rate $1 \times 10^{-4}$, $\beta_1 = 0.9$, $\beta_2 = 0.95$, and $\epsilon = 1 \times 10^{-6}$. Weight decay is set to 0.01. The learning rate follows a cosine decay schedule without warmup. Training uses \texttt{bf16} mixed precision and DeepSpeed ZeRO Stage 2 for memory efficiency, with default gradient clipping. Gradient accumulation steps are set to 1, and micro batch size per GPU is fixed at 128. Early stopping is applied based on MSE of validation sets, with a patience of 3 epochs. All components are built on standard HuggingFace and PyTorch APIs.

In Sec. \ref {subsection:Does Limited Data Volumn Hinder the Advantages of Backbone? (RQ2)}, to investigate whether the performance advantages of advanced backbone architectures are contingent on abundant training data, we conduct a controlled data-scaling study. We vary the proportion of training data sampled from each dataset — from 20\% to 100\%, in increments of 20\%, while keeping all other experimental settings identical to Sec. \ref{subsection:Does Backbone Scaling Improve Forecasting Performance? (RQ1)}.
Validation and test sets remain fixed and unchanged across runs to ensure consistent evaluation.

\subsection{Implementation Details for Sec. \ref{Does Performance Degradation Stem From Dataset Homogeneity? (RQ3)}}
\label{subsection_appendix:Implementation Details for Subsection for 4.3}
In the cross-dataset learning setting, we combine the training partitions of all source datasets into a unified corpus. To avoid bias introduced by dataset ordering or temporal structure during training, we fully shuffle all samples globally before each epoch, ensuring that batches are statistically diverse and model updates are not dominated by any single domain. Due to the large volume of combined TS data and memory constraints, we preprocess and store all datasets in \texttt{Apache Arrow} format using the \texttt{datasets} library (v4.0.0), which enables efficient, zero-copy data loading and minimizes I/O bottlenecks. 

All models are trained on 4$\times$ NVIDIA H200 140GB GPUs, using the same codebase and hyperparameters as in Sec. \ref{subsection:Does Backbone Scaling Improve Forecasting Performance? (RQ1)}, with the following specific adjustments: per-GPU batch size is set to 256, training is capped at a maximum of 3 epochs, and early stopping is triggered if no improvement in validation MSE is observed for 1 epoch. No curriculum learning, dataset weighting, or domain balancing strategy is applied. Models learn from uniformly sampled batches across all source domains, testing their raw capacity to acquire generalizable temporal representations under maximal data diversity.

\subsection{Implementation Details for Sec. \ref{subsection:Method Adaptation for Baselines}}
\label{subsection_appendix:Implementation Details for Subsection for 5}

For four LLM4TS baselines in Sec. \ref{section:Experiments and Analysis}, we adopt channel-independent strategy, each variable is processed as an independent sequence. All reproduction and pruning experiments strictly adhere to same strategy and training hyperparameters reported in original papers, ensuring faithful re-implementation. Critically, the selection of trainable versus frozen modules is held identical across all models and experimental phases. For four TSFMs baselines, All models are fine-tuned on the downstream datasets to improve their generalization capability and to align the parameter distributions before and after pruning, thereby ensuring a fair and meaningful comparison of pruning effects. This eliminates confounding factors arising from training protocol differences and allows us to attribute performance changes solely to architectural modification.

\section{Baselines}
\label{Section_appendix:Baselines}
\subsection{Baselines of LLM4TS}
\textsc{Fsca}\citep{ICLR2025_e1de63ec}: Introducing Context-Alignment to comprehend TS with the same structural and logical awareness they apply to natural language. It achieves this through two complementary alignment mechanisms: (1) \emph{Structural Alignment}, which employs dual-scale graph nodes to capture the hierarchical composition of TS, allowing LLMs to treat long sequences as coherent linguistic units while preserving fine-grained temporal features; and (2) \emph{Logical Alignment}, which utilizes directed edges to model semantic dependencies across variables or time steps, ensuring contextual coherence and relational consistency.

\textsc{Calf}\citep{liu2025calf}: A dual-branch framework is employed, where aligned textual tokens and projected TS tokens are processed through the same frozen pretrained LLM layers to extract semantically aligned features. Cross-modal interaction is established via three alignment mechanisms: linear projection, cross-attention fusion, and contrastive tuning, enabling temporal forecasting grounded in linguistic priors without modifying the original LLM parameters.

\textsc{Time-LLM}\citep{jin2023time}: The pretrained LLM is kept entirely frozen, and TS forecasting is enabled through two lightweight trainable modules, \emph{Patch Reprogramming} for input adaptation and \emph{Output Projection} for prediction mapping. It adopt a channel-independent strategy that decomposes multivariate forecasting into parallel univariate tasks.

\textsc{Ofa}\citep{zhou2023one}: A unified TS analysis framework is established by repurposing a frozen pre-trained language model without modifying its internal Transformer layers. By treating diverse TS tasks, including short- and long-term forecasting, classification, imputation, and anomaly detection, as sequence modeling problems within a common interface, the approach achieves highly competitive performance across all settings. Input TS are projected into the embedding space via lightweight trainable adapters, while outputs are mapped back through task-specific heads, preserving the LLM’s original parameters throughout. 

\subsection{Baselines of TSFMs}
\textsc{Sundial}\citep{liu2025sundialfamilyhighlycapable}: A family of native TS foundation models is pre-trained on TimeBench, a corpus containing one trillion time points from real-world and synthetic datasets. Training employs a flow-matching-based TimeFlow Loss that enables direct optimization on continuous-valued sequences without discrete tokenization. By modeling the next-patch distribution in a non-parametric generative manner, the approach supports multi-modal probabilistic forecasting and effectively mitigates mode collapse. With only minimal adaptations to the Transformer architecture, the models accept arbitrary-length inputs, achieve strong scalability, and deliver state-of-the-art performance in both point and probabilistic forecasting, while enabling zero-shot inference in just a few milliseconds.

\textsc{Chronos}\citep{ansari2024chronos}: It adapts TS to Transformer-based models through a two-step preprocessing pipeline of scaling and quantization. First, mean-scaling normalizes each value by the mean absolute magnitude of its historical context, ensuring comparability across series. Second, quantization discretizes scaled values into bins, yielding a sequence of discrete tokens. These tokens are modeled using a T5-based architecture trained with cross-entropy loss, enabling the acquisition of general-purpose TS representations while fully leveraging modeling paradigm.

\textsc{Moirai}\citep{moirai}: It is a large-scale universal time series forecasting model designed to overcome the limitations of the traditional one-model-per-dataset paradigm. By introducing architectural innovations to the standard Transformer, \textsc{Moirai} effectively handles cross-frequency learning, supports arbitrary multivariate input dimensions, and adapts to diverse distributional patterns across large-scale datasets. Trained on the Large-scale Open Time Series Archive containing over 27 billion observations across nine domains, \textsc{Moirai} demonstrates strong zero-shot forecasting capability, achieving performance comparable to or even surpassing fully fine-tuned models.

\textsc{Times-FM}\citep{das2024decoder}: As a patch-based forecasting framework, it supports variable context lengths and enables output patches longer than input patches. During training, model learns to predict extended horizons by conditioning on progressively longer prefixes. At inference time, long-term forecasting is performed in a semi-autoregressive manner: given a 256-step context, it predicts next 128 steps, then appends predictions to original context to forecast subsequent 128 steps. Internally, each decoding step processes current input patch through an input residual block, adds a positional encoding vector, feeds the result into a stacked Transformer with causal self-attention to ensure temporal consistency, and finally passes the representation through an output residual block to produce the output patch, which is compared against ground truth for loss computation.

\end{document}